%% file: main.tex
\pdfoutput=1
\documentclass[letterpaper,journal]{IEEEtran}

\usepackage{amsmath,amssymb,amsfonts,amsthm,mathtools}
\usepackage{array}
\usepackage{booktabs}
\usepackage{textcomp}
\PassOptionsToPackage{hyphens}{url}
\usepackage{url}
\usepackage{graphicx}
\usepackage{cite}
\usepackage{xcolor}
\usepackage{microtype}
\usepackage{tikz}
\usetikzlibrary{arrows.meta,positioning,shapes.geometric,fit,calc}
\usepackage{hyperref}

\definecolor{primary}{RGB}{15,41,66}
\definecolor{accent}{RGB}{11,87,208}
\definecolor{alert}{RGB}{179,38,30}
\definecolor{surface}{RGB}{244,247,251}

\hypersetup{
  colorlinks=true,
  linkcolor=primary,
  citecolor=accent,
  urlcolor=accent,
  pdftitle={Append-Observational Geometry and Nonlinear Widths of Softmax Attention States},
  pdfauthor={Jun He and Deying Yu}
}

\allowdisplaybreaks[2]

\newtheorem{theorem}{Theorem}
\newtheorem{lemma}[theorem]{Lemma}
\newtheorem{proposition}[theorem]{Proposition}
\newtheorem{corollary}[theorem]{Corollary}
\theoremstyle{definition}
\newtheorem{definition}{Definition}

\theoremstyle{remark}
\newtheorem{remark}{Remark}

\newcommand{\parhead}[1]{\par\smallskip\noindent\textbf{#1}\ }

\newcolumntype{L}[1]{>{\raggedright\arraybackslash}p{#1}}

\newcommand{\R}{\mathbb{R}}
\newcommand{\Z}{\mathbb{Z}}
\newcommand{\Q}{\mathcal{Q}}
\newcommand{\K}{\mathcal{K}_M^+}
\newcommand{\F}{\mathcal{F}}
\newcommand{\dapp}{d_{\mathrm{app}}}
\newcommand{\dQ}{d_{\mathcal{Q}}}
\newcommand{\dphys}{d_{\mathrm{phys}}}
\newcommand{\dQphys}{d_{\mathcal{Q}}^{\mathrm{phys}}}
\newcommand{\norm}[1]{\left\lVert #1\right\rVert}
\newcommand{\ip}[2]{\left\langle #1,#2\right\rangle}
\newcommand{\diag}{\operatorname{diag}}
\newcommand{\rank}{\operatorname{rank}}

\newcommand{\Var}{\operatorname{Var}}
\newcommand{\ind}{\mathrm{ind}}
\newcommand{\joint}{\mathrm{joint}}
\newcommand{\sh}{\mathrm{sh}}
\newcommand{\priv}{\mathrm{priv}}
\newcommand{\e}{\varepsilon}

\title{Append-Observational Geometry and Nonlinear Widths of Softmax Attention States}

\author{Jun~He and Deying~Yu%
\thanks{The authors are with OpenKedge LLC (e-mail: \{junhe, deying\}@openkedge.io).}}

\begin{document}

\maketitle

\begin{abstract}
Softmax attention exposes normalized outputs while concealing historical normalization mass, so distinct histories agreeing on every current query can diverge after receiving the same new key--value tokens. We model attention states as additive numerator--partition pairs equipped with an append-observational metric measuring worst-case distinguishability under common continuations, together with a positive-cone embedding establishing maximum-error merge stability and quantitative conditions for append-robust approximation. Using continuous nonlinear widths, we construct variable-key histories with identical normalized outputs that nonetheless require full local representation dimension under future appends---both in one dimension and, via an orthogonal-axis construction ($d_k \ge L$), at an $L$-independent macroscopic threshold surviving repeated permitted unit-weight appends within a linear suffix horizon $H_* = O(L)$. For multi-session sharing, we establish matching collective dimensions separating shared from private information both for a fixed-key circular harmonic family and for a variable-key family driven purely by hidden historical mass ($F_{S_i} \equiv M$). Finally, we establish indefinite-horizon maintenance via positive moment-preserving recompression, extend it to approximate numerical reductions (with logarithmic dyadic drift and non-accumulating kernel error) and relative-precision weight quantization ($O(\log\log T)$ weight bit-width), and show how existing attention coresets inherit append-robust certificates after appropriate reweighting. All results address bounded single-head softmax attention under open-loop continuations.
\end{abstract}


\section{Introduction}
\label{sec:intro}

\IEEEPARstart{A}{utoregressive} transformers retain a growing history of keys and values so that future queries can attend over past context. Shared-prefix execution, vector quantization, coreset subsampling, and low-rank summaries compress this history differently, yet all face a foundational mathematical question: \emph{what representation complexity is required to preserve attention states that are indistinguishable under current normalized observations but distinguishable under future common continuations?}

Let $q, k_j \in \R^{d_k}$ and $v_j \in \R^{d_v}$, absorbing the standard $1/\sqrt{d_k}$ scaling into the keys. For a finite key--value history of length $L$, single-head softmax attention evaluates
\begin{equation}\label{eq:base}
\begin{aligned}
  F_S(q) &= \frac{N(q)}{D(q)},\\
  N(q) &= \sum_{j=1}^{L} v_j e^{q^\top k_j},\qquad
  D(q) = \sum_{j=1}^{L} e^{q^\top k_j}.
\end{aligned}
\end{equation}
The numerator $N$ and partition function $D$ form an additive state pair $S = (N, D)$ under appends and disjoint-history merges. Rescaling $S$ to $cS$ ($c > 0$) leaves $F_S(q)$ unchanged for every $q \in \R^{d_k}$, yet appending a common suffix $X$ weights the historical contribution differently for $S$ and $cS$. While exact key--value tensor reconstruction preserves append behavior (and sufficiently accurate tensor reconstruction can control append error under boundedness assumptions), tensor-reconstruction losses do not directly characterize the intrinsic dynamic equivalence classes of attention states or their minimal representation complexity. Conversely, evaluating compression solely through current normalized outputs misses the historical normalization mass governing how past context competes with future tokens.

Prior work has recognized the operational role of normalization mass from several angles: shared-prefix kernels merge output and log-normalizer statistics~\cite{cascade,flashinfer,hydragen}, query-oblivious attention coresets control numerator and denominator errors jointly inside their proofs~\cite{liberty,cohen}, residual-cache methods restore omitted denominator mass~\cite{reskv}, and Yang and Li~\cite{yang2026} derive an a posteriori future-query error bound governed by the hyperbolic tangent of log-partition discrepancy. Building on this recognition, our primary contribution is to quantify the \emph{continuous nonlinear representation complexity} forced by this dynamic observability gap---particularly for variable-key histories whose current normalized outputs are identically constant---and to connect dynamic equivalence to compositional approximation geometry, collective widths, and certified state maintenance.

Note also why raw score-matrix rank is an unreliable complexity measure: the rank-one score matrix $(x_i y_j)$ maps under entrywise exponentiation to $(e^{x_i y_j})$, which is strictly totally positive for strictly increasing scalar nodes~\cite{gk,karlin} and hence full rank whenever query and key counts coincide; row normalization preserves rank. Low-dimensional keys need not yield low-rank attention operators, whereas full-rank exponential operators can still exhibit rapidly decaying singular values. We therefore measure complexity directly in the geometry of append-observable state functions.

\begin{figure*}[!t]
\centering
\resizebox{\textwidth}{!}{%
\begin{tikzpicture}[
  font=\footnotesize\sffamily,
  box/.style={draw=primary, thick, rounded corners=3pt, fill=surface, minimum height=10.5mm, align=center, inner sep=3.4pt},
  core/.style={draw=accent, very thick, rounded corners=3pt, fill=accent!8, minimum height=10.5mm, align=center, inner sep=3.6pt},
  leaf/.style={draw=primary!80, semithick, rounded corners=2pt, fill=white, minimum width=80mm, minimum height=8.2mm, align=center, inner sep=3.0pt},
  arr/.style={-{Stealth[length=2.2mm,width=1.5mm]}, thick, color=primary},
  bluearr/.style={-{Stealth[length=2.2mm,width=1.5mm]}, thick, color=accent}
]
\node[box, minimum width=44mm] (state)
  {\textbf{Additive State Space $\K$}\\
   \scriptsize $S=(N,D)\in C(\Q,\R^{d_v})\!\times\! C(\Q,\R)$\\[-1pt]
   \scriptsize $D>0,\ \norm{N}\le MD$};

\node[core, minimum width=48mm, right=7.5mm of state] (geom)
  {\textbf{Dynamic Observability Geometry}\\
   \scriptsize Append metric $\dQ$ \& identifiability (Sec.~\ref{sec:algebra})\\[-1pt]
   \scriptsize Positive lift $P_S\mapsto\tau$: $\max$-merge stability};

\node[leaf, right=10mm of geom] (coll)
  {\textbf{Collective Widths (Sec.~\ref{sec:collective})}\\
   \scriptsize Fixed-key $\Q_L$ harmonic \& variable-key hidden-mass ($F_i{\equiv}M$);\\[-1pt]
   \scriptsize Exact $R_{\joint}(\e){=}s{+}Bp$, gain $G_C{=}\tfrac{B(s+p)}{s+Bp}$};

\node[leaf, above=3.2mm of coll] (width)
  {\textbf{Single-Session Widths (Sec.~\ref{sec:width})}\\
   \scriptsize Exact $r{\ge}L(d_k{+}d_v)$; fixed-key spectral widths;\\[-1pt]
   \scriptsize Hidden-mass ($F{\equiv}M$): 1D \& orth.\ ($d_k{\ge}L$), $r_{\min}{=}L$, $H_*{\,=\,}O(L)$};

\node[leaf, below=3.2mm of coll] (maint)
  {\textbf{Compositional Maintenance (Sec.~\ref{sec:maintenance})}\\
   \scriptsize Sharp kernel distortion; binary-counter bounds; positive cubature;\\[-1pt]
   \scriptsize $O(\delta_{\mathrm{num}}\log T)$ solver drift \& $O(\log\log T)$ weight bits};

\node[above=2.0mm of state, font=\footnotesize\bfseries\sffamily, color=primary] {1. State Algebra};
\node[above=2.0mm of geom, font=\footnotesize\bfseries\sffamily, color=accent] {2. Append Metric \& Cone Lift};
\node[above=2.0mm of width, font=\footnotesize\bfseries\sffamily, color=primary] {3. Widths, Sharing \& Maintenance};

\coordinate (bus) at ($(geom.east)!0.48!(coll.west)$);
\draw[bluearr] (state.east) -- (geom.west);
\draw[thick, color=accent] (geom.east) -- (bus);
\draw[bluearr] (bus) |- (width.west);
\draw[bluearr] (bus) -- (coll.west);
\draw[bluearr] (bus) |- (maint.west);
\end{tikzpicture}%
}
\caption{Relationship between dynamic observability, continuous nonlinear representation widths, and compositional state maintenance. The append-observational metric $\dQ$ and positive-cone lift $P_S\mapsto\tau$ on $\K$ link hidden-mass single-session widths (1D local matching $r_{\min}(\e)=L$ in Theorem~\ref{thm:w4} and Corollary~\ref{cor:w4phys}, plus the well-conditioned orthogonal-axis counterpart with $L$-independent threshold $\e_{\mathrm{orth}}$ and linear physical horizon $H_* = O(L)$ in Theorem~\ref{thm:w4orth}), collective dimension separation (matching $R_{\joint}(\e)=s+Bp$ for fixed-key circular families on $\Q_L$ in Theorem~\ref{thm:c2} and variable-key hidden-mass families on general $\Q$ in Theorem~\ref{thm:c3}), and certified maintenance with numerical-drift and weight-quantization bounds (Section~\ref{sec:maintenance}, Corollaries~\ref{cor:numdrift}--\ref{cor:bitwidth}).}
\label{fig:overview}
\end{figure*}
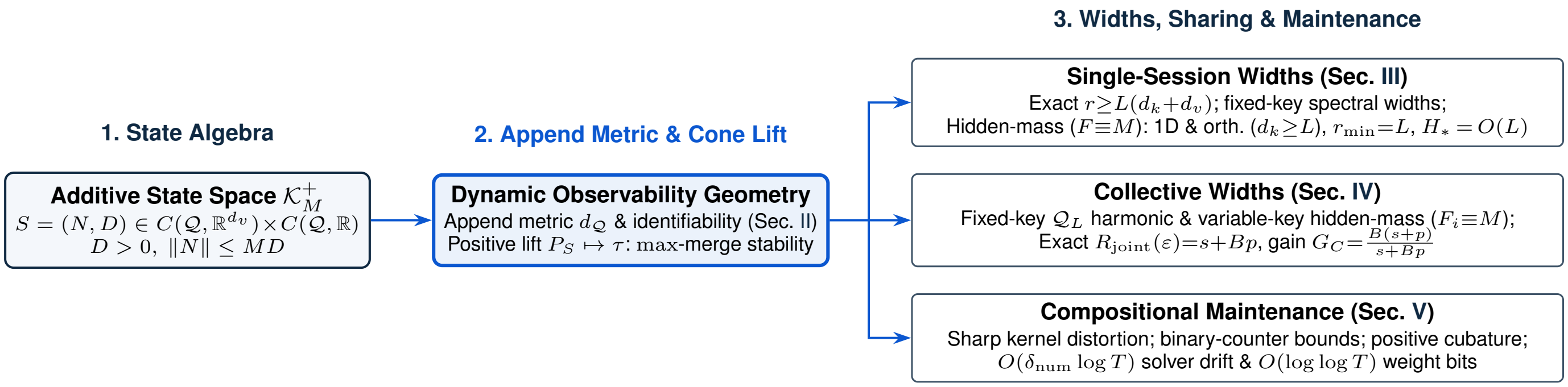

\parhead{Primary results versus supporting framework.}
Fig.~\ref{fig:overview} summarizes how the append-observational metric $\dQ$ links our results:
\begin{itemize}
\item \textbf{Primary results.} (i)~Dynamic equivalence and operational metric characterization on $\K$, distinguishing physical, weighted-atomic, and relaxed continuations (Section~\ref{sec:algebra}) and establishing quantitative necessity and sufficiency linking $\dQ$ to normalized-output and relative-mass preservation (Propositions~\ref{prop:coreset}--\ref{prop:converse}); (ii)~constant-output ($F \equiv M$) variable-key hidden-mass width obstructions proving $r_{\min} = L$ both in one dimension (Theorem~\ref{thm:w4}) and, without Vandermonde conditioning decay, along orthogonal axes in $d_k \ge L$ at an $L$-independent macroscopic threshold $\e_{\mathrm{orth}} = M\tanh(Q_{\max}a/4)$ (Theorem~\ref{thm:w4orth}); (iii)~physical-continuation survival showing these obstructions persist under repeated permitted unit-weight appends up to a factor of $8/9$, within a linear suffix horizon $H_* = O(L)$ in the orthogonal regime (Corollary~\ref{cor:w4phys}, Theorem~\ref{thm:w4orth}(2)); and (iv)~collective multi-session width separation under variable-key hidden historical mass ($F_{S_i} \equiv M$), proving exact matching dimensions $R_{\joint}(\e) = s + Bp$ and $G_C(\e) = B(s+p)/(s+Bp)$ on arbitrary compact query domains $\Q$ containing orthogonal probes (Theorem~\ref{thm:c3}).
\item \textbf{Supporting mathematical results.} Positive-cone merge stability via Thompson's metric and common positive-gauge invariance (Section~\ref{sec:lift}); generic exact continuous-dimension and fixed-key finite-query width references (Sections~\ref{sec:t2}--\ref{sec:fixedkey}); matching collective dimensions for a fixed-key circular harmonic family with spectral truncation on $\Q_L$ (Section~\ref{sec:c2}); kernel distortion, conditional binary-counter accounting, exact-arithmetic positive moment-preserving maintenance, and extensions to approximate numerical reductions ($O(\delta_{\mathrm{num}}\log T)$ dyadic drift) and quantized positive weights ($O(\log\log T)$ shared-exponent bit-width; Section~\ref{sec:maintenance}, Corollaries~\ref{cor:numdrift}--\ref{cor:bitwidth}); and an audit transferring established query-oblivious coreset guarantees to $(\dQ,\tau)$ (Corollary~\ref{cor:coreset_inst}).
\end{itemize}

\parhead{Scope and attribution.}
All results address bounded single-head softmax attention under fixed visibility, a static compact query domain $\Q$, and open-loop common continuations; our representation lower bounds quantify continuous real-coordinate dimensions, while Section~\ref{sec:p1b} bridges exact-arithmetic maintenance to numerical solver residuals and $O(\log\log T)$ weight bit-width (excluding key--value coordinates), stopping short of finite-bit key--value lower bounds, hardware speedups, or closed-loop LLM benchmarks. Prior mathematical and systems ingredients are explicitly attributed where used, including state-pair merging~\cite{cascade,flashinfer,hydragen}, Thompson's metric~\cite{thompson}, Brouwer's invariance of domain~\cite{brouwer}, Borsuk--Ulam~\cite{borsuk}, continuous nonlinear and ellipsoid widths~\cite{devore}, Fourier--Bessel expansions~\cite{dlmf}, sharp Hilbert--TV bounds~\cite{cohenfausti}, Carath\'eodory--Tchakaloff cubature~\cite{bayer,piazzon2017}, Bentley--Saxe dynamization~\cite{bentley}, simultaneous $(N,D)$ control in attention coresets~\cite{liberty,cohen}, and the $\tanh(\delta/4)$ log-mass sensitivity envelope~\cite{yang2026}; Section~\ref{sec:scope} and Table~\ref{tab:related} provide a detailed comparison. Core notation appears in Table~\ref{tab:notation}.

\begin{table}[!t]
\centering
\caption{Core Notation}
\label{tab:notation}
\setlength{\tabcolsep}{3.0pt}
\renewcommand{\arraystretch}{1.10}
\begin{tabular}{@{}L{30.5mm} L{\dimexpr\columnwidth-30.5mm-2\tabcolsep\relax}@{}}
\toprule
\textbf{Symbol} & \textbf{Meaning} \\
\midrule
$\Q,\ \Q_L$ & Compact query domain $\Q\subset\R^{d_k}$; finite circular grid of $L$ points \\
$M,\ \norm{\cdot},\ \norm{\cdot}_*$ & Value bound $M>0$; norm on $\R^{d_v}$ and its dual norm \\
$\K$ & Continuous pairs $(N,D)$ on $\Q$ with $D>0$, $\norm{N}\le MD$ \\
$F_S,\ H_{t,g}(S;q)$ & Normalized output $N/D$; suffix observation $(N{+}tg)/(D{+}t)$ \\
$\dapp,\ \dQ,\ \dQphys$ & Per-query and uniform relaxed and physical append distances \\
$P_S(q,u),\ \tau$ & Positive-cone lift $D{+}\ip{u}{N}/(2M)$ and Thompson metric \\
$w_r^{\mathrm{app}}(\F),\ r_{\min}(\e)$ & Continuous nonlinear width in $\dQ$; minimum achievable dim. \\
$w_R^{(B)},\ R_{\joint},\ G_C$ & Collective width across $B$ sessions, joint dim., gain $R_{\ind}/R_{\joint}$ \\
$\mu,\ \Psi(\mu),\ S(\mu)$ & Weighted atomic measure, feature moment vector, induced state \\
\bottomrule
\end{tabular}
\end{table}

\section{State Algebra, Continuation Models, and Metrics}
\label{sec:algebra}

\subsection{Admissible States and Continuation Models}
\label{sec:states}

Fix a nonempty compact query domain $\Q \subset \R^{d_k}$, a norm $\norm{\cdot}$ on $\R^{d_v}$ with dual norm $\norm{\cdot}_*$, and a value bound $M > 0$. Define the admissible state space explicitly as a set of continuous function pairs:
\begin{equation}\label{eq:cone}
\begin{aligned}
  \K = \bigl\{&(N,D) \in C(\Q,\R^{d_v}) \times C(\Q,\R) :\\
  &D(q) > 0,\ \norm{N(q)} \le M D(q)\ \forall q \in \Q\bigr\}.
\end{aligned}
\end{equation}
Continuity on compact $\Q$ guarantees $\min_{q \in \Q} D(q) > 0$ and $\max_{q \in \Q} D(q) < \infty$, so $F_S = N/D \in C(\Q,\R^{d_v})$ with $\norm{F_S(q)} \le M$, and all log-ratios below are finite. Without continuity, strictly positive functions on a compact domain can approach zero arbitrarily closely and yield unbounded log-ratios.

Every finite positive weighted atomic measure $\mu = \sum_{j=1}^L a_j \delta_{(k_j,v_j)}$ ($a_j > 0$, $\norm{v_j} \le M$) induces $S(\mu) = (N_\mu, D_\mu) \in \K$ via $N_\mu(q) = \sum_j a_j v_j e^{q^\top k_j}$ and $D_\mu(q) = \sum_j a_j e^{q^\top k_j}$. States in $\K$ are identified by their functions on $\Q$, not by any specific atomic representation.

Define composition by pointwise addition:
\begin{equation}\label{eq:add}
  S_1 + S_2 = (N_1 + N_2,\, D_1 + D_2).
\end{equation}
Addition is associative, commutative, and closed on $\K$ (since $\norm{N_1+N_2} \le M(D_1+D_2)$). Because $\K$ contains no additive identity, $(\K, +)$ is a commutative \emph{semigroup}. When an empty history is needed, we adjoin $0 = (0,0)$ to form the commutative \emph{monoid} $\K \cup \{0\}$; we never assign $N/D$ to $0$, and its observations are defined only for strictly positive appended mass. The finite Thompson metric $\tau$ in Section~\ref{sec:lift} remains restricted strictly to $\K$.

\parhead{Continuation models.}
We distinguish three nested common-continuation models at each query $q \in \Q$ (Table~\ref{tab:continuations}):
\begin{enumerate}
\item \emph{Physical token append:} merging a finite sequence of ordinary unit-weight KV pairs $(k_j, v_j)$ from a permitted key domain with $\norm{v_j} \le M$, adding mass $t(q) = \sum_j e^{q^\top k_j}$ and numerator $\sum_j v_j e^{q^\top k_j}$.
\item \emph{Weighted atomic append:} merging finitely many KV pairs with arbitrary positive weights $a_j > 0$ and $\norm{v_j} \le M$.
\item \emph{Relaxed observational append:} merging an arbitrary nonnegative mass $t \ge 0$ and bounded average value $g \in \R^{d_v}$ ($\norm{g} \le M$) independently at each query $q$, producing
\begin{equation}\label{eq:obs}
  H_{t,g}(S; q) = \frac{N(q) + t g}{D(q) + t}.
\end{equation}
\end{enumerate}
Every physical or weighted-atomic append induces an admissible pair $(t(q), g(q))$ with $t(q) > 0$ and $\norm{g(q)} \le M$ at each $q \in \Q$. The relaxed model therefore upper-bounds worst-case discrepancies under ordinary physical appends, whereas a relaxed maximizer $t^*$ (such as $t^* = \sqrt{D\hat D}$) need not be realizable by a fixed-length physical sequence under bounded keys (addressed in Corollary~\ref{cor:w4phys}).

\begin{table}[!t]
\centering
\caption{Hierarchy of Common-Continuation Models at Query $q\in\Q$}
\label{tab:continuations}
\setlength{\tabcolsep}{2.8pt}
\renewcommand{\arraystretch}{1.12}
\begin{tabular}{@{}L{22mm} L{\dimexpr\columnwidth-22mm-2\tabcolsep\relax}@{}}
\toprule
\textbf{Model} & \textbf{Permitted appended suffix} \\
\midrule
Physical token & Unit-weight KV pairs $(k_j,v_j)$ with exponential weights $e^{q^\top k_j}$ and $\norm{v_j}\le M$. \\
Weighted atomic & KV pairs with arbitrary positive weights $a_j>0$ and $\norm{v_j}\le M$. \\
Relaxed observational & Arbitrary mass $t\ge 0$ and bounded value $\norm{g}\le M$ per query; upper-bounds physical appends. \\
\bottomrule
\end{tabular}
\end{table}

\begin{theorem}[Dynamic identifiability]\label{thm:t1}
For $S = (N,D)$ and $\hat S = (\hat N, \hat D)$ in $\K$, the following are equivalent:
\begin{enumerate}
\item $N(q) = \hat N(q)$ and $D(q) = \hat D(q)$ for every $q \in \Q$.
\item $H_{t,g}(S;q) = H_{t,g}(\hat S;q)$ for every $q \in \Q$, $t \ge 0$, and $\norm{g} \le M$ (relaxed observational appends).
\item $F_{S+X}(q) = F_{\hat S+X}(q)$ for every $q \in \Q$ and every physical single-token append $X = (v e^{q^\top k_0}, e^{q^\top k_0})$ drawn from a fixed permitted key $k_0 \in \R^{d_k}$ and two distinct permitted values $v_1 \ne v_2$ with $\norm{v_1},\norm{v_2} \le M$.
\end{enumerate}
\end{theorem}
\begin{proof}
$(1)\Rightarrow(2)\Rightarrow(3)$ is immediate from \eqref{eq:obs}, since a single physical token at $(k_0, v)$ induces $t_0(q) = e^{q^\top k_0} > 0$ and $g = v$. For $(3)\Rightarrow(1)$, fix $q \in \Q$ and subtract the equal observations at $v_1, v_2$:
\[
  \frac{t_0 (v_1 - v_2)}{D(q) + t_0}
  = H_{t_0, v_1}(S;q) - H_{t_0, v_2}(S;q)
  = \frac{t_0 (v_1 - v_2)}{\hat D(q) + t_0}.
\]
Because $t_0 > 0$ and $v_1 \ne v_2$, we obtain $D(q) = \hat D(q)$, and substituting back into $H_{t_0, v_1}(S;q) = H_{t_0, v_1}(\hat S;q)$ yields $N(q) = \hat N(q)$.
\end{proof}

\subsection{Append-Observational Metric and Stability}
\label{sec:dapp}

\begin{definition}[Relaxed append-observational distance]\label{def:dapp}
For $S, \hat S \in \K$ and $q \in \Q$, define the relaxed, adversarial common-continuation distances
\begin{align}
  \dapp(S,\hat S; q) &= \sup_{t \ge 0,\,\norm{g} \le M} \norm{H_{t,g}(S;q) - H_{t,g}(\hat S;q)}, \label{eq:dapp}\\
  \dQ(S,\hat S) &= \sup_{q \in \Q} \dapp(S,\hat S; q). \label{eq:dQ}
\end{align}
\end{definition}

\begin{proposition}[Metric properties of $\dQ$]\label{prop:metric}
For each $q \in \Q$, $\dapp(\cdot,\cdot;q)$ is a pseudometric on $\K$ (and a metric on evaluated pairs $(N(q), D(q))$), and $\dQ$ is a bounded metric on $\K$ taking values in $[0, 2M]$. Extending $H_{t,g}(0;q) = g$ for $t > 0$ and taking the supremum over $t > 0$ extends $\dQ$ to a metric on the monoid $\K \cup \{0\}$.
\end{proposition}
\begin{proof}
For all $S,\hat S \in \K$, $t \ge 0$, and $\norm{g} \le M$, $\norm{H_{t,g}(S;q)} \le (\norm{N(q)} + t\norm{g})/(D(q)+t) \le M$, so $0 \le \dapp(S,\hat S;q) \le 2M$ and $0 \le \dQ(S,\hat S) \le 2M$. Symmetry, the triangle inequality, and $\dQ(S,S) = 0$ follow from the norm axioms and subadditivity of the supremum. If $\dQ(S,\hat S) = 0$, then $H_{t,g}(S;q) = H_{t,g}(\hat S;q)$ for all $q \in \Q$, $t \ge 0$, $\norm{g} \le M$, so Theorem~\ref{thm:t1} gives $S = \hat S$.

For the monoid extension, continuity of $t \mapsto H_{t,g}(S;q)$ at $t=0$ leaves $\dQ$ unchanged on $\K$ when restricted to $t>0$. If $S \in \K$ and $\hat S = 0$, choosing $t > 0$ and $\norm{g} = M$ opposite to $F_S(q)$ (or arbitrary if $F_S(q)=0$) yields $\norm{H_{t,g}(S;q) - g} = \frac{D(q)}{D(q)+t}\norm{F_S(q) - g} > 0$, so $\dQ(S,0) > 0$.
\end{proof}

\begin{lemma}[Historical mass sensitivity]\label{lem:mass}
At a fixed query $q \in \Q$, let $\delta = |\log(\hat D/D)|$, $\alpha = D/(D+t)$, and $\hat\alpha = \hat D/(\hat D+t)$. Then
\begin{equation}\label{eq:mass}
  \sup_{t \ge 0} |\alpha - \hat\alpha| = \tanh(\delta / 4),
\end{equation}
attained uniquely at $t^* = \sqrt{D \hat D}$ when $D \ne \hat D$ (and at every $t \ge 0$ when $D = \hat D$).
\end{lemma}
\begin{proof}
By symmetry assume $r = \hat D/D \ge 1$ and set $x = t/D \ge 0$, giving $|\alpha - \hat\alpha| = (r-1)x / [(1+x)(r+x)]$. If $r=1$ both sides vanish. For $r>1$, the derivative in $x$ has the sign of $r - x^2$, so the unique maximizer is $x = \sqrt{r}$ ($t^* = \sqrt{D\hat D}$), yielding $(\sqrt{r}-1)/(\sqrt{r}+1) = \tanh(\tfrac{1}{4}\log r) = \tanh(\delta/4)$.
\end{proof}

\begin{proposition}[Observational characterization]\label{prop:obs}
At a fixed query $q \in \Q$, let $F = N/D$, $\hat F = \hat N/\hat D$, $e = \norm{F - \hat F}$, and $\delta = |\log(\hat D/D)|$. Then
\begin{align}
  \dapp(S,\hat S; q) &= \sup_{t \ge 0}\bigl(\|\alpha F - \hat\alpha \hat F\| + M|\alpha - \hat\alpha|\bigr), \label{eq:exactobs}\\
  \max\{e,\, M\tanh(\delta/4)\} &\le \dapp(S,\hat S; q) \nonumber\\
  &\le e + 2M\tanh(\delta/4). \label{eq:bounds}
\end{align}
In particular, $\dapp(S,\hat S;q)$ is bi-Lipschitz equivalent to $e + M\tanh(\delta/4)$ with constants $1/2$ and $2$.
\end{proposition}
\begin{proof}
Since $H_{t,g}(S;q) - H_{t,g}(\hat S;q) = (\alpha F - \hat\alpha \hat F) + (\hat\alpha - \alpha) g$, the triangle inequality gives $\norm{H_{t,g}(S;q) - H_{t,g}(\hat S;q)} \le \norm{\alpha F - \hat\alpha \hat F} + M|\alpha - \hat\alpha|$. If $\hat\alpha - \alpha = 0$ (i.e., $t = 0$ or $D(q) = \hat D(q)$), the term $(\hat\alpha - \alpha)g$ vanishes and every admissible $g$ ($\norm{g} \le M$) yields the same value $\norm{\alpha F - \hat\alpha \hat F}$. If $\hat\alpha - \alpha \ne 0$, equality is attained by taking $g = M\operatorname{sgn}(\hat\alpha - \alpha)\,w$, where $w = (\alpha F - \hat\alpha \hat F)/\norm{\alpha F - \hat\alpha \hat F}$ when $\alpha F \ne \hat\alpha \hat F$ and $w$ is any unit vector ($\norm{w} = 1$) when $\alpha F = \hat\alpha \hat F$. This establishes \eqref{eq:exactobs}. Setting $t=0$ gives $\dapp \ge e$; dropping $\norm{\alpha F - \hat\alpha \hat F}$ and applying Lemma~\ref{lem:mass} gives $\dapp \ge M\tanh(\delta/4)$. Writing $\alpha F - \hat\alpha \hat F = \alpha(F-\hat F) + (\alpha - \hat\alpha)\hat F$ with $\alpha \le 1$ and $\norm{\hat F} \le M$ yields the upper bound in \eqref{eq:bounds}. Bi-Lipschitz equivalence follows from $\max(u,v) \ge (u+v)/2$.
\end{proof}

Because $\tanh(\delta/4)$ saturates at $1$, $\dQ$ is locally---not globally---equivalent to an unbounded log-mass norm. Proposition~\ref{prop:obs} also confirms identity of indiscernibles directly: $\dQ(S,\hat S) = 0 \implies e = \delta = 0$ on $\Q$, hence $F = \hat F$, $D = \hat D$, and $N = \hat N$.

\begin{theorem}[Common-merge stability and subadditivity]\label{thm:m1}
For all $S, \hat S, X, S_1, \hat S_1, S_2, \hat S_2 \in \K$,
\begin{align}
  \dQ(S + X, \hat S + X) &\le \dQ(S, \hat S), \label{eq:appendcontract}\\
  \dQ(S_1 + S_2, \hat S_1 + \hat S_2) &\le \dQ(S_1, \hat S_1) + \dQ(S_2, \hat S_2). \label{eq:subadd}
\end{align}
\end{theorem}
\begin{proof}
At any $q \in \Q$, composing $S$ with $X$ followed by $(t,g)$ yields $H_{t,g}(S+X; q) = H_{t', g'}(S; q)$ where $t' = D_X(q) + t > 0$ and $g' = (N_X(q) + tg)/(D_X(q) + t)$ obeys $\norm{g'} \le M$, giving \eqref{eq:appendcontract}. Equation~\eqref{eq:subadd} follows via the triangle inequality through $\hat S_1 + S_2$ and \eqref{eq:appendcontract}.
\end{proof}

\subsection{Positive-Cone Lift and Maximum-Error Merge Law}
\label{sec:lift}

Subadditivity in \eqref{eq:subadd} allows errors to grow linearly with the number of merged blocks. To eliminate merge-count accumulation, let $B_* = \{u \in \R^{d_v} : \norm{u}_* \le 1\}$ be the compact dual unit ball and define the linear positive-cone lift
\begin{equation}\label{eq:lift}
  P_S(q, u) = D(q) + \frac{\ip{u}{N(q)}}{2M},\qquad (q,u) \in \Q \times B_*.
\end{equation}
By \eqref{eq:cone} and $|\ip{u}{N(q)}| \le M D(q)$, every $S \in \K$ satisfies
\begin{equation}\label{eq:liftbounds}
  \frac{1}{2} D(q) \le P_S(q,u) \le \frac{3}{2} D(q) \qquad \forall (q,u) \in \Q \times B_*.
\end{equation}
Thus $P_S \in C(\Q \times B_*, (0,\infty))$. The lift is injective because $P_S(q,0) = D(q)$ and $2M(P_S(q,u) - P_S(q,0)) = \ip{u}{N(q)}$ for all $u \in B_*$ determines $N(q)$. Pulling back Thompson's part metric~\cite{thompson} on the cone of positive continuous functions yields
\begin{equation}\label{eq:tau}
  \tau(S, \hat S) = \sup_{(q,u) \in \Q \times B_*} \left|\log \frac{P_{\hat S}(q,u)}{P_S(q,u)}\right|.
\end{equation}
Continuity on compact $\Q \times B_*$ and \eqref{eq:liftbounds} ensure $\tau(S,\hat S) < \infty$; symmetry, the triangle inequality, and $\tau(S,\hat S) = 0 \iff S = \hat S$ follow from $|\log(\cdot)|$ and injectivity of $P_S$. Thus $\tau$ is a finite metric on $\K$.

\begin{theorem}[Maximum-error merge law]\label{thm:m3}
For any finite collections $S_1,\ldots,S_m \in \K$ and $\hat S_1,\ldots,\hat S_m \in \K$,
\begin{equation}\label{eq:m3}
  \tau\!\left(\sum_{j=1}^m S_j,\ \sum_{j=1}^m \hat S_j\right)
  \le \max_{1 \le j \le m} \tau(S_j, \hat S_j).
\end{equation}
\end{theorem}
\begin{proof}
Let $b = \max_{1\le j\le m} \tau(S_j, \hat S_j) < \infty$. For every $j$ and $(q,u) \in \Q \times B_*$, $e^{-b} P_{S_j}(q,u) \le P_{\hat S_j}(q,u) \le e^b P_{S_j}(q,u)$. Summing over $j$ and using linearity of $S \mapsto P_S$ preserves both inequalities; dividing by $P_{\sum_j S_j}(q,u) > 0$, taking logarithms, and taking suprema over $\Q \times B_*$ yields \eqref{eq:m3}.
\end{proof}

Theorem~\ref{thm:m3} prevents error growth across parallel merges of independently compressed states; distortion from \emph{repeated recompressions} is treated in Section~\ref{sec:maintenance}.

\begin{theorem}[Observable control and reverse bound]\label{thm:m4}
For all $S, \hat S \in \K$,
\begin{equation}\label{eq:m4}
  \dQ(S, \hat S) \le \min\bigl\{2M,\, 4M\tanh(\tau(S,\hat S)/2)\bigr\},
\end{equation}
and conversely,
\begin{equation}\label{eq:reverse}
  \tau(S, \hat S) \le \sup_{q \in \Q}\!\left(\left|\log\frac{\hat D(q)}{D(q)}\right| + \frac{\norm{F_S(q) - F_{\hat S}(q)}}{M}\right).
\end{equation}
\end{theorem}
\begin{proof}
Let $\tau = \tau(S,\hat S)$, fix $q \in \Q$, $t \ge 0$, $\norm{g} \le M$, and $u \in B_*$. Define the suffix-extended positive lift
\begin{align*}
  P_S^{\mathrm{ext}}(q,u) &= P_S(q,u) + c(u),\\
  c(u) &= t\!\left(1 + \frac{\ip{u}{g}}{2M}\right) \ge \frac{t}{2} \ge 0,
\end{align*}
and $P_{\hat S}^{\mathrm{ext}}(q,u) = P_{\hat S}(q,u) + c(u)$. Because $e^{-\tau} \le 1 \le e^{\tau}$ and $c(u) \ge 0$, adding $c(u)$ preserves $e^{-\tau} P_S^{\mathrm{ext}}(q,u) \le P_{\hat S}^{\mathrm{ext}}(q,u) \le e^{\tau} P_S^{\mathrm{ext}}(q,u)$. Put $P^\pm = P_S^{\mathrm{ext}}(q, \pm u)$ and $\hat P^\pm = P_{\hat S}^{\mathrm{ext}}(q, \pm u)$, so
\begin{align*}
  p &= \frac{P^+}{P^+ + P^-} = \frac{1}{2} + \frac{\ip{u}{H_{t,g}(S;q)}}{4M},\\
  \hat p &= \frac{\hat P^+}{\hat P^+ + \hat P^-} = \frac{1}{2} + \frac{\ip{u}{H_{t,g}(\hat S;q)}}{4M}.
\end{align*}
With $\sigma(z) = (1+e^{-z})^{-1}$, we have $p = \sigma(x)$ and $\hat p = \sigma(\hat x)$ where $x = \log(P^+/P^-)$ and $\hat x = \log(\hat P^+/\hat P^-)$ satisfy $|x - \hat x| \le |\log(\hat P^+/P^+)| + |\log(\hat P^-/P^-)| \le 2\tau$. Consider $h(y) = \sigma(y+b) - \sigma(y)$ for $b \ge 0$. When $b = 0$, $h(y) \equiv 0 = \tanh(0)$ identically. When $b > 0$, $h'(y) = \sigma'(y+b) - \sigma'(y)$; because $\sigma'$ is even and strictly decreasing on $[0,\infty)$, $h'(y) > 0$ for $y < -b/2$ and $h'(y) < 0$ for $y > -b/2$, so $h$ attains its unique maximum at $y = -b/2$, where $h(-b/2) = \tanh(b/4)$. Taking $b = 2\tau \ge 0$ gives
\begin{align*}
  \bigl|\ip{u}{H_{t,g}(S;q) - H_{t,g}(\hat S;q)}\bigr|
  &= 4M|p - \hat p| \\
  &\le 4M\tanh(\tau/2).
\end{align*}
Taking suprema over $u \in B_*$, $t \ge 0$, $\norm{g} \le M$, $q \in \Q$ and noting $\dQ \le 2M$ proves \eqref{eq:m4}.

For \eqref{eq:reverse}, factor $P_S(q,u) = D(q)\bigl(1 + \ip{u}{F_S(q)}/(2M)\bigr)$. Since $1 + \ip{u}{F_S(q)}/(2M) \in [1/2, 3/2]$ and $\log$ is $2$-Lipschitz on $[1/2, 3/2]$,
\[
  \left|\log\frac{P_{\hat S}(q,u)}{P_S(q,u)}\right|
  \le \left|\log\frac{\hat D(q)}{D(q)}\right| + 2\cdot\frac{\left|\ip{u}{F_{\hat S}(q) - F_S(q)}\right|}{2M},
\]
yielding \eqref{eq:reverse} upon taking suprema over $u \in B_*$ and $q \in \Q$.
\end{proof}

\begin{remark}[Topological equivalence]\label{rem:topology}
By Proposition~\ref{prop:obs} and Theorem~\ref{thm:m4}, $\dQ$ and $\tau$ induce the same topology on $\K$: $\dQ(S_n, S) \to 0$ implies uniform convergence $F_{S_n} \to F_S$ and $\log D_{S_n} \to \log D_S$, hence $\tau(S_n, S) \to 0$ by \eqref{eq:reverse}; the converse holds by \eqref{eq:m4}.
\end{remark}

\begin{proposition}[Common positive-gauge invariance]\label{prop:gauge}
For any continuous positive function $c \in C(\Q, (0,\infty))$ and any $S = (N, D) \in \K$, define the rescaled state $cS = (cN, cD) \in \K$. Then for all $S, \hat S \in \K$, $q \in \Q$, $t \ge 0$, and $\norm{g} \le M$,
\begin{equation}\label{eq:gauge_obs}
  H_{t,g}(cS; q) = H_{t/c(q),\, g}(S; q),
\end{equation}
and consequently both relaxed metrics are invariant under common positive rescaling:
\begin{equation}\label{eq:gauge_inv}
  \dQ(cS, c\hat S) = \dQ(S, \hat S),\qquad \tau(cS, c\hat S) = \tau(S, \hat S).
\end{equation}
\end{proposition}
\begin{proof}
Since $c$ is continuous and strictly positive on compact $\Q$, $cS \in \K$, and dividing numerator and denominator of $(c(q)N(q)+tg)/(c(q)D(q)+t)$ by $c(q) > 0$ yields \eqref{eq:gauge_obs}. Because $t \mapsto t/c(q)$ maps $[0,\infty)$ bijectively onto $[0,\infty)$, taking suprema over $t \ge 0$, $\norm{g} \le M$, and $q \in \Q$ gives $\dQ(cS, c\hat S) = \dQ(S, \hat S)$. Similarly, linearity of the positive lift \eqref{eq:lift} gives $P_{cS}(q,u) = c(q)P_S(q,u)$, so $P_{c\hat S}(q,u)/P_{cS}(q,u) = P_{\hat S}(q,u)/P_S(q,u)$ on $\Q \times B_*$, proving $\tau(cS, c\hat S) = \tau(S, \hat S)$.
\end{proof}

Three operational distinctions follow immediately from Proposition~\ref{prop:gauge}: (1)~consistently centering keys of both $S$ and $\hat S$ by $-\bar k$ multiplies both states by $c(q) = e^{-q^\top\bar k} > 0$ and leaves $\dQ$ and $\tau$ unchanged; (2)~before \emph{composing} a centered state $cS$ with future uncentered physical KV tokens $X$, one must restore $S$ to its original key gauge (or transform future tokens by the same shift $-\bar k$), since $cS + X \ne c(S + X)$ whenever $c \not\equiv 1$; and (3)~the restricted integer-mass physical-continuation metric $\dQphys$ in Section~\ref{sec:w4} generally lacks this invariance because rescaling $S \mapsto cS$ alters the realizable integer suffix masses $n/c(q)$ ($n \in \Z_{\ge 0}$).

\section{Dynamic Functional Widths}
\label{sec:width}

\subsection{Continuous Nonlinear Widths and Streaming Complexity}
\label{sec:widthdef}

Adapting the continuous nonlinear widths of DeVore, Howard, and Micchelli~\cite{devore} to $\dQ$, we measure the minimum number of continuous real coordinates required to encode a state family up to uniform error $\e$.

\begin{definition}[Continuous dynamic functional width]\label{def:width}
For a family $\F \subseteq \K$ and integer $r \ge 0$, define
\begin{equation}\label{eq:width}
  w_r^{\mathrm{app}}(\F) = \inf_{E,\, G} \sup_{S \in \F} \dQ(S, G(E(S))),
\end{equation}
where $E: \F \to \R^r$ is continuous in the $\dQ$-topology and $G: \R^r \to \K$ is an arbitrary (possibly discontinuous) decoder (for $r=0$, $\R^0$ is a singleton and $G$ outputs a single constant state). For $\e \ge 0$, define $r_{\min}(\F,\e)$ as the smallest $r \ge 0$ admitting continuous $E: \F \to \R^r$ and arbitrary $G: \R^r \to \K$ with $\sup_{S \in \F} \dQ(S, G(E(S))) \le \e$ (or $\infty$ if none exists).
\end{definition}

The decoder $G$ may use fixed family parameters (such as fixed keys in Section~\ref{sec:fixedkey} or public Fourier bases in Section~\ref{sec:c2}), while $r$ counts variable continuous coordinates. Our lower bounds apply the Borsuk--Ulam collision principle~\cite{borsuk,devore}: if $\varphi: B \to \F$ is continuous on a centrally symmetric norm ball $B \subset \R^m$ ($m \ge 1$) and $E: \F \to \R^r$ ($r < m$) is continuous, then composing $E \circ \varphi$ with the radial antipodal homeomorphism $S^{m-1} \to \partial B$ yields an antipodal pair $z, -z \in \partial B$ with $E(\varphi(z)) = E(\varphi(-z))$; any decoder $G$ incurs error at least $\tfrac{1}{2}\dQ(\varphi(z), \varphi(-z))$ on one of the two inputs.

\parhead{Why encoder continuity?}
Requiring continuity of $E: (\F, \dQ) \to \R^r$ enforces \emph{perturbation stability} of the compressed coordinates, the approximation-theoretic analogue of well-posedness in inverse problems~\cite{devore}: small changes to an attention state in $\dQ$ cannot trigger arbitrarily large jumps in its representation. Without continuity, a space-filling bijection can collapse any $\R^L$ parameter family into $\R^1$ with zero reconstruction error, making real-coordinate dimension vacuous. Continuity is imposed only on $E$ (never on the decoder $G$), and lower bounds on $w_r^{\mathrm{app}}(\F)$ therefore quantify the minimum stable coordinate dimension even when specific constructive algorithms---such as greedy Carath\'eodory--Tchakaloff elimination (Theorem~\ref{thm:p1b}) or discrete cache-eviction heuristics---are themselves discontinuous.

\parhead{Relationship to finite-bit streaming complexity.}
Continuous real-coordinate widths and finite-bit streaming space bounds~\cite{haris,balancekv,tightstream} capture complementary obstructions (Table~\ref{tab:width_vs_bits}). Streaming bounds count retained bits via communication complexity, whereas continuous widths count independent real coordinates via topological obstructions (invariance of domain and Borsuk--Ulam), characterizing robust continuous state geometries and collective dimension laws. A coordinate lower bound $r \ge L$ does \emph{not} imply a byte-level GPU memory lower bound without an additional precision model.

\begin{table}[!t]
\centering
\caption{Complementary Roles of Continuous Nonlinear Widths and Finite-Bit Streaming Space Complexity}
\label{tab:width_vs_bits}
\setlength{\tabcolsep}{2.8pt}
\renewcommand{\arraystretch}{1.12}
\begin{tabular}{@{}L{\dimexpr0.5\columnwidth-\tabcolsep\relax} L{\dimexpr0.5\columnwidth-\tabcolsep\relax}@{}}
\toprule
\textbf{Continuous representation width} & \textbf{Streaming space complexity} \\
\midrule
Counts independent real coordinates & Counts retained bits \\
Geometric / topological obstruction & Communication / info.-theoretic obstruction \\
Characterizes robust continuous states & Characterizes digital memory requirements \\
Identifies structural collective gains & Establishes finite-memory impossibility \\
No direct byte bound without precision & No direct continuous-geometry characterization \\
\bottomrule
\end{tabular}
\end{table}

\subsection{Generic Exact Continuous State Dimension}
\label{sec:t2}

\begin{theorem}[Generic exact dynamic state dimension]\label{thm:t2}
Suppose compact $\Q \subset \R^{d_k}$ contains a nonempty open set $U \subset \R^{d_k}$. Consider the family $\F_L \subseteq \K$ of unit-weight histories with $L$ pairwise distinct keys $k_1,\ldots,k_L \in \R^{d_k}$ and independently variable values $v_1,\ldots,v_L \in \R^{d_v}$ satisfying $\norm{v_j} < M$. Any continuous, exact dynamically sufficient encoder $E: \F_L \to \R^r$ (admitting $G: \R^r \to \K$ with $\dQ(S, G(E(S))) = 0$ for all $S \in \F_L$) must satisfy
\begin{equation}\label{eq:t2}
  r \ge L(d_k + d_v).
\end{equation}
\end{theorem}
\begin{proof}
\emph{Step 1 (Linear independence of exponentials).} Let $c_1,\ldots,c_m \in \R$ and let $a_1,\ldots,a_m \in \R^{d_k}$ be pairwise distinct, with $\sum_{j=1}^m c_j e^{q^\top a_j} = 0$ on $U$. Because $\bigcup_{j < j'} \{w \in \R^{d_k} : w^\top (a_j - a_{j'}) = 0\}$ has empty interior, pick $w \in \R^{d_k}$ so that $\lambda_j = w^\top a_j$ are pairwise distinct, and pick $q_0 \in U$, $\eta > 0$ so that $q_0 + t w \in U$ for $|t| < \eta$. Differentiating $g(t) = \sum_{j=1}^m (c_j e^{q_0^\top a_j}) e^{t \lambda_j} \equiv 0$ at $t=0$ for orders $0,\ldots,m-1$ gives a homogeneous Vandermonde system with distinct nodes $\lambda_j$, forcing $c_j = 0$ for all $j$.

\emph{Step 2 (Recovery up to permutation).} If two unit-weight histories $\theta = (k_1,\ldots,k_L, v_1,\ldots,v_L)$ and $\hat\theta = (\hat k_1,\ldots,\hat k_L, \hat v_1,\ldots,\hat v_L)$ with distinct keys satisfy $S(\theta) = S(\hat\theta)$, Theorem~\ref{thm:t1} gives $D_\theta = D_{\hat\theta}$ and $N_\theta = N_{\hat\theta}$ on $U$. Applying Step~1 to $D_\theta - D_{\hat\theta} = 0$ yields a unique permutation $\pi$ of $\{1,\ldots,L\}$ with $\hat k_j = k_{\pi(j)}$; applying Step~1 coordinatewise to $N_\theta - N_{\hat\theta} = 0$ forces $\hat v_j = v_{\pi(j)}$.

\emph{Step 3 (Open consistently ordered domain).} Pick a base tuple $\theta^0$ with $k_{1,1}^0 < \cdots < k_{L,1}^0$ and $\norm{v_j^0} < M$, and let $\Omega \subset (\R^{d_k})^L \times (\R^{d_v})^L$ be an open neighborhood of $\theta^0$ on which $k_{1,1} < \cdots < k_{L,1}$ and $\norm{v_j} < M$. Because no nontrivial permutation preserves $\Omega$, Step~2 makes $\Phi: \Omega \to \F_L$ injective. Compactness of $\Q$ and uniform positivity of $D_\theta$ on compact subsets of $\Omega$ make $\Phi$ continuous into $(\K,\tau)$, hence into $(\K,\dQ)$ by Theorem~\ref{thm:m4}.

\emph{Step 4 (Invariance of domain).} By Proposition~\ref{prop:metric}, exactness makes $E$ injective on $\F_L$, so $E \circ \Phi: \Omega \to \R^r$ is a continuous injection from a nonempty open subset of $\R^{L(d_k+d_v)}$ into $\R^r$. If $r < L(d_k+d_v)$, composing with the coordinate inclusion $\R^r \hookrightarrow \R^{L(d_k+d_v)}$ ($y \mapsto (y, 0)$) maps an open set injectively and continuously into a proper subspace of empty interior, contradicting Brouwer's invariance of domain~\cite{brouwer}. Thus $r \ge L(d_k+d_v)$.
\end{proof}

\subsection{Fixed-Key Widths and Finite-Query Spectral Reference}
\label{sec:fixedkey}

As a linear, fixed-key reference baseline contrasting with the nonlinear variable-key hidden-mass obstructions in Section~\ref{sec:w4}, we first characterize continuous widths when keys are fixed and only values vary, culminating in the finite-query ellipsoid-width benchmark of Theorem~\ref{thm:w3}. Fix keys $k_1,\ldots,k_L \in \R^{d_k}$ and scalar values $v = (v_1,\ldots,v_L)^\top \in \R^L$ ($d_v = 1$, $|v_j| \le M$), defining $S_v = (N_v, D_{\mathrm{fix}}) \in \K$ with fixed denominator $D_{\mathrm{fix}}(q) = \sum_{j=1}^L e^{q^\top k_j}$. Given probe queries $\Q_J = \{q_1,\ldots,q_J\} \subseteq \Q$, define the $J \times L$ row-stochastic response matrix
\begin{equation}\label{eq:A}
  A_{ij} = \frac{e^{q_i^\top k_j}}{\sum_{h=1}^{L} e^{q_i^\top k_h}},\qquad
  1 \le i \le J,\ 1 \le j \le L.
\end{equation}
On $\Q_J$, $F_{S_v}(\Q_J) = Av$. Whenever a decoded state $\hat S$ retains the exact denominator $D_{\mathrm{fix}}$, $|H_{t,g}(S_v;q) - H_{t,g}(\hat S;q)| = \frac{D_{\mathrm{fix}}(q)}{D_{\mathrm{fix}}(q)+t}|F_{S_v}(q) - F_{\hat S}(q)|$, maximized at $t=0$; hence $\dapp(S_v, \hat S; q) = |F_{S_v}(q) - F_{\hat S}(q)|$.

\begin{theorem}[Fixed-key observability lower bound]\label{thm:w1}
Let $\F_{\mathrm{fix}} = \{S_v : v \in [-M, M]^L\} \subseteq \K$ for fixed keys $k_1,\ldots,k_L$. If $J = L$ probe queries $q_1,\ldots,q_L \in \Q$ make $A$ in \eqref{eq:A} invertible, then for every $r < L$,
\begin{equation}\label{eq:w1}
  w_r^{\mathrm{app}}(\F_{\mathrm{fix}}) \ge \frac{M}{\norm{A^{-1}}_{\infty \to \infty}}.
\end{equation}
\end{theorem}
\begin{proof}
Because $v \mapsto S_v$ is continuous on $[-M,M]^L$, Borsuk--Ulam gives an antipodal collision $v, -v$ with $\norm{v}_\infty = M$ for any continuous $E: \F_{\mathrm{fix}} \to \R^r$ ($r < L$). Evaluating at $t=0$ on $\Q_L$ gives $\dQ(S_v, S_{-v}) \ge 2\norm{Av}_\infty$, so one input incurs error at least $\norm{Av}_\infty \ge M / \norm{A^{-1}}_{\infty \to \infty}$.
\end{proof}

\begin{corollary}[Selectively addressable histories]\label{cor:w2}
Under Theorem~\ref{thm:w1}, if each key $k_i$ has a probe query $q_i \in \Q$ with $q_i^\top k_i - q_i^\top k_j \ge \Delta$ for all $j \ne i$ and $\rho_0 = (L-1)e^{-\Delta} < 1$, then for all $r < L$,
\begin{equation}\label{eq:w2}
  M \frac{1 - \rho_0}{1 + \rho_0} \le w_r^{\mathrm{app}}(\F_{\mathrm{fix}}) \le M.
\end{equation}
\end{corollary}
\begin{proof}
Here $A_{ii} \ge (1+\rho_0)^{-1}$ and $\sum_{j\ne i} A_{ij} = 1 - A_{ii}$, so $\norm{Ax}_\infty \ge \min_i(2A_{ii}-1)\norm{x}_\infty \ge \frac{1-\rho_0}{1+\rho_0}\norm{x}_\infty$; hence $A$ is invertible and Theorem~\ref{thm:w1} yields the lower bound. Decoding $S_0$ (using the known fixed keys) gives error at most $M$.
\end{proof}

For a finite query set $\Q_J = \{q_1,\ldots,q_J\} \subseteq \Q$, define the finite-query append-observational distances
\begin{align}
  d_{\Q_J, 2}(S, \hat S) &= \left(\sum_{i=1}^J \dapp(S, \hat S; q_i)^2\right)^{1/2}, \label{eq:dQ2}\\
  d_{\Q_J}(S, \hat S) &= \max_{1 \le i \le J} \dapp(S, \hat S; q_i). \label{eq:dQinf}
\end{align}
These are metrics when $\K$ is restricted to the finite domain $\Q = \Q_J$, and observation pseudometrics when $\Q \supsetneq \Q_J$. Let $w_{r,2}^{\mathrm{app}}(\F; \Q_J)$ and $w_{r,\infty}^{\mathrm{app}}(\F; \Q_J)$ denote the corresponding continuous widths (Definition~\ref{def:width} with error and encoder continuity measured in $d_{\Q_J,2}$ and $d_{\Q_J}$, respectively).

\begin{theorem}[Exact finite-query spectral width]\label{thm:w3}
Let $\F_{\mathrm{fix},2} = \{S_v : v \in \R^L,\ \norm{v}_2 \le M\} \subseteq \K$ for fixed keys $k_1,\ldots,k_L$ (either on $\Q = \Q_J$ or on a larger compact $\Q \supseteq \Q_J$ equipped with the pseudometrics \eqref{eq:dQ2}--\eqref{eq:dQinf}). Let $\sigma_1(A) \ge \cdots \ge \sigma_{\min(J,L)}(A) \ge 0$ be the singular values of $A$ in \eqref{eq:A}, with convention $\sigma_k(A) = 0$ for $k > \min(J,L)$. Then for every integer $r \ge 0$,
\begin{equation}\label{eq:w3}
  w_{r,2}^{\mathrm{app}}(\F_{\mathrm{fix},2}; \Q_J) = M \sigma_{r+1}(A),
\end{equation}
and on the same finite query set $\Q_J$,
\begin{equation}\label{eq:w3inf}
  \frac{M \sigma_{r+1}(A)}{\sqrt{J}} \le w_{r,\infty}^{\mathrm{app}}(\F_{\mathrm{fix},2}; \Q_J) \le M \sigma_{r+1}(A).
\end{equation}
\end{theorem}
\begin{proof}
Since $\norm{v}_\infty \le \norm{v}_2 \le M$, $\F_{\mathrm{fix},2} \subseteq \K$. Let $k_A = \rank(A) \le \min(J,L)$ and $r_* = \min\{r, k_A\}$.

\emph{Upper bound.} Write the compact SVD $A = U_{k_A} \Sigma_{k_A} V_{k_A}^\top$ with $\Sigma_{k_A} = \diag(\sigma_1,\ldots,\sigma_{k_A}) \succ 0$, and let $U_{r_*}, \Sigma_{r_*}, V_{r_*}$ denote the top $r_*$ singular factors. If $r = 0$, $E: \F_{\mathrm{fix},2} \to \R^0$ is constant and $G$ returns $S_0$ ($\hat v = 0$), yielding error $\sup_{\norm{v}_2\le M} \norm{Av}_2 = M\sigma_1(A)$. If $r \ge 1$ and $r_* = 0$, set $E \equiv 0 \in \R^r$ and $\hat v = 0$ (error $0 = M\sigma_{r+1}(A)$). Otherwise ($r_* \ge 1$), define $E: \F_{\mathrm{fix},2} \to \R^r$ by
\begin{align*}
  z_* &= \Sigma_{r_*}^{-1} U_{r_*}^\top F_{S_v}(\Q_J) = V_{r_*}^\top v \in \R^{r_*},\\
  E(S_v) &= (z_*, 0_{r - r_*}) \in \R^r.
\end{align*}
Because $E(S_v)$ depends linearly on the observable vector $F_{S_v}(\Q_J)$, it is well-defined and continuous in both the $d_{\Q_J,2}$- and $d_{\Q_J}$-topologies even when $A$ is rank-deficient or $r > k_A$ or $r > L$. Define $G: \R^r \to \K$ by extracting the first $r_*$ coordinates $\tilde z \in \R^{r_*}$, projecting onto the closed radius-$M$ Euclidean ball via $\Pi_M(\tilde z)$, setting $\hat v = V_{r_*} \Pi_M(\tilde z)$ (which obeys $\norm{\hat v}_\infty \le \norm{\hat v}_2 \le M$), and outputting $S_{\hat v} \in \K$ with fixed denominator $D_{\mathrm{fix}}$. Because $\norm{z_*}_2 = \norm{V_{r_*}^\top v}_2 \le M$, $\Pi_M(z_*) = z_*$ on $\F_{\mathrm{fix},2}$, so $A\hat v = U_{r_*}\Sigma_{r_*} V_{r_*}^\top v$ and
\[
  d_{\Q_J,2}(S_v, S_{\hat v}) = \norm{A(v - \hat v)}_2 \le M\sigma_{r_*+1}(A) = M\sigma_{r+1}(A)
\]
(noting that when $r \ge k_A$, $r_* = k_A$ and $A\hat v = Av$, giving error $0 = M\sigma_{r+1}(A)$).

\emph{Lower bound.} If $r \ge k_A$, $\sigma_{r+1}(A) = 0$ and the lower bound $0$ is trivial. If $r < k_A$, then $r + 1 \le k_A$ and $\sigma_{r+1}(A) > 0$. Restrict to the radius-$M$ Euclidean sphere in $\operatorname{span}\{v_1,\ldots,v_{r+1}\}$ (the first $r+1$ right singular vectors). Because $v \mapsto S_v$ is continuous into $(\F_{\mathrm{fix},2}, d_{\Q_J,2})$, Borsuk--Ulam yields an antipodal collision $v, -v$ with $d_{\Q_J,2}(S_v, S_{-v}) \ge 2\norm{Av}_2 \ge 2M\sigma_{r+1}(A)$, giving worst-case error at least $M\sigma_{r+1}(A)$ (for $r=0$, the two antipodal points $\pm M v_1$ trivially collide). Equation~\eqref{eq:w3inf} follows from $J^{-1/2}\norm{x}_2 \le \norm{x}_\infty \le \norm{x}_2$.
\end{proof}

Theorem~\ref{thm:w3} is an attention-specific ellipsoid width~\cite{devore} valid strictly on $\Q_J$; it contrasts with the nonlinear variable-key family next.

\subsection{Variable Keys: Hidden Historical Mass in Relaxed and Physical Models}
\label{sec:w4}

Let $d_k \ge 1$ and $d_v = 1$, and assign every key in a length-$L$ unit-weight history $k = (k_1,\ldots,k_L) \in (\R^{d_k})^L$ the identical scalar value $v_j = M$ ($j = 1,\ldots,L$). Every such key configuration produces $S_k = (M D_k, D_k) \in \K$ with
\[
  F_{S_k}(q) = \frac{M D_k(q)}{D_k(q)} \equiv M \qquad \forall q \in \Q.
\]
Every state in such a constant-value family is \emph{completely indistinguishable under all current normalized attention observations}. Under the relaxed observational append model (Definition~\ref{def:dapp}), however, Proposition~\ref{prop:obs} and Lemma~\ref{lem:mass} yield the exact per-query distance
\begin{equation}\label{eq:keydist}
  \dapp(S_k, S_{\hat k}; q) = 2M \tanh\!\left(\frac{|\log(D_k(q)/D_{\hat k}(q))|}{4}\right),
\end{equation}
attained at $g = -M$ and $t^* = \sqrt{D_k(q) D_{\hat k}(q)}$. We first show that this hidden-mass obstruction already forces full local dimension even in one key dimension ($d_k = 1$; Theorem~\ref{thm:w4} and Corollary~\ref{cor:w4phys}), and then eliminate the resulting 1D Vandermonde conditioning decay in $d_k \ge L$ via an orthogonal-axis construction yielding an $L$-independent macroscopic threshold within a linear physical suffix horizon (Theorem~\ref{thm:w4orth}).

\begin{theorem}[Matching local nonlinear key width in 1D]\label{thm:w4}
Let $d_k = d_v = 1$, fix a bounded key interval $[-K_{\max}, K_{\max}]$ with strictly ordered interior reference keys $-K_{\max} < k_1^0 < \cdots < k_L^0 < K_{\max}$, and assume $\Q \subset \R$ contains $q_i = i h$ for $i = 1,\ldots,L$ and some $h > 0$. Define the log-partition map $\Phi: \R^L \to \R^L$, its Jacobian $J$ at $k^0$, conditioning constant $\mu$, and maximum probe magnitude $Q_{\max}$ by
\begin{equation}\label{eq:w4defs}
\begin{aligned}
  \Phi_i(k) &= \log \sum_{j=1}^{L} e^{q_i k_j},\qquad
  J_{ij} = \frac{q_i e^{q_i k_j^0}}{D_{k^0}(q_i)},\\
  \mu &= \norm{J^{-1}}_{\infty \to \infty}^{-1},\qquad
  Q_{\max} = \max_{1 \le i \le L} |q_i| = L h.
\end{aligned}
\end{equation}
Then $J$ is invertible and $\mu > 0$. Choose $a > 0$ small enough that $C_a = k^0 + [-a, a]^L \subset [-K_{\max}, K_{\max}]^L$ and (when $L \ge 2$) $k_1 < \cdots < k_L$ for all $k \in C_a$, i.e., $a < \min\bigl\{K_{\max} - |k_1^0|,\, K_{\max} - |k_L^0|,\, \tfrac{1}{2}\min_{1\le j<L}(k_{j+1}^0 - k_j^0)\bigr\}$ with the empty-minimum convention $\min\emptyset = +\infty$ when $L = 1$ (where the adjacent-key ordering constraint is vacuous and $r < 1$ reduces to $r = 0$, with antipodal points $k_1^0 \pm a$). Put $\F_a = \{S_k : k \in C_a\} \subseteq \K$. In the relaxed observational append model, if $a \le 2\mu / Q_{\max}^2$, then setting $\e_{\mathrm{hm}} := M \tanh\bigl((2a\mu - \tfrac{1}{2}Q_{\max}^2 a^2)/4\bigr)$, we have for every $r < L$,
\begin{equation}\label{eq:w4lower}
  w_r^{\mathrm{app}}(\F_a)
  \ge \e_{\mathrm{hm}}
  \ge M \tanh\!\left(\frac{a\mu}{4}\right) > 0.
\end{equation}
Conversely, $w_L^{\mathrm{app}}(\F_a) = 0$ via a continuous exact encoder into $\R^L$; thus for every $0 \le \e < \e_{\mathrm{hm}}$,
\begin{equation}\label{eq:w4exact}
  r_{\min}(\F_a, \e) = L.
\end{equation}
\end{theorem}
\begin{proof}
\emph{Invertibility of $J$.} With $z_j = e^{h k_j^0} > 0$ pairwise distinct, $J = \diag\bigl(q_i / D_{k^0}(q_i)\bigr)(z_j^i)_{i,j=1}^L$ is a nonsingular diagonal matrix times a transposed Vandermonde matrix times $\diag(z_1,\ldots,z_L)$, so $\det J \ne 0$ and $\mu > 0$.

\emph{Lower bound.} Because $x \mapsto S_{k^0 + x}$ is continuous on $[-a,a]^L$, Borsuk--Ulam gives an antipodal collision $x, -x$ ($\norm{x}_\infty = a$) for any continuous $E: \F_a \to \R^r$ ($r < L$). Setting $f_i(s) = \Phi_i(k^0 + s x)$ ($s \in [-1,1]$),
\begin{equation}\label{eq:hessian}
  f_i''(s) = D^2\Phi_i(k^0+sx)[x,x] = q_i^2 \Var_{p_i(k^0+sx)}(x_j),
\end{equation}
where $p_i(k)_j = e^{q_i k_j}/D_k(q_i)$. Since $|x_j| \le a$, $0 \le f_i''(s) \le q_i^2 a^2$. By Taylor's theorem with integral remainder,
\begin{align*}
  f_i(1) - f_i(-1) &= 2 f_i'(0) + e_i,\\
  e_i &= \int_0^1 (1-s)\bigl(f_i''(s) - f_i''(-s)\bigr)\,ds.
\end{align*}
Because $f_i''(s), f_i''(-s) \in [0, q_i^2 a^2]$, we have $|f_i''(s) - f_i''(-s)| \le q_i^2 a^2$, yielding $|e_i| \le \tfrac{1}{2} q_i^2 a^2 \le \tfrac{1}{2} Q_{\max}^2 a^2$. Taking the $\ell_\infty$ norm over $i = 1,\ldots,L$ gives
\begin{align*}
  \norm{\Phi(k^0+x) - \Phi(k^0-x)}_\infty
  &\ge 2\norm{Jx}_\infty - \norm{e}_\infty \\
  &\ge 2a\mu - \frac{1}{2} Q_{\max}^2 a^2 \ge a\mu > 0.
\end{align*}
Evaluating \eqref{eq:keydist} at a maximizing query $q_i$ and taking half the antipodal state distance proves \eqref{eq:w4lower}.

\emph{Matching upper bound.} Let $\Psi_{\mathrm{key}}: C_a \to \F_a$ be $k \mapsto S_k$. For any $k \in C_a$, $D_k(ih) = \sum_{j=1}^L z_j^i$ ($i = 1,\ldots,L$, $z_j = e^{h k_j} > 0$) are the first $L$ power sums of $(z_1,\ldots,z_L)$. Newton's identities uniquely recover the elementary symmetric polynomials and hence the multiset $\{z_1,\ldots,z_L\}$; strict ordering on $C_a$ uniquely recovers $k$. Because $C_a$ is compact and $(\F_a, \dQ)$ is Hausdorff, $\Psi_{\mathrm{key}}$ is a homeomorphism onto $\F_a$. Thus $E = \Psi_{\mathrm{key}}^{-1}: \F_a \to \R^L$ is continuous, and $G(y) = S_{\Pi_{C_a}(y)}$ (clipping coordinates into $C_a$) achieves error $0$, proving $w_L^{\mathrm{app}}(\F_a) = 0$ and \eqref{eq:w4exact}.
\end{proof}

One might ask whether Theorem~\ref{thm:w4} is merely an artifact of allowing arbitrary real-valued suffix masses $t \ge 0$. The following corollary shows that the hidden-mass obstruction survives under ordinary unit-weight KV appends, losing at most a factor of $8/9$.

\begin{corollary}[Survival under repeated unit-weight appends]\label{cor:w4phys}
In the setting of Theorem~\ref{thm:w4}, restrict further to positive strictly ordered reference keys $0 < k_1^0 < \cdots < k_L^0 < K_{\max}$ and choose $a > 0$ small enough that $C_a \subset (0, K_{\max}]^L$ preserves strict ordering when $L \ge 2$, i.e., $a < \min\bigl\{k_1^0,\, K_{\max}-k_L^0,\, \tfrac{1}{2}\min_{1\le j<L}(k_{j+1}^0 - k_j^0)\bigr\}$ with $\min\emptyset = +\infty$ when $L = 1$. Suppose the unit-weight KV pair $(k_{\mathrm{app}}, v_{\mathrm{app}}) = (0, -M)$ is permitted and may be appended in any finite multiplicity $n \in \Z_{\ge 0}$ (adding integer mass $n e^{q\cdot 0} = n$ and numerator $-nM$). Define the physical-continuation distance
\begin{equation}\label{eq:dphys}
\begin{aligned}
  \dphys(S, \hat S; q) &= \sup_{n \in \Z_{\ge 0}} \left|H_{n,-M}(S; q) - H_{n,-M}(\hat S; q)\right|,\\
  \dQphys(S, \hat S) &= \sup_{q \in \Q} \dphys(S, \hat S; q),
\end{aligned}
\end{equation}
and the corresponding physical continuous width
\[
  w_r^{\mathrm{phys}}(\F_a) = \inf_{E,\, G} \sup_{S \in \F_a} \dQphys\bigl(S, G(E(S))\bigr),
\]
where $E: (\F_a, \dQphys) \to \R^r$ is continuous in the $\dQphys$-topology and $G: \R^r \to \K$ is arbitrary. If $a \le 2\mu / Q_{\max}^2$, then for every $r < L$,
\begin{equation}\label{eq:w4phys}
  w_r^{\mathrm{phys}}(\F_a)
  \ge \frac{8}{9} \e_{\mathrm{hm}}
  \ge \frac{8}{9} M \tanh\!\left(\frac{a\mu}{4}\right) > 0.
\end{equation}
Moreover, $\dQphys$ is a metric on $\F_a$ under which $w_L^{\mathrm{phys}}(\F_a) = 0$, so for every $0 \le \e < \frac{8}{9}\e_{\mathrm{hm}}$, the operational minimum physical dimension is $r_{\min}^{\mathrm{phys}}(\F_a, \e) = L$.
\end{corollary}
\begin{proof}
For any two states $S_k, S_{\hat k} \in \F_a$ with historical masses $D = D_k(q)$ and $\hat D = D_{\hat k}(q)$ at a query $q$, appending $n \in \Z_{\ge 0}$ copies of $(0,-M)$ yields discrepancy $2M f(n)$, where
\[
  f(t) = \frac{|D - \hat D|\, t}{(D + t)(\hat D + t)},\qquad t \ge 0.
\]
If $D = \hat D$, then both $f(t) \equiv 0$ and $\dapp(S_k, S_{\hat k}; q) = 0$, so $\dphys(S_k, S_{\hat k}; q) = \frac{8}{9}\dapp(S_k, S_{\hat k}; q) = 0$ trivially.

Now suppose $D \ne \hat D$ and $t^* = \sqrt{D \hat D} \ge 1$. Choose the realizable integer multiplicity $n = \lceil t^* \rceil \in \Z_{\ge 1}$, and write $c = n / t^* \in [1, 2)$ and $z = \sqrt{\hat D / D} > 0$, so $D = t^*/z$ and $\hat D = z t^*$. Direct substitution gives $f(t^*) = |D-\hat D|z / [t^*(1+z)^2]$ and $f(ct^*) = |D-\hat D|cz / [t^*(1+cz)(z+c)]$. Expanding the denominator via $1 + z^2 = (1+z)^2 - 2z$ yields $(1+cz)(z+c) = c(1+z)^2 + (c-1)^2 z$, so
\[
  \frac{f(c t^*)}{f(t^*)}
  = \frac{c(1+z)^2}{(1+cz)(z+c)}
  = \frac{c}{c + (c-1)^2 \frac{z}{(1+z)^2}}.
\]
Because $(1+z)^2 - 4z = (1-z)^2 \ge 0$, we have $\frac{z}{(1+z)^2} \le \frac{1}{4}$ for all $z > 0$, hence
\begin{equation}\label{eq:eightninths}
  \frac{f(c t^*)}{f(t^*)}
  \ge \frac{c}{c + \tfrac{1}{4}(c-1)^2}
  = \frac{4c}{(1+c)^2}
  \ge \frac{8}{9}
\end{equation}
for all $c \in [1, 2]$, since $c \mapsto 4c/(1+c)^2$ has derivative $4(1-c)/(1+c)^3 \le 0$ on $[1,2]$ and attains its minimum $8/9$ at $c = 2$. Because every $k \in C_a$ has $k_j > 0$ and every probe $q_i = ih > 0$, we have $D_k(q_i) = \sum_{j=1}^L e^{q_i k_j} > L \ge 1$ for all $k \in C_a$ and $i = 1,\ldots,L$; hence $t^* = \sqrt{D_k(q_i) D_{\hat k}(q_i)} > 1$ at every probe query, yielding $\dphys(S_k, S_{\hat k}; q_i) \ge \frac{8}{9}\dapp(S_k, S_{\hat k}; q_i)$ for all $k, \hat k \in C_a$ (including equal masses).

\emph{Continuity and lower bound.} Every $(n, -M)$ is an admissible relaxed continuation, so $\dQphys \le \dQ$ on $\K$. Because $\Psi_{\mathrm{key}}: C_a \to (\F_a, \dQ)$ is continuous, the domination $\dQphys \le \dQ$ immediately implies that $\Psi_{\mathrm{key}}: C_a \to (\F_a, \dQphys)$ is continuous. Thus any $\dQphys$-continuous encoder $E: \F_a \to \R^r$ ($r < L$) induces a continuous map $x \mapsto E(S_{k^0+x})$ on $[-a,a]^L$, yielding an antipodal collision $x, -x$ ($\norm{x}_\infty = a$). At a maximizing probe $q_i$ from Theorem~\ref{thm:w4}, $|\Phi_i(k^0+x) - \Phi_i(k^0-x)| \ge 2a\mu - \tfrac{1}{2}Q_{\max}^2 a^2 > 0$, so $D_{k^0+x}(q_i) \ne D_{k^0-x}(q_i)$ and \eqref{eq:eightninths} yields \eqref{eq:w4phys}.

\emph{Topology for the matching upper bound.} Although $\dQphys$ is only a pseudometric on all of $\K$ (states with $F_S \equiv -M$ are collapsed), on $\F_a$ it is a \emph{metric}: if $\dQphys(S_k, S_{\hat k}) = 0$, evaluating at $n = 1$ gives $H_{1,-M}(S_k; q_i) = M\frac{D_k(q_i)-1}{D_k(q_i)+1} = H_{1,-M}(S_{\hat k}; q_i)$, forcing $D_k(q_i) = D_{\hat k}(q_i)$ for all $i = 1,\ldots,L$, and hence $k = \hat k$ by Newton's identities and strict ordering on $C_a$. Thus $\Psi_{\mathrm{key}}: C_a \to (\F_a, \dQphys)$ is a continuous bijection from a compact space onto a Hausdorff metric space, hence a homeomorphism. Consequently $E = \Psi_{\mathrm{key}}^{-1}: (\F_a, \dQphys) \to \R^L$ is continuous in the $\dQphys$-topology, and $G(y) = S_{\Pi_{C_a}(y)}$ achieves zero error, proving $w_L^{\mathrm{phys}}(\F_a) = 0$ and $r_{\min}^{\mathrm{phys}}(\F_a, \e) = L$.
\end{proof}

\begin{remark}[Conditioning, ceiling choice, and open-loop horizon]\label{rem:w4scope}
In Corollary~\ref{cor:w4phys}, choosing the ceiling multiplicity $n = \lceil t^* \rceil$ guarantees $c = n / t^* \in [1, 2)$ uniformly whenever $t^* > 1$, yielding the universal analytical floor $8/9$ via \eqref{eq:eightninths} without case splitting. This $8/9$ factor is a clean analytical floor rather than a tight integer optimum: in Fig.~\ref{fig:validation}(a) ($D = 1$, $\hat D = 10$), $n = \lfloor t^* \rfloor = 3$ achieves $1.0385M$ versus $1.0286M$ for $n = \lceil t^* \rceil = 4$. Two quantitative trade-offs govern these 1D local open-loop bounds. First, because $J$ is a scaled transposed Vandermonde matrix, $\mu = \norm{J^{-1}}_{\infty\to\infty}^{-1}$ decays rapidly with $L$. Second, under $k \in C_a \subset (0, K_{\max}]^L$ and probes $q_i \in (0, Q_{\max}]$, every denominator obeys $D_k(q_i) = \sum_{j=1}^L e^{q_i k_j} \le L e^{Q_{\max} K_{\max}}$, so $n = \lceil t^* \rceil$ requires at most $H_* = \lceil L e^{Q_{\max} K_{\max}} \rceil$ appended copies of the permitted unit-weight token $(0,-M)$ uniformly over $\F_a$ (an open-loop horizon bound rather than a short-horizon or closed-loop guarantee). Appendix~\ref{app:conditioning} tabulates exact high-precision values of $\mu$, the maximal admissible radius $a_{\max}$, $\e_{\mathrm{hm}}/M$, and $H_*$ for $L \in \{2, 3, 4, 6\}$, exhibiting this trade-off. As Theorem~\ref{thm:w4orth} next shows, both the super-exponential conditioning decay and the exponential suffix horizon are artifacts of forcing $d_k = 1$ rather than intrinsic limitations of the hidden-mass obstruction.
\end{remark}

\begin{theorem}[Well-conditioned high-dimensional hidden-mass obstruction]\label{thm:w4orth}
Let $d_k \ge L \ge 2$, $d_v = 1$, and let $e_1,\ldots,e_L \in \R^{d_k}$ be orthonormal vectors. Fix a probe scale $Q_{\max} > 0$, a perturbation radius $a > 0$, and set the reference key radius $K_0 := Q_{\max}^{-1} \log(L - 1)$ (so $e^{Q_{\max} K_0} = L - 1$). Assume the compact query domain $\Q \subset \R^{d_k}$ contains the $L$ probe queries $q_i = Q_{\max} e_i$ ($i = 1,\ldots,L$). For each $\theta = (\theta_1,\ldots,\theta_L)^\top \in [-a, a]^L$, define the $L$ variable keys and constant values
\begin{equation}\label{eq:orthkeys}
  k_j(\theta) = (K_0 + \theta_j) e_j,\qquad
  v_j \equiv M,\qquad j = 1,\ldots,L,
\end{equation}
inducing $S_\theta = (M D_\theta, D_\theta) \in \K$ with $F_{S_\theta}(q) \equiv M$ for all $q \in \Q$, and set $\F_a^{\mathrm{orth}} := \{S_\theta : \theta \in [-a, a]^L\} \subseteq \K$. Define the $L$-independent threshold $\e_{\mathrm{orth}} := M \tanh(Q_{\max} a / 4)$.
\begin{enumerate}
\item \emph{Relaxed observational appends:} For every $r < L$,
\begin{equation}\label{eq:orthrel}
  w_r^{\mathrm{app}}(\F_a^{\mathrm{orth}}) \ge \e_{\mathrm{orth}},
\end{equation}
and $w_L^{\mathrm{app}}(\F_a^{\mathrm{orth}}) = 0$ via a continuous exact encoder into $\R^L$. Consequently, for every $0 \le \e < \e_{\mathrm{orth}}$,
\begin{equation}\label{eq:orthexact}
  r_{\min}(\F_a^{\mathrm{orth}}, \e) = L.
\end{equation}
\item \emph{Physical unit-weight appends:} Suppose the unit-weight KV pair $(k_{\mathrm{app}}, v_{\mathrm{app}}) = (0, -M)$ is permitted in any finite multiplicity $n \in \Z_{\ge 0}$. Under $\dQphys$ from \eqref{eq:dphys}, for every $r < L$,
\begin{equation}\label{eq:orthphys}
  w_r^{\mathrm{phys}}(\F_a^{\mathrm{orth}}) \ge \frac{8}{9}\e_{\mathrm{orth}},
\end{equation}
and $w_L^{\mathrm{phys}}(\F_a^{\mathrm{orth}}) = 0$, so $r_{\min}^{\mathrm{phys}}(\F_a^{\mathrm{orth}}, \e) = L$ for every $0 \le \e < \frac{8}{9}\e_{\mathrm{orth}}$. Moreover, the lower bound is attained within the uniform linear suffix horizon
\begin{equation}\label{eq:orthH}
  H_* = \left\lceil (L - 1)\bigl(e^{Q_{\max} a} + 1\bigr) \right\rceil = O(L).
\end{equation}
\end{enumerate}
\end{theorem}
\begin{proof}
\emph{Exact decoupling on probes.} Because $e_1,\ldots,e_L$ are orthonormal, $q_i^\top k_j(\theta) = Q_{\max}(K_0 + \theta_i)$ for $j = i$ and $q_i^\top k_j(\theta) = 0$ for $j \ne i$. Using $e^{Q_{\max} K_0} = L - 1$, the historical denominator at probe $q_i$ depends solely on $\theta_i$:
\begin{equation}\label{eq:orthdecouple}
  D_\theta(q_i)
  = e^{Q_{\max}(K_0 + \theta_i)} + (L - 1)
  = (L - 1)\bigl(e^{Q_{\max} \theta_i} + 1\bigr).
\end{equation}

\emph{Part~(1) (Relaxed lower and upper bounds).} By compactness of $\Q \times [-a,a]^L$ and uniform positivity of $D_\theta$, the map $\Psi_{\mathrm{orth}}: [-a, a]^L \to \F_a^{\mathrm{orth}}$ ($\theta \mapsto S_\theta$) is continuous into $(\K, \tau)$ by \eqref{eq:reverse} and hence into $(\K, \dQ)$ by \eqref{eq:m4} of Theorem~\ref{thm:m4}. For any continuous encoder $E: (\F_a^{\mathrm{orth}}, \dQ) \to \R^r$ with $r < L$, Borsuk--Ulam yields an antipodal collision $\theta, -\theta \in \partial[-a, a]^L$ ($\norm{\theta}_\infty = a$). Pick $i \in \{1,\ldots,L\}$ with $|\theta_i| = a$; by symmetry assume $\theta_i = a$. By \eqref{eq:orthdecouple},
\begin{align*}
  \frac{D_\theta(q_i)}{D_{-\theta}(q_i)}
  &= \frac{(L - 1)(e^{Q_{\max} a} + 1)}{(L - 1)(e^{-Q_{\max} a} + 1)} \\
  &= \frac{e^{Q_{\max} a} + 1}{e^{-Q_{\max} a}(1 + e^{Q_{\max} a})}
  = e^{Q_{\max} a}.
\end{align*}
Hence $|\log(D_\theta(q_i)/D_{-\theta}(q_i))| = Q_{\max} a$ identically, independent of $L$. Evaluating \eqref{eq:keydist} at $q_i$ gives
\[
  \dQ(S_\theta, S_{-\theta}) \ge 2M\tanh(Q_{\max} a / 4) = 2\e_{\mathrm{orth}},
\]
so any decoder $G$ incurs error at least $\e_{\mathrm{orth}}$ on one of the two colliding inputs, proving \eqref{eq:orthrel}.

For the matching upper bound, \eqref{eq:orthdecouple} uniquely recovers each coordinate $\theta_i = Q_{\max}^{-1}\log\bigl(\frac{D_\theta(q_i)}{L - 1} - 1\bigr)$ ($i = 1,\ldots,L$), so $\Psi_{\mathrm{orth}}$ is a continuous bijection from the compact cube $[-a, a]^L$ onto the Hausdorff metric space $(\F_a^{\mathrm{orth}}, \dQ)$, hence a homeomorphism. Setting $E = \Psi_{\mathrm{orth}}^{-1}$ and $G(y) = S_{\Pi_{[-a,a]^L}(y)}$ (coordinatewise clipping into $[-a,a]^L$) achieves zero error, proving $w_L^{\mathrm{app}}(\F_a^{\mathrm{orth}}) = 0$ and \eqref{eq:orthexact}.

\emph{Part~(2) (Physical-continuation survival and linear horizon).} Because $\dQphys \le \dQ$, $\Psi_{\mathrm{orth}}: [-a, a]^L \to (\F_a^{\mathrm{orth}}, \dQphys)$ is continuous. Any $\dQphys$-continuous encoder $E: \F_a^{\mathrm{orth}} \to \R^r$ ($r < L$) therefore admits an antipodal collision $\theta, -\theta \in \partial[-a,a]^L$ with $|\theta_i| = a$ for some $i$. By \eqref{eq:orthdecouple}, $D_\theta(q_i) > L - 1 \ge 1$ for every $\theta \in [-a,a]^L$ and $i \in \{1,\ldots,L\}$, and $D_\theta(q_i) \ne D_{-\theta}(q_i)$, so $t^* = \sqrt{D_\theta(q_i) D_{-\theta}(q_i)} > 1$. Choosing the realizable integer multiplicity $n = \lceil t^* \rceil \in \Z_{\ge 1}$ and applying \eqref{eq:eightninths} from Corollary~\ref{cor:w4phys} yields
\[
  \dphys(S_\theta, S_{-\theta}; q_i)
  \ge \frac{8}{9}\dapp(S_\theta, S_{-\theta}; q_i)
  \ge \frac{16}{9}\e_{\mathrm{orth}},
\]
proving \eqref{eq:orthphys}. Moreover, $D_\theta(q_i) \le (L - 1)(e^{Q_{\max} a} + 1)$ uniformly over $\theta \in [-a, a]^L$ and $i \in \{1,\ldots,L\}$, so every required multiplicity obeys $n = \lceil t^* \rceil \le H_*$ in \eqref{eq:orthH} (at antipodal collisions in fact $t^* = 2(L-1)\cosh(Q_{\max} a/2) \le (L-1)(e^{Q_{\max} a} + 1)$).

Finally, $\dQphys$ is a metric on $\F_a^{\mathrm{orth}}$: if $\dQphys(S_\theta, S_{\hat\theta}) = 0$, evaluating at $n = 1$ gives $H_{1,-M}(S_\theta; q_i) = M\frac{D_\theta(q_i) - 1}{D_\theta(q_i) + 1} = H_{1,-M}(S_{\hat\theta}; q_i)$, forcing $D_\theta(q_i) = D_{\hat\theta}(q_i)$ and hence $\theta_i = \hat\theta_i$ for all $i$. By compactness of $[-a,a]^L$, $\Psi_{\mathrm{orth}}: [-a,a]^L \to (\F_a^{\mathrm{orth}}, \dQphys)$ is a homeomorphism, so $E = \Psi_{\mathrm{orth}}^{-1}$ and $G(y) = S_{\Pi_{[-a,a]^L}(y)}$ achieve zero error, proving $w_L^{\mathrm{phys}}(\F_a^{\mathrm{orth}}) = 0$ and $r_{\min}^{\mathrm{phys}}(\F_a^{\mathrm{orth}}, \e) = L$ for all $0 \le \e < \frac{8}{9}\e_{\mathrm{orth}}$.
\end{proof}

\parhead{Eliminating 1D conditioning decay.}
Unlike Theorem~\ref{thm:w4}, where probing along one axis produces a transposed Vandermonde Jacobian whose conditioning decays super-exponentially with $L$ (Appendix~\ref{app:conditioning}), the orthogonal-axis log-partition map $\Phi_i(\theta) = \log D_\theta(q_i)$ has strictly diagonal Jacobian $J = \tfrac{1}{2}Q_{\max} I_L$ at $\theta = 0$ ($\mu = \tfrac{1}{2}Q_{\max}$ independent of $L$), and exact decoupling removes any Taylor remainder. For instance, choosing $Q_{\max} a = 2$ gives $\e_{\mathrm{orth}}/M = \tanh(1/2) \approx 0.4621$ (physical threshold $\frac{8}{9}\e_{\mathrm{orth}}/M \approx 0.4108$) uniformly for \emph{all} $L \ge 2$ within the linear suffix horizon $H_* = \lceil (L-1)(e^2 + 1) \rceil \le 9(L-1) < 9L$, while each key has only logarithmic Euclidean norm $\norm{k_j(\theta)}_2 = |K_0 + \theta_j| \le Q_{\max}^{-1}\log(L - 1) + a = O(\log L)$. Thus Theorem~\ref{thm:w4} demonstrates that the hidden-mass obstruction exists even in 1D, whereas Theorem~\ref{thm:w4orth} proves it remains macroscopically large ($\e_{\mathrm{orth}}/M = \Omega(1)$) at arbitrary $L$.

\begin{remark}[Practical regime $1 < d_k < L$]\label{rem:practical_dk}
Theorem~\ref{thm:w4orth} assumes $d_k \ge L$, whereas in practical transformers context length $L$ typically far exceeds head dimension $d_k$. For any $L > d_k \ge 1$ ($L \ge 2$), setting $d := \min\{L, d_k\} = d_k$, placing $d$ variable keys $k_j(\theta) = (K_0 + \theta_j)e_j$ ($j = 1,\ldots,d$, $\theta \in [-a,a]^d$) along orthonormal axes $e_1,\ldots,e_d \in \R^{d_k}$ with $K_0 = Q_{\max}^{-1}\log(L-1)$, and fixing the remaining $L - d$ history keys at $k_j \equiv 0$ ($j = d+1,\ldots,L$) with constant values $v_j \equiv M$ ($F_{S_\theta} \equiv M$) preserves the exact decoupled probe denominator \eqref{eq:orthdecouple},
\begin{align*}
  D_\theta(q_i) &= e^{Q_{\max}(K_0+\theta_i)} + (d - 1) + (L - d) \\
  &= (L - 1)\bigl(e^{Q_{\max}\theta_i} + 1\bigr)
\end{align*}
for every $i = 1,\ldots,d$. Applying the identical Borsuk--Ulam argument on $[-a, a]^d$ immediately shows that any family $\F$ of length-$L$ constant-output histories in $\R^{d_k}$ containing this $d$-parameter subfamily obeys the macroscopic, $L$-independent lower bounds $r_{\min}(\F, \e) \ge \min\{L, d_k\}$ ($0 \le \e < \e_{\mathrm{orth}}$) and $r_{\min}^{\mathrm{phys}}(\F, \e) \ge \min\{L, d_k\}$ ($0 \le \e < \frac{8}{9}\e_{\mathrm{orth}}$) within the same linear suffix horizon $H_* = \lceil (L - 1)(e^{Q_{\max}a} + 1)\rceil = O(L)$ and logarithmic key norm $O(\log L)$. Extending well-conditioned hidden-mass lower bounds beyond $d_k$ when $L \gg d_k$ requires non-orthogonal key geometry (Open Problem~(i) in Section~\ref{sec:conclusion}).
\end{remark}

\section{Collective Width Separation}
\label{sec:collective}

\subsection{Collective Model and Operational Accounting}
\label{sec:c1}

Section~\ref{sec:w4} isolated the single-session nonlinear obstruction created by \emph{unobserved denominator mass} when keys vary and current normalized outputs remain identically constant. We now extend the append-observational width framework to $B$ concurrent sessions under both mechanisms: Section~\ref{sec:c2} holds keys---and hence $D_{\mathrm{fix}}$---fixed across sessions on a finite circular grid $\Q_L$ to isolate \emph{cross-session value-subspace sharing} with nontrivial spectral truncation (Theorem~\ref{thm:c2}), while Section~\ref{sec:c3} unifies Sections~\ref{sec:w4} and~\ref{sec:collective} by proving exact matching collective dimensions ($R_{\joint}(\e) = s + Bp$ versus $R_{\ind}(\e) = B(s + p)$) driven purely by \emph{variable-key hidden historical mass} ($F_{S_i} \equiv M$) over arbitrary compact query domains $\Q$ containing orthogonal probes, surviving repeated physical appends within a linear suffix horizon (Theorem~\ref{thm:c3}). Extending these collective widths to \emph{unknown} (unaligned) shared subspaces and non-orthogonal key geometry remains Open Problem~(i) in Section~\ref{sec:conclusion}.

\begin{definition}[Collective dynamic functional width and accounting]\label{def:collective}
Fix a query domain $\Q$, session count $B \ge 1$, and a joint family $\F_{\joint} \subseteq (\K)^B$, equipped with the product metric $d_{\Q}^{(B)}(\mathbf{S}, \hat{\mathbf{S}}) = \max_{1 \le i \le B} \dQ(S_i, \hat S_i)$. For $R \ge 0$, define
\begin{equation}\label{eq:jointwidth}
\begin{aligned}
  &w_R^{(B)}(\F_{\joint}; \Q)\\
  &\quad = \inf_{E,\,\{G_i\}} \sup_{\mathbf{S} \in \F_{\joint}} \max_{1 \le i \le B} \dQ\bigl(S_i, G_i(E(\mathbf{S}))\bigr),
\end{aligned}
\end{equation}
where $E: \F_{\joint} \to \R^R$ is continuous in $d_{\Q}^{(B)}$ and $G_i: \R^R \to \K$ are arbitrary decoders. Let $R_{\joint}(\e)$ be the smallest $R \ge 0$ admitting continuous $E$ and arbitrary $G_1,\ldots,G_B$ with uniform worst-session error at most $\e$. Let $\F_i = \{S_i : \mathbf{S} \in \F_{\joint}\}$ be the $i$-th marginal, $r_i^{\ind}(\e) = r_{\min}(\F_i, \e)$ under the same metric $\dQ$, and $R_{\ind}(\e) = \sum_{i=1}^B r_i^{\ind}(\e)$ ($= B r_{\ind}(\e)$ for symmetric marginals). When $R_{\joint}(\e) > 0$, define the collective gain
\begin{equation}\label{eq:gc}
  G_C(\e) = \frac{R_{\ind}(\e)}{R_{\joint}(\e)}.
\end{equation}
\end{definition}

Lower bounds permit every decoder $G_i$ to read all $R$ coordinates; upper bounds use the stricter shared-plus-private form $E(\mathbf{S}) = (z_{\sh}, z_1,\ldots,z_B)$ with $\hat S_i = G_i(z_{\sh}, z_i)$ and $R = r_{\sh} + B r_{\priv}$ (a functional separation of inputs, not a cryptographic privacy guarantee). Fixed public family parameters are excluded equally from $R_{\joint}(\e)$ and $R_{\ind}(\e)$.

\begin{figure*}[!t]
\centering
\input{figures/validation_panels.tex}
\caption{Analytical and numerical illustrations of the main theoretical bounds (reproducible via \href{https://github.com/openkedge/pdds/blob/main/attention-nonlinear-widths/scripts/generate_validation_figures.py}{\texttt{scripts/generate\_validation\_figures.py}}, available at \url{https://github.com/openkedge/pdds/tree/main/attention-nonlinear-widths}). \textbf{(a)}~Hidden historical mass for constant-output states ($F\equiv M$) with $D=1$, $\hat D=10$, where $\Delta(t)/M = 18t/[(1+t)(10+t)]$: zero error at $t=0$, relaxed peak $\approx 1.0390\,M$ at $t^*=\sqrt{10}\approx 3.16$, ceiling-based proof construction $n=\lceil t^*\rceil=4$ yielding $\approx 1.0286\,M$ (compared with $n=3$ yielding $\approx 1.0385\,M$; see Remark~\ref{rem:w4scope}), and the universal $\tfrac{8}{9}$ analytical floor ($\approx 0.9235\,M$). \textbf{(b)}~Exact thresholds $\e_-(\kappa),\e_+(\kappa)$ (solid) for the $L=16$ circular family at $a=b=1$ (admissible under \eqref{eq:family} whenever $M \ge 2 + \max_i |\beta_i|$) versus small-$\kappa$ asymptotics (dashed); over the evaluated range $\kappa \in [0.2, 12]$, $\e_+(\kappa) < \e_-(\kappa)$ holds for $\kappa < \kappa_\times \approx 7.83$. \textbf{(c)}~Collective compression gain $G_C(B) = B(s+p)/(s+Bp)$ of Theorems~\ref{thm:c2} and~\ref{thm:c3} for $s/p \in \{1,3,7\}$, approaching the saturation limits $1 + s/p \in \{2,4,8\}$.}
\label{fig:validation}
\end{figure*}
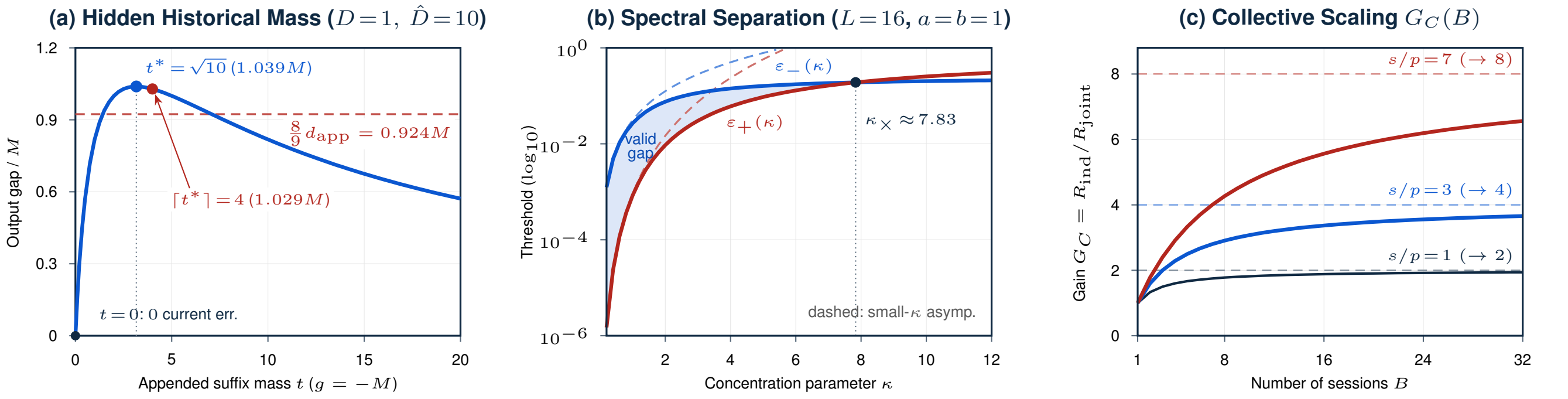

\subsection{Circular Harmonic Family on a Finite Query Grid}
\label{sec:c2}

Let $L \ge 3$ and $\theta_j = 2\pi j / L$ ($j = 0,\ldots,L-1$). We explicitly restrict the query domain to the finite circular grid
\begin{equation}\label{eq:QL}
  \Q_L = \bigl\{(\cos\theta_j,\, \sin\theta_j) : j = 0,\ldots,L-1\bigr\} \subset \R^2,
\end{equation}
with fixed keys $k_j = \kappa(\cos\theta_j, \sin\theta_j) \in \R^2$ ($\kappa > 0$), so $q_i^\top k_j = \kappa\cos(\theta_i - \theta_j)$. The normalized response matrix on $\Q_L$,
\begin{equation}\label{eq:circular}
  A_{ij} = \frac{e^{\kappa \cos(\theta_i - \theta_j)}}{\sum_{h=0}^{L-1} e^{\kappa \cos\theta_h}},\qquad 0 \le i,j \le L-1,
\end{equation}
is real symmetric and circulant. Inserting the Fourier--Bessel expansion $e^{\kappa\cos t} = \sum_{n \in \Z} I_{|n|}(\kappa) e^{int}$~\cite{dlmf} into the discrete Fourier transform gives strictly positive eigenvalues
\begin{equation}\label{eq:eigen}
  \lambda_n = \frac{\sum_{h \in \Z} I_{|n+hL|}(\kappa)}{\sum_{h \in \Z} I_{|hL|}(\kappa)} > 0,\qquad n = 0,\ldots,L-1,
\end{equation}
so $A$ is invertible. Real invariant subspaces of $A$ are built from complete frequency groups: conjugate sine/cosine pairs $\{n, L-n\}$ (dimension $2$) and, when $L$ is even, the Nyquist mode $n = L/2$ (dimension $1$).

Choose two orthogonal invariant real Fourier subspaces $U, V \subset \mathbf{1}^\perp \subset \R^L$ of complete frequency groups, radii $a, b > 0$, and distinct known session means $\beta_1,\ldots,\beta_B \in \R$ with $\max_i |\beta_i| + a + b \le M$. Define $\F_{\joint} \subset (\K)^B$ on $\Q_L$ by $S_i = S_{v_i}$ with
\begin{equation}\label{eq:family}
\begin{aligned}
  v_i &= \beta_i \mathbf{1} + x + y_i,\\
  x &\in U,\ \norm{x}_2 \le a,\qquad y_i \in V,\ \norm{y_i}_2 \le b.
\end{aligned}
\end{equation}
Each session obeys $\norm{v_i}_\infty \le |\beta_i| + a + b \le M$ and $F_i = \beta_i \mathbf{1} + Ax + Ay_i$ with the same fixed denominator on $\Q_L$.

Choose nonempty retained invariant subspaces $U_s \subseteq U$ and $V_p \subseteq V$ of dimensions $s, p \ge 1$ composed of complete frequency groups. Let $\Pi_U, \Pi_V, \Pi_s, \Pi_p$ denote the orthogonal projections onto $U, V, U_s, V_p$, and let $\Omega_{\sh}, \Omega_{\priv} \subseteq \{0,\ldots,L-1\}$ be the DFT frequency index sets spanning the omitted orthogonal complements $U \cap U_s^\perp$ and $V \cap V_p^\perp$ (counting each conjugate pair with multiplicity $2$ and the Nyquist mode with multiplicity $1$, matching real dimension). Set $\lambda_s^{\sh} = \min_{n \in U_s} \lambda_n$ and $\lambda_p^{\priv} = \min_{n \in V_p} \lambda_n$, and define
\begin{align}
  T_{\sh}(s) &= \left(\frac{1}{L}\sum_{n \in \Omega_{\sh}} \lambda_n^2\right)^{1/2}, \nonumber\\
  T_{\priv}(p) &= \left(\frac{1}{L}\sum_{n \in \Omega_{\priv}} \lambda_n^2\right)^{1/2}, \nonumber\\
  \e_+(s,p) &= a T_{\sh}(s) + b T_{\priv}(p), \label{eq:thresholds}\\
  \e_-(s,p) &= L^{-1/2} \min\bigl\{a \lambda_s^{\sh},\, b \lambda_p^{\priv}\bigr\}, \nonumber
\end{align}
with empty sums equal to zero. (In shorthand we write $n \in U \setminus U_s$ for $n \in \Omega_{\sh}$ and $n \in V \setminus V_p$ for $n \in \Omega_{\priv}$.)

\begin{theorem}[Matching collective dimensions on $\Q_L$]\label{thm:c2}
For the circular harmonic family \eqref{eq:family} measured in $d_{\Q_L}$ over the finite query grid $\Q_L$ in \eqref{eq:QL},
\begin{align}
  R < s + Bp &\implies w_R^{(B)}(\F_{\joint}; \Q_L) \ge \e_-(s,p), \label{eq:c2lower}\\
  w_{s + Bp}^{(B)}(\F_{\joint}; \Q_L) &\le \e_+(s,p). \label{eq:c2upper}
\end{align}
Whenever $\e_+(s,p) \le \e < \e_-(s,p)$, the collective and independent operational dimensions under $d_{\Q_L}$ match exactly:
\begin{equation}\label{eq:gain}
\begin{aligned}
  R_{\joint}(\e) &= s + Bp,\qquad r_{\ind}(\e) = s + p,\\
  G_C(\e) &= \frac{B(s+p)}{s + Bp}.
\end{aligned}
\end{equation}
The attaining encoder is shared-plus-private with admissible reconstructions satisfying $\tau_{\Q_L}(S_i, \hat S_i) \le \e_+(s,p)/M$ (where $\tau_{\Q_L}$ denotes $\tau$ evaluated on the finite query grid $\Q_L$).
\end{theorem}
\begin{proof}
\emph{Lower bound.} Restrict \eqref{eq:family} to the centrally symmetric norm ball
\begin{multline*}
  \mathcal{B} = \bigl\{(x, y_1,\ldots,y_B) \in U_s \times V_p^B :\\
  \max\{\norm{x}_2/a,\, \max_i \norm{y_i}_2/b\} \le 1\bigr\}
\end{multline*}
of dimension $m = s + Bp$. By Borsuk--Ulam, any continuous $E: \F_{\joint} \to \R^R$ ($R < m$) collides on an antipodal boundary pair $z, -z \in \partial\mathcal{B}$. At $t=0$, $d_{\Q_L}^{(B)}(\mathbf{S}(z), \mathbf{S}(-z)) \ge 2\max_i \norm{A(x+y_i)}_\infty$, so one input incurs error at least $\max_i \norm{A(x+y_i)}_\infty$. Orthogonal $A$-invariance of $U_s$ and $V_p$ gives
\begin{align*}
  \norm{A(x+y_i)}_\infty
  &\ge L^{-1/2} \norm{A(x+y_i)}_2\\
  &= L^{-1/2}\bigl(\norm{Ax}_2^2 + \norm{Ay_i}_2^2\bigr)^{1/2}.
\end{align*}
On $\partial\mathcal{B}$, either $\norm{x}_2 = a$ (giving $\norm{Ax}_2 \ge a\lambda_s^{\sh}$) or $\norm{y_i}_2 = b$ for some $i$ (giving $\norm{Ay_i}_2 \ge b\lambda_p^{\priv}$), proving \eqref{eq:c2lower} even when every decoder sees the full code.

\emph{Upper bound and equality.} Because $A$ is invertible and $U \perp V \perp \mathbf{1}$, $S_i \mapsto \Pi_s A^{-1}(F_{S_i}(\Q_L) - \beta_i\mathbf{1}) = \Pi_s x$ and $S_i \mapsto \Pi_p A^{-1}(F_{S_i}(\Q_L) - \beta_i\mathbf{1}) = \Pi_p y_i$ are continuous on $\F_{\joint}$ in $d_{\Q_L}^{(B)}$. Encode the $s$ coordinates of $\Pi_s x$ once and the $p$ coordinates of each $\Pi_p y_i$ per session ($R = s + Bp$). Each decoder $G_i$ projects received coordinates back into the radius-$a$ and radius-$b$ balls in $U_s, V_p$ and reconstructs $\hat v_i = \beta_i \mathbf{1} + \hat x + \hat y_i$ with the fixed keys and exact denominator on $\Q_L$, ensuring $\norm{\hat v_i}_\infty \le M$ and $\hat S_i \in \K$.

Because $x \in U$ and $y_i \in V$, we have $\Pi_U x = x$ and $\Pi_V y_i = y_i$, so the response residual on $\Q_L$ is
\[
  F_i - \hat F_i = C_{\sh} x + C_{\priv} y_i,
\]
where we define the subspace-restricted residual operators
\begin{equation}\label{eq:resops}
  C_{\sh} = A(\Pi_U - \Pi_s),\qquad C_{\priv} = A(\Pi_V - \Pi_p).
\end{equation}
Let $W \in \mathbb{C}^{L \times L}$ be the unitary DFT matrix ($W_{jn} = L^{-1/2} e^{i 2\pi j n / L}$). Because $U, U_s$ consist of complete frequency groups, $\Omega_{\sh}$ is closed under negation modulo $L$, $A = W \diag(\lambda_0,\ldots,\lambda_{L-1}) W^*$, and $\Pi_U - \Pi_s = W \diag(\mathbf{1}_{n \in \Omega_{\sh}}) W^*$; hence $C_{\sh} = W \diag(\lambda_n \mathbf{1}_{n \in \Omega_{\sh}}) W^*$ is real symmetric and circulant. Every row $j \in \{0,\ldots,L-1\}$ of $C_{\sh}$ therefore has squared Euclidean norm
\begin{align*}
  \norm{(C_{\sh})_{j,:}}_2^2
  &= (C_{\sh} C_{\sh}^*)_{jj} \\
  &= \sum_{n \in \Omega_{\sh}} |W_{jn}|^2 \lambda_n^2
  = \frac{1}{L} \sum_{n \in U \setminus U_s} \lambda_n^2.
\end{align*}
By Cauchy--Schwarz, $\norm{C_{\sh}}_{2\to\infty} = \max_j \norm{(C_{\sh})_{j,:}}_2$, and the identical derivation applies to $C_{\priv}$ on $\Omega_{\priv}$; thus
\begin{equation}\label{eq:rowops}
\begin{aligned}
  \norm{C_{\sh}}_{2\to\infty}^2 &= \frac{1}{L} \sum_{n \in U \setminus U_s} \lambda_n^2 = T_{\sh}(s)^2,\\
  \norm{C_{\priv}}_{2\to\infty}^2 &= \frac{1}{L} \sum_{n \in V \setminus V_p} \lambda_n^2 = T_{\priv}(p)^2.
\end{aligned}
\end{equation}
By the triangle inequality and $\norm{x}_2 \le a$, $\norm{y_i}_2 \le b$,
\[
  \norm{F_i - \hat F_i}_{\infty,\Q_L} \le \norm{C_{\sh}}_{2\to\infty} a + \norm{C_{\priv}}_{2\to\infty} b = \e_+(s,p).
\]
Since the denominator on $\Q_L$ is exact, $d_{\Q_L}(S_i, \hat S_i) = \norm{F_i - \hat F_i}_{\infty,\Q_L} \le \e_+(s,p)$, proving \eqref{eq:c2upper}, and \eqref{eq:reverse} gives $\tau_{\Q_L}(S_i, \hat S_i) \le \e_+(s,p)/M$. Under $\e_+(s,p) \le \e < \e_-(s,p)$, $R_{\joint}(\e) = s + Bp$; applying the same proof with $B = 1$ to each marginal yields $r_{\ind}(\e) = s + p$ and \eqref{eq:gain}.
\end{proof}

\begin{remark}[Necessity of the finite grid $\Q_L$]\label{rem:grid}
Restricting $\Q$ to $\Q_L$ is required for \eqref{eq:c2upper}: \eqref{eq:c2lower} holds on any domain containing $\Q_L$, whereas controlling $\sup_{\theta \in [0,2\pi)} |F_i(\theta) - \hat F_i(\theta)|$ between grid points would require a continuous-domain interpolation bound.
\end{remark}

\begin{proposition}[Nonvacuous spectral separation regime]\label{prop:nonvac}
Set $L = 16$, let $U$ be spanned by real frequency groups $\{1, 4\}$ (dimension $4$), and let $V$ be spanned by real frequency groups $\{2, 5\}$ (dimension $4$). Retain frequency $1$ in $U_s$ ($s=2$, discarding frequency $4$) and frequency $2$ in $V_p$ ($p=2$, discarding frequency $5$). Then for fixed $a, b > 0$, as $\kappa \to 0^+$,
\begin{equation}\label{eq:nonvac}
\begin{aligned}
  \e_+(2,2) &= \Theta(\kappa^4),\qquad \e_-(2,2) = \Theta(\kappa^2),\\
  \frac{\e_+(2,2)}{\e_-(2,2)} &= \Theta(\kappa^2) \to 0.
\end{aligned}
\end{equation}
Hence for all sufficiently small $\kappa > 0$, $[\e_+(2,2), \e_-(2,2))$ is nonempty with genuinely discarded shared and private modes.
\end{proposition}
\begin{proof}
Using $I_m(\kappa) = \frac{(\kappa/2)^m}{m!}(1 + O(\kappa^2))$ as $\kappa \to 0^+$, for $L = 16$ and $n \in \{1,2,4,5\}$ all nonzero aliases obey $|n + 16h| \ge 11$ and the denominator of \eqref{eq:eigen} is $1 + O(\kappa^2)$, so $\lambda_n = \frac{(\kappa/2)^n}{n!}(1 + O(\kappa^2))$. Counting both real modes in each discarded group yields $T_{\sh}(2) = \sqrt{2/16}\,\lambda_4 = \Theta(\kappa^4)$ and $T_{\priv}(2) = \sqrt{2/16}\,\lambda_5 = \Theta(\kappa^5)$, so $\e_+(2,2) = \Theta(\kappa^4)$. Meanwhile $\lambda_2^{\sh} = \lambda_1 = \Theta(\kappa)$ and $\lambda_2^{\priv} = \lambda_2 = \Theta(\kappa^2)$, giving $\e_-(2,2) = \tfrac{1}{4}\min\{a\lambda_1, b\lambda_2\} = \Theta(\kappa^2)$ and \eqref{eq:nonvac}.
\end{proof}

\parhead{Scaling and interpretation.}
For $p > 0$ and fixed $s, p$, $\lim_{B \to \infty} G_C(\e) = 1 + s/p$ with $G_C(\e) \le B$ (Fig.~\ref{fig:validation}(c)). Panel~(b) evaluates $\e_+(\kappa)$ and $\e_-(\kappa)$ for $L=16$ and $a=b=1$ (admissible under \eqref{eq:family} whenever $M \ge 2 + \max_i |\beta_i|$, distinct from the normalized ratio $\e_+/M$ governing $\tau_{\Q_L}$). Over the evaluated range $\kappa \in [0.2, 12]$, $\e_+(\kappa) < \e_-(\kappa)$ holds for $\kappa < \kappa_\times \approx 7.83$ (as $\kappa \to \infty$, every $\lambda_n(\kappa) \to 1$ on $\Q_{16}$, so $\e_+(\kappa) \to 1/\sqrt{2} > 1/4 \leftarrow \e_-(\kappa)$). Theorem~\ref{thm:c2} thus provides an exact collective-dimension baseline under prescribed fixed-key shared and private spectral structure.

\subsection{Collective Width Separation Under Hidden Historical Mass}
\label{sec:c3}

We now lift the orthogonal-axis hidden-mass construction of Theorem~\ref{thm:w4orth} to $B$ concurrent sessions. Every session has identically constant normalized output $F_{S_i}(q) \equiv M$ across all $q \in \Q$---zero current-output differences anywhere---yet shared and private variable keys force exact collective width separation $R_{\joint}(\e) = s + Bp$ versus $r_{\ind}(\e) = s + p$ purely through unobserved historical denominator mass. Because the shared-plus-private encoder recovers the key parameters exactly, zero error holds across the entire compact domain $\Q$, removing the finite-grid restriction $\Q_L$ noted in Remark~\ref{rem:grid}.

\begin{theorem}[Collective width separation under hidden historical mass]\label{thm:c3}
Let $B \ge 1$, $s, p \ge 1$, $L := s + p \ge 2$, $d_k \ge L$, and $d_v = 1$. Let $e_1,\ldots,e_s, f_1,\ldots,f_p \in \R^{d_k}$ be $L$ mutually orthonormal vectors. Fix a probe scale $Q_{\max} > 0$, perturbation radii $a, b > 0$, and set the reference key radius $K_0 := Q_{\max}^{-1} \log(L - 1)$ (so $e^{Q_{\max} K_0} = L - 1$). Assume the compact query domain $\Q \subset \R^{d_k}$ contains the $L$ probe queries
\begin{equation}\label{eq:collprobes}
\begin{aligned}
  q_u^{\sh} &= Q_{\max} e_u \;\; (1 \le u \le s),\\
  q_w^{\priv} &= Q_{\max} f_w \;\; (1 \le w \le p).
\end{aligned}
\end{equation}
For each shared parameter $\theta \in [-a, a]^s$ and private parameters $\phi_1,\ldots,\phi_B \in [-b, b]^p$, define the $i$-th session state $S_i(\theta, \phi_i) \in \K$ ($1 \le i \le B$) from the $L$ variable keys and constant values
\begin{equation}\label{eq:collkeys}
\begin{aligned}
  k_u^{\sh}(\theta) &= (K_0 + \theta_u) e_u \;\; (1 \le u \le s),\\
  k_{i,w}^{\priv}(\phi_i) &= (K_0 + \phi_{i,w}) f_w \;\; (1 \le w \le p),\qquad
  v \equiv M.
\end{aligned}
\end{equation}
Every session in the resulting joint family $\F_{\joint}^{\mathrm{hm}} \subseteq (\K)^B$ produces the identical constant normalized output $F_{S_i}(q) \equiv M$ for all $q \in \Q$. Set $\e_{\mathrm{coll}} := M \tanh\bigl(Q_{\max} \min\{a, b\} / 4\bigr)$.
\begin{enumerate}
\item \emph{Relaxed observational appends:} Under the product metric $d_{\Q}^{(B)}$ of Definition~\ref{def:collective},
\begin{equation}\label{eq:c3rel}
\begin{aligned}
  &R < s + Bp \implies w_R^{(B)}(\F_{\joint}^{\mathrm{hm}}; \Q) \ge \e_{\mathrm{coll}},\\
  &w_{s + Bp}^{(B)}(\F_{\joint}^{\mathrm{hm}}; \Q) = 0.
\end{aligned}
\end{equation}
Consequently, for every $0 \le \e < \e_{\mathrm{coll}}$, the collective and independent operational dimensions match exactly:
\begin{equation}\label{eq:c3gain}
\begin{aligned}
  R_{\joint}(\e) &= s + Bp,\qquad r_{\ind}(\e) = s + p,\\
  G_C(\e) &= \frac{B(s + p)}{s + Bp},
\end{aligned}
\end{equation}
attained by a shared-plus-private encoder with zero error across all $q \in \Q$.
\item \emph{Physical unit-weight appends:} Under the product physical-continuation metric $d_{\Q}^{\mathrm{phys},(B)}(\mathbf{S}, \hat{\mathbf{S}}) := \max_{1 \le i \le B} \dQphys(S_i, \hat S_i)$ using repeated appends of $(0, -M)$ (as in Corollary~\ref{cor:w4phys}), let $w_R^{\mathrm{phys},(B)}(\F_{\joint}^{\mathrm{hm}}; \Q)$ denote the corresponding collective physical width over $d_{\Q}^{\mathrm{phys},(B)}$-continuous encoders $E$ and arbitrary decoders $G_i: \R^R \to \K$. Then
\begin{equation}\label{eq:c3phys}
\begin{aligned}
  &R < s + Bp \implies w_R^{\mathrm{phys},(B)}(\F_{\joint}^{\mathrm{hm}}; \Q) \ge \frac{8}{9}\e_{\mathrm{coll}},\\
  &w_{s + Bp}^{\mathrm{phys},(B)}(\F_{\joint}^{\mathrm{hm}}; \Q) = 0.
\end{aligned}
\end{equation}
Thus $R_{\joint}^{\mathrm{phys}}(\e) = s + Bp$, $r_{\ind}^{\mathrm{phys}}(\e) = s + p$, and $G_C^{\mathrm{phys}}(\e) = \frac{B(s+p)}{s+Bp}$ for all $0 \le \e < \frac{8}{9}\e_{\mathrm{coll}}$, attained within the uniform linear suffix horizon
\begin{equation}\label{eq:c3H}
  H_* = \left\lceil (L - 1)\bigl(e^{Q_{\max}\max\{a, b\}} + 1\bigr) \right\rceil = O(s + p).
\end{equation}
\end{enumerate}
\end{theorem}
\begin{proof}
\emph{Decoupled probe denominators.} By orthonormality of $\{e_u, f_w\}$ and $e^{Q_{\max} K_0} = L - 1$, the historical denominator of session $i$ at each shared and private probe decouples completely:
\begin{equation}\label{eq:c3decouple}
\begin{aligned}
  D_{S_i}(q_u^{\sh}) &= (L - 1)\bigl(e^{Q_{\max}\theta_u} + 1\bigr),\\
  D_{S_i}(q_w^{\priv}) &= (L - 1)\bigl(e^{Q_{\max}\phi_{i,w}} + 1\bigr).
\end{aligned}
\end{equation}

\emph{Part~(1) (Relaxed lower and upper bounds).} Consider the centrally symmetric parameter box
\begin{multline*}
  \mathcal{B} := \Bigl\{ z = (\theta, \phi_1,\ldots,\phi_B) \in \R^{s + Bp} :\\
  \max\!\Bigl\{\frac{\norm{\theta}_\infty}{a},\, \max_{1 \le i \le B}\frac{\norm{\phi_i}_\infty}{b}\Bigr\} \le 1 \Bigr\}
\end{multline*}
of dimension $m = s + Bp$. By Theorem~\ref{thm:m4}, each coordinate map $(\theta, \phi_i) \mapsto S_i(\theta, \phi_i)$ is continuous into $(\K, \dQ)$, so $\Psi: \mathcal{B} \to (\F_{\joint}^{\mathrm{hm}}, d_{\Q}^{(B)})$, $z \mapsto \mathbf{S}(z) = (S_1(\theta,\phi_1),\ldots,S_B(\theta,\phi_B))$, is continuous. For any continuous encoder $E: (\F_{\joint}^{\mathrm{hm}}, d_{\Q}^{(B)}) \to \R^R$ with $R < s + Bp$, Borsuk--Ulam yields an antipodal boundary collision $z, -z \in \partial\mathcal{B}$:
\begin{itemize}
\item If $\norm{\theta}_\infty = a$, pick $u$ with $|\theta_u| = a$. Evaluating session $i = 1$ at $q_u^{\sh}$ via \eqref{eq:c3decouple} gives
\[
  \left|\log\frac{D_{S_1(z)}(q_u^{\sh})}{D_{S_1(-z)}(q_u^{\sh})}\right|
  = \left|\log\frac{e^{Q_{\max} a} + 1}{e^{-Q_{\max} a} + 1}\right|
  = Q_{\max} a.
\]
\item Otherwise, $\norm{\phi_i}_\infty = b$ for some $i \in \{1,\ldots,B\}$; picking $w$ with $|\phi_{i,w}| = b$ gives $|\log(D_{S_i(z)}(q_w^{\priv})/D_{S_i(-z)}(q_w^{\priv}))| = Q_{\max} b$.
\end{itemize}
In either case, \eqref{eq:keydist} gives $d_{\Q}^{(B)}(\mathbf{S}(z), \mathbf{S}(-z)) \ge 2M\tanh(Q_{\max}\min\{a, b\}/4) = 2\e_{\mathrm{coll}}$, so any collection of decoders $\{G_i\}_{i=1}^B$ (even if every $G_i$ reads all $R$ coordinates) incurs worst-session error at least $\e_{\mathrm{coll}}$ on one of the two inputs.

For the matching upper bound, \eqref{eq:c3decouple} uniquely inverts each parameter coordinate via
\begin{align*}
  \theta_u &= Q_{\max}^{-1}\log\!\left(\frac{D_{S_1}(q_u^{\sh})}{L - 1} - 1\right),\\
  \phi_{i,w} &= Q_{\max}^{-1}\log\!\left(\frac{D_{S_i}(q_w^{\priv})}{L - 1} - 1\right)
\end{align*}
for $1 \le u \le s$, $1 \le i \le B$, and $1 \le w \le p$, making $\Psi: \mathcal{B} \to (\F_{\joint}^{\mathrm{hm}}, d_{\Q}^{(B)})$ a continuous bijection from a compact space onto a Hausdorff metric space, hence a homeomorphism. The shared-plus-private encoder $E(\mathbf{S}) = \Psi^{-1}(\mathbf{S}) = (\theta, \phi_1,\ldots,\phi_B) \in \R^{s + Bp}$ is continuous in $d_{\Q}^{(B)}$, and each decoder $G_i(z_{\sh}, z_i)$ clips $(z_{\sh}, z_i)$ coordinatewise into $[-a, a]^s \times [-b, b]^p$ and outputs $S_i$ exactly across all $q \in \Q$, yielding $w_{s + Bp}^{(B)}(\F_{\joint}^{\mathrm{hm}}; \Q) = 0$ and proving \eqref{eq:c3rel}. Applying the identical argument with $B = 1$ to each marginal $\F_i$ gives $r_{\ind}(\e) = s + p$ and \eqref{eq:c3gain} for all $0 \le \e < \e_{\mathrm{coll}}$.

\emph{Part~(2) (Physical-continuation survival).} Because $\dQphys \le \dQ$, we have $d_{\Q}^{\mathrm{phys},(B)} \le d_{\Q}^{(B)}$, so $\Psi: \mathcal{B} \to (\F_{\joint}^{\mathrm{hm}}, d_{\Q}^{\mathrm{phys},(B)})$ is continuous. For any $d_{\Q}^{\mathrm{phys},(B)}$-continuous $E: \F_{\joint}^{\mathrm{hm}} \to \R^R$ ($R < s + Bp$), Borsuk--Ulam yields an antipodal collision $z, -z \in \partial\mathcal{B}$. At the maximizing session $i$ and probe $q \in \{q_u^{\sh}, q_w^{\priv}\}$ above, \eqref{eq:c3decouple} guarantees $D_{S_i}(q) > L - 1 \ge 1$ uniformly on $\mathcal{B}$ and $D_{S_i(z)}(q) \ne D_{S_i(-z)}(q)$, so $t^* = \sqrt{D_{S_i(z)}(q) D_{S_i(-z)}(q)} > 1$. Moreover, $D_{S_i}(q) \le (L - 1)(e^{Q_{\max}\max\{a,b\}} + 1)$ uniformly on $\mathcal{B}$, so $n = \lceil t^* \rceil \le H_*$ in \eqref{eq:c3H}. Applying \eqref{eq:eightninths} from Corollary~\ref{cor:w4phys} gives $d_{\Q}^{\mathrm{phys},(B)}(\mathbf{S}(z), \mathbf{S}(-z)) \ge \frac{16}{9}\e_{\mathrm{coll}}$, establishing the lower bound in \eqref{eq:c3phys}.

Finally, if $d_{\Q}^{\mathrm{phys},(B)}(\mathbf{S}(z), \mathbf{S}(\hat z)) = 0$, evaluating $H_{1,-M}$ at $n = 1$ across all sessions and probes forces $D_{S_i(z)}(q) = D_{S_i(\hat z)}(q)$ and hence $z = \hat z$, showing $d_{\Q}^{\mathrm{phys},(B)}$ is a metric on $\F_{\joint}^{\mathrm{hm}}$. Compactness of $\mathcal{B}$ makes $\Psi: \mathcal{B} \to (\F_{\joint}^{\mathrm{hm}}, d_{\Q}^{\mathrm{phys},(B)})$ a homeomorphism, so the same shared-plus-private encoder and decoders achieve zero error across all $q \in \Q$, yielding $w_{s + Bp}^{\mathrm{phys},(B)}(\F_{\joint}^{\mathrm{hm}}; \Q) = 0$ and matching physical collective dimensions (with $B = 1$ for the marginals) for all $0 \le \e < \frac{8}{9}\e_{\mathrm{coll}}$.
\end{proof}

\section{Indefinite-Horizon Certified Maintenance}
\label{sec:maintenance}

\subsection{Cone-Preserving Kernel Approximation}
\label{sec:p1a}

Fix a compact key domain $\mathcal{X} \subset \R^{d_k}$ and a continuous strictly positive kernel $K: \Q \times \mathcal{X} \to (0,\infty)$. A finite positive weighted atomic measure $\mu = \sum_{j=1}^L a_j \delta_{(k_j, v_j)}$ ($a_j > 0$, $k_j \in \mathcal{X}$, $\norm{v_j} \le M$) induces $S(\mu) = (N_\mu, D_\mu) \in \K$ via $N_\mu(q) = \sum_j a_j v_j K(q,k_j)$ and $D_\mu(q) = \sum_j a_j K(q,k_j)$.

\begin{lemma}[Sharp relative-error threshold and probability distortion]\label{lem:p1a}
Let $\hat K: \Q \times \mathcal{X} \to \R$ be a continuous approximate kernel satisfying
\begin{equation}\label{eq:relkernel}
  |\hat K(q,k) - K(q,k)| \le \gamma K(q,k)\qquad \forall (q,k) \in \Q \times \mathcal{X}
\end{equation}
for $0 \le \gamma < 1$. Replacing $K$ consistently by $\hat K$ in both numerator and denominator of any positive weighted atomic state $S = S(\mu)$ yields $\hat S \in \K$ satisfying
\begin{align}
  \tau(S, \hat S) &\le -\log(1 - \gamma), \label{eq:kernelcert}\\
  \dQ(S, \hat S) &\le 2M \tanh\!\left(\frac{1}{4}\log\frac{1+\gamma}{1-\gamma}\right) = \frac{2M\gamma}{1 + \sqrt{1 - \gamma^2}}. \label{eq:sharp}
\end{align}
Moreover, $\gamma < 1$ is the sharp universal admissibility threshold from \eqref{eq:relkernel} alone, and \eqref{eq:sharp} is sharp over the class of admissible two-atom weighted histories whenever $|\mathcal{X}| \ge 2$.
\end{lemma}
\begin{proof}
For $\gamma = 0$, $\hat K \equiv K$ gives $\hat S = S$ and $\tau = \dQ = 0$, matching \eqref{eq:kernelcert}--\eqref{eq:sharp} with equality. Now fix $0 < \gamma < 1$ and put $l = 1-\gamma > 0$, $h = 1+\gamma$. Since $l K \le \hat K \le h K$, $\hat K > 0$, $\hat D > 0$, and $\norm{\hat N} \le M\hat D$, so $\hat S \in \K$. Writing
\[
  P_S(q,u) = \sum_{j=1}^L a_j K(q,k_j)\left(1 + \frac{\ip{u}{v_j}}{2M}\right),
\]
every factor $1 + \ip{u}{v_j}/(2M) \ge 1/2$ is positive, so $l P_S \le P_{\hat S} \le h P_S$; since $h \le l^{-1}$, taking logarithms yields \eqref{eq:kernelcert}. At $\gamma = 1$, $\hat K \equiv 0$ satisfies \eqref{eq:relkernel} yet violates $\hat D > 0$, proving sharpness of $\gamma < 1$.

For \eqref{eq:sharp}, fix $q \in \Q$ and a relaxed suffix $(t, g)$ ($t \ge 0$, $\norm{g} \le M$). Form the common positive support $I$ as follows: if $t = 0$, omit the zero-mass suffix atom and set $I = \{1,\ldots,L\}$; if $t > 0$, adjoin atom $j = 0$ with positive weight $w_0 = t > 0$, value $v_0 = g$, and ratio $r_0 = 1 \in [l, h]$, setting $I = \{0, 1, \ldots, L\}$. For $j \ge 1$, put $w_j = a_j K(q,k_j) > 0$ and $r_j = \hat K(q,k_j)/K(q,k_j) \in [l, h]$. With strictly positive probability weights $p_j = w_j / \sum_{s \in I} w_s > 0$ ($j \in I$) and $m_0 = \sum_{j \in I} p_j r_j \in [l, h]$, the perturbed weights $\hat p_j = p_j r_j / m_0 > 0$ ($j \in I$) satisfy $\norm{p - \hat p}_1 = \mathbb{E}_p[|r - m_0|]/m_0$. Convexity of $r \mapsto |r - m_0|$ on $[l, h]$ gives
\[
  |r - m_0| \le \frac{h - r}{h - l}(m_0 - l) + \frac{r - l}{h - l}(h - m_0),
\]
and taking expectations under $p$ yields
\begin{align*}
  \norm{p - \hat p}_1
  &\le \frac{2(h - m_0)(m_0 - l)}{m_0(h - l)}\\
  &= \frac{2}{h - l}\!\left(h + l - m_0 - \frac{h l}{m_0}\right).
\end{align*}
By AM--GM, $m_0 + hl/m_0 \ge 2\sqrt{hl}$, attained at $m_0 = \sqrt{hl} \in [l, h]$, giving
\begin{align*}
  \norm{p - \hat p}_1
  &\le 2\frac{\sqrt{h} - \sqrt{l}}{\sqrt{h} + \sqrt{l}}
  = \frac{2\gamma}{1 + \sqrt{1 - \gamma^2}}\\
  &= 2\tanh\!\left(\frac{1}{4}\log\frac{1+\gamma}{1-\gamma}\right).
\end{align*}
Since $\norm{v_j} \le M$ for all $j \in I$,
\begin{align*}
  \norm{H_{t,g}(S;q) - H_{t,g}(\hat S;q)}
  &= \norm{\sum_{j \in I}(p_j - \hat p_j) v_j}\\
  &\le M \norm{p - \hat p}_1,
\end{align*}
which yields \eqref{eq:sharp} upon taking suprema over $t, g, q$.

\emph{Two-atom sharpness extremizer.} Given two distinct keys $k_l, k_h \in \mathcal{X}$, fix any $q_0 \in \Q$ and define a continuous function $r: \mathcal{X} \to [l, h]$ with $r(k_l) = l$ and $r(k_h) = h$, setting $\hat K(q,k) = r(k) K(q,k)$ (which satisfies \eqref{eq:relkernel} uniformly). Choose normalized weights
\[
  p_l = \frac{\sqrt{h}}{\sqrt{h} + \sqrt{l}},\qquad
  p_h = \frac{\sqrt{l}}{\sqrt{h} + \sqrt{l}},
\]
which satisfy $p_l + p_h = 1$ and $p_l l + p_h h = \sqrt{lh}$, and set positive atom weights $a_l = p_l / K(q_0, k_l)$ and $a_h = p_h / K(q_0, k_h)$. Assign opposite values $v_l = M u_0$ and $v_h = -M u_0$ for any unit vector $\norm{u_0} = 1$. Then at $q_0$, $\hat p_l = p_l l / \sqrt{lh} = p_h$ and $\hat p_h = p_h h / \sqrt{lh} = p_l$, so $p_l - \hat p_l = -(p_h - \hat p_h) = \frac{\sqrt{h}-\sqrt{l}}{\sqrt{h}+\sqrt{l}}$, and the zero-suffix ($t=0$) output error is
\begin{align*}
  \norm{F_S(q_0) - F_{\hat S}(q_0)}
  &= \norm{(p_l - \hat p_l)M u_0 + (p_h - \hat p_h)(-M u_0)}\\
  &= 2M\frac{\sqrt{h} - \sqrt{l}}{\sqrt{h} + \sqrt{l}},
\end{align*}
attaining equality in \eqref{eq:sharp}.
\end{proof}

The underlying probability inequality is an instance of Cohen and Fausti~\cite[Theorem~5.1]{cohenfausti}: restricting to the common positive support $I$ (omitting atom $j=0$ when $t=0$), $p$ and $\hat p$ are strictly positive probability vectors whose Hilbert projective distance obeys $\mathcal{H}(p,\hat p) = \log\bigl((\max_{j\in I}\hat p_j/p_j)/(\min_{j\in I}\hat p_j/p_j)\bigr) \le \log(h/l) = \log\frac{1+\gamma}{1-\gamma}$, yielding $\norm{p - \hat p}_1 \le 2\tanh\bigl(\mathcal{H}(p,\hat p)/4\bigr)$. Notice also that \eqref{eq:sharp} is roughly twice as tight as inserting \eqref{eq:kernelcert} into Theorem~\ref{thm:m4} for small $\gamma$ ($M\gamma$ vs.\ $2M\gamma$), and requires applying $\hat K$ consistently in both numerator and denominator.

\parhead{Finite-feature examples.}
On a circular plane with $K(\theta,\phi) = e^{\kappa\cos(\theta-\phi)}$, truncating the Fourier--Bessel series at order $H$,
\begin{equation}\label{eq:harmonic}
  K_H(\theta,\phi) = I_0(\kappa) + 2\sum_{n=1}^H I_n(\kappa)\cos\bigl(n(\theta-\phi)\bigr),
\end{equation}
yields admissibility whenever $\gamma_H = 2e^\kappa \sum_{n>H} I_n(\kappa) < 1$, with explicit real key-feature basis
\begin{equation}\label{eq:realfourier}
  \bigl\{1,\, \cos\phi,\, \sin\phi,\, \ldots,\, \cos(H\phi),\, \sin(H\phi)\bigr\}
\end{equation}
of real dimension $m = 2H + 1$. On any bounded dot-product domain ($|q^\top k| \le \rho$), degree-$p_0$ Taylor truncation achieves $\gamma \le e^{2\rho}\rho^{p_0+1}/(p_0+1)!$ with $\binom{d_k+p_0}{p_0}$ real monomial features including $1$; hence finite admissible feature expansions always exist on bounded domains, though their dimension can grow rapidly with $d_k$ and $\rho$.

\subsection{Depth-Controlled Dyadic Merge-and-Reduce}
\label{sec:d1}

Using Theorem~\ref{thm:m3}, a merge-and-reduce stream controls approximation error by tree \emph{depth}~\cite{bentley}. Below, Part~1 holds for any strictly-increasing-level merge forest, whereas Part~2 imposes binary-counter scheduling.

\begin{theorem}[Dynamic error and binary-counter resource bounds]\label{thm:d1}
Suppose each level-$\ell$ reduction $\mathcal{R}_\ell$ ($\ell \ge 1$) maps every merged summary it receives to a nonempty admissible state in $\K$ with $\tau(X, \mathcal{R}_\ell(X)) \le \eta_\ell$.
\begin{enumerate}
\item \emph{Depth-controlled error (general strictly-increasing merge forest):} Consider any binary merge-and-reduce forest over exact leaves in which compression levels strictly increase along every leaf-to-root path up to maximum active level $K_T \ge 0$. Let $S_T$ be the exact union of all $T$ tokens and $\hat S_T$ the merge of all active roots and any exact unreduced tokens. Then
\begin{equation}\label{eq:d1}
  \tau(S_T, \hat S_T) \le \sum_{\ell=1}^{K_T} \eta_\ell
\end{equation}
(with an empty sum equal to $0$). For any target $0 < \e < 2M$ with $\tau_* = 2\operatorname{artanh}(\e/(4M))$, the horizon-free schedule $\eta_\ell = 6\tau_*/(\pi^2 \ell^2)$ guarantees $\dQ(S_T, \hat S_T) \le \e$ for all $T \ge 1$.
\item \emph{Logarithmic levels and conditional resource bounds (binary-counter schedule):} Suppose tokens arrive into an incomplete exact insertion buffer of size $< c$ ($c \ge 1$, storage cost at most $R_{\mathrm{buffer}}$, with $R_{\mathrm{buffer}} = 0$ when $c = 1$). Whenever $c$ tokens accumulate, they form a complete exact level-$0$ leaf block (storage cost at most $C_0$). Complete blocks are maintained by a binary counter: after completed carries, at most one active exact level-$0$ block and at most one active reduced summary per level $\ell \ge 1$ exist simultaneously; whenever two blocks occupy level $\ell - 1$, they are merged and reduced via $\mathcal{R}_\ell$ into level $\ell$ (cascading carries upward). Then $K_T = 0$ for $T < 2c$ and $K_T = \lfloor \log_2 \lfloor T/c \rfloor \rfloor$ for $T \ge 2c$. If $C_\ell(\eta_\ell)$ bounds the persistent representation size of a level-$\ell$ summary and $W_\ell$ bounds the computational work of one level-$\ell$ merge-and-reduce call, then at every step $T \ge 1$ the persistent active storage $R_T$ after completed carries (excluding temporary merge workspace) and amortized reduction work per token $U_{\mathrm{amortized}}$ (beyond buffer insertion) satisfy
\begin{equation}\label{eq:d1resources}
\begin{aligned}
  R_T &\le R_{\mathrm{buffer}} + C_0 + \sum_{\ell=1}^{K_T} C_\ell(\eta_\ell),\\
  U_{\mathrm{amortized}} &= O\!\left(\sum_{\ell=1}^{K_T} \frac{W_\ell}{2^\ell c}\right).
\end{aligned}
\end{equation}
\end{enumerate}
\end{theorem}
\begin{proof}
For (1), at an internal level-$\ell$ node merging exact children $S_1, S_2$ approximated by $\hat S_1, \hat S_2$, the triangle inequality and Theorem~\ref{thm:m3} yield
\begin{align*}
  \tau\bigl(S_1+S_2,\, &\mathcal{R}_\ell(\hat S_1+\hat S_2)\bigr) \\
  &\le \tau(S_1+S_2, \hat S_1+\hat S_2) + \eta_\ell \\
  &\le \max\{\tau(S_1,\hat S_1), \tau(S_2,\hat S_2)\} + \eta_\ell.
\end{align*}
Induction from exact leaves bounds each active root's error by the maximum path sum of level budgets; merging all active roots and exact leaves via Theorem~\ref{thm:m3} gives \eqref{eq:d1}. Since $\sum_{\ell=1}^\infty \eta_\ell = \tau_*$, Theorem~\ref{thm:m4} yields $\dQ(S_T, \hat S_T) \le \e$.

For (2), write $T = q c + r$ with $q = \lfloor T/c \rfloor$ complete leaf blocks and $r = T \bmod c \in \{0,\ldots,c-1\}$ tokens in the incomplete insertion buffer. After completed carries, the binary expansion $q = \sum_{\ell \ge 0} b_\ell 2^\ell$ ($b_\ell \in \{0,1\}$) records exactly which levels hold a persistent active block: $b_0 = 1$ denotes an active full exact level-$0$ block (cost $\le C_0$), and $b_\ell = 1$ ($\ell \ge 1$) denotes an active level-$\ell$ summary (cost $\le C_\ell(\eta_\ell)$). Hence $K_T = 0$ when $q \le 1$ ($T < 2c$) and $K_T = \lfloor \log_2 q \rfloor$ when $q \ge 2$ ($T \ge 2c$). Summing persistent active components gives the bound on $R_T$ in \eqref{eq:d1resources} (excluding temporary merge workspace during carries; Appendix~\ref{app:d1} verifies boundary horizons explicitly for $c = 1$ and $c \ge 2$). Finally, a level-$\ell$ reduction ($\ell \ge 1$) consumes $2^\ell c$ original tokens and executes exactly $\lfloor T / (2^\ell c) \rfloor$ times during the first $T$ tokens, yielding total reduction work $\sum_{\ell=1}^{K_T} \lfloor T/(2^\ell c)\rfloor W_\ell$ and the amortized bound in \eqref{eq:d1resources}.
\end{proof}

Each token participates in at most $O(\log(1+T/c))$ reductions, but participation count must not be equated with computational runtime without specifying the reduction cost $W_\ell$. Table~\ref{tab:claims} delineates the exact scope of these dynamic maintenance guarantees.

\begin{table}[!t]
\centering
\caption{Exact Scope of the Dynamic Maintenance Guarantees}
\label{tab:claims}
\setlength{\tabcolsep}{3.0pt}
\renewcommand{\arraystretch}{1.12}
\begin{tabular}{@{}L{38mm} L{\dimexpr\columnwidth-38mm-2\tabcolsep\relax}@{}}
\toprule
\textbf{Property} & \textbf{Established here?} \\
\midrule
Error independent of stream length $T$ & Yes in exact arithmetic (Thm.~\ref{thm:d1}, Cor.~\ref{cor:fixed}) \\
Numerical solver drift & Yes, $O(\delta_{\mathrm{num}}\log T)$ in dyadic merge forest (Cor.~\ref{cor:numdrift}) \\
Atom count independent of $T$ & Yes (fixed-feature positive summaries, Thm.~\ref{thm:p1b}) \\
Constant bit complexity & No, but $O(\log\log T)$ weight bits via relative-precision quantization (Cor.~\ref{cor:bitwidth}; excl.\ KV coords.) \\
Efficient amortized update time & Conditional on reduction cost $W_\ell$ \\
GPU inference speedup & Not established \\
\bottomrule
\end{tabular}
\end{table}

\subsection{Positive Moment-Preserving Recompression}
\label{sec:p1b}

To construct a positive atomic class closed under repeated merges and recompressions, we apply Carath\'eodory--Tchakaloff positive cubature~\cite{bayer,piazzon2017} directly on weighted atomic measures. For $\mu = \sum_{j=1}^n a_j \delta_{(k_j,v_j)}$ ($a_j > 0$, $k_j \in \mathcal{X}$, $\norm{v_j} \le M$) and continuous real key features $g_1,\ldots,g_m: \mathcal{X} \to \R$, define $\psi(k,v) = \bigl(g_\ell(k),\, g_\ell(k)v\bigr)_{\ell=1}^m \in \R^{m(d_v+1)}$ and the additive moment vector
\begin{equation}\label{eq:moment}
  \Psi(\mu) = \int \psi(k,v)\,d\mu(k,v) = \sum_{j=1}^n a_j \psi(k_j, v_j) \in \R^{m(d_v+1)}.
\end{equation}
Defining $\Psi$ on $\mu$ rather than on an abstract pair $(N,D) \in \K$ avoids assuming that $(N,D)$ uniquely determines all chosen moments for an arbitrary kernel.

\begin{theorem}[Fixed-moment positive realization]\label{thm:p1b}
Suppose a continuous positive kernel $K: \Q \times \mathcal{X} \to (0,\infty)$ admits a real separable approximation
\begin{equation}\label{eq:features}
  \hat K(q,k) = \sum_{\ell=1}^m f_\ell(q) g_\ell(k),\qquad
  |\hat K - K| \le \gamma K,\quad 0 \le \gamma < 1,
\end{equation}
uniformly on $\Q \times \mathcal{X}$, with $1 \in \operatorname{span}\{g_1,\ldots,g_m\}$ (adjoining $g \equiv 1$ into $m$ if absent). Then for every finite positive weighted atomic measure $\mu$ on $\mathcal{X} \times \{\norm{v}\le M\}$, there exists a positive weighted atomic measure $\tilde\mu$ supported on at most $m(d_v+1)$ original atoms of $\mu$ such that $\Psi(\tilde\mu) = \Psi(\mu)$, its induced state $S(\tilde\mu)$ (evaluated with the original kernel $K$) lies in $\K$, and
\begin{equation}\label{eq:p1b}
\begin{aligned}
  \tau\bigl(S(\mu), S(\tilde\mu)\bigr) &\le \log\frac{1+\gamma}{1-\gamma},\\
  \dQ\bigl(S(\mu), S(\tilde\mu)\bigr) &\le \min\{2M, 4M\gamma\}.
\end{aligned}
\end{equation}
\end{theorem}
\begin{proof}
Write $\mu = \sum_{j=1}^n a_j \delta_{(k_j,v_j)}$ ($a_j > 0$). While $n > m(d_v+1)$, the $n$ vectors $\psi(k_j, v_j) \in \R^{m(d_v+1)}$ are linearly dependent, so some nonzero $c \in \R^n$ satisfies $\sum_{j=1}^n c_j \psi(k_j, v_j) = 0$. Since $c \ne 0$ and $1 \in \operatorname{span}\{g_1,\ldots,g_m\}$, $\sum_{j=1}^n c_j = 0$, so $c$ has a strictly positive entry. Set $t = \min_{c_j > 0} a_j / c_j > 0$ and replace $a_j \leftarrow a_j - t c_j$: all new weights are nonnegative, at least one vanishes, and $\Psi$ is unchanged. Dropping zero-weight atoms and iterating terminates with $n \le m(d_v+1)$; the constant-feature assumption $1 \in \operatorname{span}\{g_1,\ldots,g_m\}$ establishes exact preservation of total atomic mass ($\sum \tilde a_j = \sum a_j > 0$), so $\tilde\mu$ is nonempty and $S(\tilde\mu) \in \K$.

Let $\hat P_\mu, \hat P_{\tilde\mu}$ be the lifted functions \eqref{eq:lift} evaluated with $\hat K$. Separability \eqref{eq:features} and $\Psi(\mu) = \Psi(\tilde\mu)$ give $\hat P_\mu = \hat P_{\tilde\mu}$ on $\Q \times B_*$. Applying Lemma~\ref{lem:p1a} to both $\mu$ and $\tilde\mu$ yields $(1-\gamma)P_{S(\mu)} \le \hat P_\mu = \hat P_{\tilde\mu} \le (1+\gamma)P_{S(\tilde\mu)}$ and symmetrically $(1-\gamma)P_{S(\tilde\mu)} \le (1+\gamma)P_{S(\mu)}$, proving the $\tau$ bound in \eqref{eq:p1b}. Theorem~\ref{thm:m4} and $\tanh\bigl(\tfrac{1}{2}\log\frac{1+\gamma}{1-\gamma}\bigr) = \gamma$ yield the $\dQ$ bound.
\end{proof}

\begin{corollary}[Indefinite-horizon fixed-feature maintenance]\label{cor:fixed}
Fix a single feature expansion \eqref{eq:features} with $1 \in \operatorname{span}\{g_1,\ldots,g_m\}$, and apply Theorem~\ref{thm:p1b} after every append or disjoint-history merge of weighted atomic measures. In exact arithmetic, for every $T \ge 1$ the maintained summary $\tilde\mu_T$ has at most $m(d_v+1)$ positive atoms, preserves total atomic mass exactly ($\sum \tilde a_j = \sum a_j$), and satisfies
\begin{equation}\label{eq:fixed}
\begin{aligned}
  \Psi(\tilde\mu_T) &= \Psi(\mu_T),\\
  \dQ\bigl(S(\mu_T), S(\tilde\mu_T)\bigr) &\le \min\{2M, 4M\gamma\},
\end{aligned}
\end{equation}
where $\mu_T$ is the exact unreduced history. Choosing $\gamma \le \e/(4M)$ guarantees uniform error $\le \e$ independently of the number of merges or reductions.
\end{corollary}
\begin{proof}
Additivity of $\Psi$ and exact preservation at each reduction step give $\Psi(\tilde\mu_T) = \Psi(\mu_T)$ by induction; Theorem~\ref{thm:p1b}'s ratio comparison directly between $\mu_T$ and $\tilde\mu_T$ removes depth-dependent accumulation entirely.
\end{proof}

In numerical implementations of Carath\'eodory--Tchakaloff subsampling (such as Lawson--Hanson NNLS or LP solvers), moment matching is not exact and stored positive weights require finite precision. The following two corollaries bridge Theorem~\ref{thm:p1b} and Corollary~\ref{cor:fixed} to approximate numerical reductions and quantized weight bit-widths.

\begin{corollary}[Approximate moment-preserving reduction and logarithmic numerical drift]\label{cor:numdrift}
In the setting of Theorem~\ref{thm:p1b}, let $\hat K(q, k) = \sum_{\ell=1}^m f_\ell(q) g_\ell(k)$ satisfy $|\hat K - K| \le \gamma K$ ($0 \le \gamma < 1$) on $\Q \times \mathcal{X}$, and let $\hat P_\mu(q, u)$ denote the positive-cone lift \eqref{eq:lift} evaluated with $\hat K$. Suppose a numerical reduction operator $\mathcal{R}$ maps any finite positive weighted atomic measure $\mu$ on $\mathcal{X} \times \{\norm{v} \le M\}$ to a positive weighted atomic measure $\tilde\mu = \mathcal{R}(\mu)$ supported on at most $m(d_v + 1)$ original atoms, satisfying the uniform relative surrogate-cone residual
\begin{equation}\label{eq:numres}
\begin{aligned}
  (1 - \delta_{\mathrm{num}}) \hat P_\mu(q, u)
  &\le \hat P_{\tilde\mu}(q, u)
  \le (1 + \delta_{\mathrm{num}}) \hat P_\mu(q, u)\\
  &\qquad\qquad \forall (q, u) \in \Q \times B_*
\end{aligned}
\end{equation}
for some $0 \le \delta_{\mathrm{num}} < 1$.
\begin{enumerate}
\item \emph{Single-step bound:} The induced state $S(\tilde\mu)$ lies in $\K$ and obeys
\begin{equation}\label{eq:numsingle}
  \tau\bigl(S(\mu), S(\tilde\mu)\bigr)
  \le \log\frac{1 + \gamma}{1 - \gamma} - \log(1 - \delta_{\mathrm{num}}).
\end{equation}
\item \emph{Depth-controlled streaming drift:} If $\mathcal{R}$ is used as the reduction operator at every internal level $\ell \ge 1$ of a strictly-increasing merge forest of maximum active level $K_T \ge 0$ (Theorem~\ref{thm:d1}(1)), then for every $T \ge 1$ the maintained summary $\tilde\mu_T$ satisfies
\begin{equation}\label{eq:numstream}
  \tau\bigl(S(\mu_T), S(\tilde\mu_T)\bigr)
  \le \log\frac{1 + \gamma}{1 - \gamma} - K_T \log(1 - \delta_{\mathrm{num}}),
\end{equation}
where $\mu_T$ is the exact unreduced history. In particular, under the binary-counter schedule of Theorem~\ref{thm:d1}(2), $K_T = \lfloor \log_2 \lfloor T/c \rfloor \rfloor$ ($T \ge 2c$), so numerical solver drift grows at most logarithmically in $T$ ($O(\delta_{\mathrm{num}}\log T)$ rather than $O(\delta_{\mathrm{num}} T)$) while the kernel approximation error remains independent of $T$.
\end{enumerate}
\end{corollary}
\begin{proof}
Because $0 \le \gamma < 1$, $\hat K$ is continuous and strictly positive on compact $\Q \times \mathcal{X}$; by Lemma~\ref{lem:p1a}, every finite positive weighted atomic measure $\mu$ induces an admissible surrogate state $\hat S(\mu) = (\hat N_\mu, \hat D_\mu) \in \K$ with lift $\hat P_\mu = P_{\hat S(\mu)}$. Define the surrogate Thompson pseudometric $\hat\tau(\mu, \nu) := \tau(\hat S(\mu), \hat S(\nu)) = \sup_{(q,u) \in \Q \times B_*} |\log(\hat P_\nu(q,u)/\hat P_\mu(q,u))|$. By linearity of $\mu \mapsto \hat S(\mu)$, $\hat\tau$ satisfies the triangle inequality and the maximum-error merge law (Theorem~\ref{thm:m3}). Because $\delta_{\mathrm{num}} < 1$ and $\hat P_\mu > 0$, \eqref{eq:numres} guarantees $\hat P_{\tilde\mu} > 0$, so $\tilde\mu$ is nonempty and $S(\tilde\mu) \in \K$; moreover, since $1 + \delta_{\mathrm{num}} \le (1 - \delta_{\mathrm{num}})^{-1}$, \eqref{eq:numres} implies $\hat\tau(\mu, \mathcal{R}(\mu)) \le -\log(1 - \delta_{\mathrm{num}})$.

Applying the induction of Theorem~\ref{thm:d1}(1) in $\hat\tau$ from exact leaves ($\hat\tau = 0$) yields $\hat\tau(\mu_T, \tilde\mu_T) \le -K_T \log(1 - \delta_{\mathrm{num}})$ (with $K_T = 1$ for Part~(1)). By Lemma~\ref{lem:p1a} applied to both $\mu_T$ and $\tilde\mu_T$, $(1-\gamma)P_{S(\mu_T)} \le \hat P_{\mu_T} \le (1+\gamma)P_{S(\mu_T)}$ and $(1-\gamma)P_{S(\tilde\mu_T)} \le \hat P_{\tilde\mu_T} \le (1+\gamma)P_{S(\tilde\mu_T)}$. Chaining these inequalities with $(1 - \delta_{\mathrm{num}})^{K_T} \hat P_{\mu_T} \le \hat P_{\tilde\mu_T} \le (1 - \delta_{\mathrm{num}})^{-K_T} \hat P_{\mu_T}$ yields
\begin{align*}
  \frac{1 - \gamma}{1 + \gamma}(1 - \delta_{\mathrm{num}})^{K_T} P_{S(\mu_T)}
  &\le P_{S(\tilde\mu_T)} \\
  &\le \frac{1 + \gamma}{1 - \gamma}(1 - \delta_{\mathrm{num}})^{-K_T} P_{S(\mu_T)}.
\end{align*}
Taking logarithms and suprema over $\Q \times B_*$ proves both \eqref{eq:numsingle} and \eqref{eq:numstream} (and transfers to $\dQ$ via \eqref{eq:m4}).
\end{proof}

By separability \eqref{eq:features} and compactness of $\Q \times \mathcal{X}$, any moment-vector solver residual $\norm{\Psi(\tilde\mu) - \Psi(\mu)}_2$ directly induces a uniform bound $\delta_{\mathrm{num}}$ in \eqref{eq:numres} proportional to $\norm{\Psi(\tilde\mu) - \Psi(\mu)}_2 / \sum a_j$.

\begin{corollary}[Quantized positive weights and $O(\log\log T)$ weight bit-width]\label{cor:bitwidth}
Let $\tilde\mu_T = \sum_{j=1}^{n_T} \tilde a_j \delta_{(k_j, v_j)}$ be a positive weighted atomic summary of a unit-weight stream of length $T \ge 1$ on $\mathcal{X} \times \{\norm{v} \le M\}$ with $1 \le n_T \le m(d_v + 1)$ atoms and preserved total atomic mass $\sum_{j=1}^{n_T} \tilde a_j = T$ (as in Corollary~\ref{cor:fixed}). Let $\kappa_K := \frac{\sup_{\Q \times \mathcal{X}} K}{\inf_{\Q \times \mathcal{X}} K} \in [1, \infty)$ be the finite kernel dynamic range on compact $\Q \times \mathcal{X}$. Fix truncation and rounding tolerances $\delta_{\mathrm{drop}}, \delta_w \in (0, 1)$, and form the quantized summary $\bar\mu_T$ by:
\begin{enumerate}
\item dropping any atom with weight $\tilde a_j < \frac{\delta_{\mathrm{drop}}}{3 \kappa_K n_T} T$, and
\item rounding each surviving weight $\tilde a_j$ to $\bar a_j \in [(1 - \delta_w)\tilde a_j,\, (1 + \delta_w)\tilde a_j]$.
\end{enumerate}
Then $\bar\mu_T$ is nonempty, $S(\bar\mu_T) \in \K$, and
\begin{equation}\label{eq:quanttau}
  \tau\bigl(S(\tilde\mu_T), S(\bar\mu_T)\bigr)
  \le -\log(1 - \delta_{\mathrm{drop}}) - \log(1 - \delta_w).
\end{equation}
Moreover, storing a shared scale exponent $\lfloor \log_2 T \rfloor$ alongside relative floating-point representations of the surviving weights requires total weight bit-width
\begin{equation}\label{eq:bitwidth}
\begin{aligned}
  O\Bigl(&\log\log(T + 3) + m(d_v + 1)\log\frac{2}{\delta_w} \\
  &+ m(d_v + 1)\log\log\frac{3\kappa_K m(d_v + 1)}{\delta_{\mathrm{drop}}}\Bigr)
\end{aligned}
\end{equation}
bits (excluding key--value coordinates), scaling as $O(\log\log T)$ in stream length $T$.
\end{corollary}
\begin{proof}
Let $K_{\min} = \inf_{\Q \times \mathcal{X}} K > 0$ and $K_{\max} = \sup_{\Q \times \mathcal{X}} K = \kappa_K K_{\min}$. By \eqref{eq:liftbounds},
\[
  P_{S(\tilde\mu_T)}(q, u)
  \ge \frac{1}{2} D_{\tilde\mu_T}(q)
  \ge \frac{1}{2}\!\left(\sum_{j=1}^{n_T} \tilde a_j\right) K_{\min}
  = \frac{1}{2} T K_{\min}
\]
uniformly on $\Q \times B_*$. Meanwhile, each atom contributes at most $\frac{3}{2} \tilde a_j K_{\max}$ to $P_{S(\tilde\mu_T)}(q, u)$, so the total contribution of all dropped atoms is at most
\[
  n_T \cdot \frac{\delta_{\mathrm{drop}} T}{3 \kappa_K n_T} \cdot \frac{3}{2} K_{\max}
  = \frac{1}{2}\delta_{\mathrm{drop}} T K_{\min}
  \le \delta_{\mathrm{drop}} P_{S(\tilde\mu_T)}(q, u).
\]
Thus the post-truncation measure $\tilde\mu_T'$ is nonempty and obeys $(1 - \delta_{\mathrm{drop}}) P_{S(\tilde\mu_T)} \le P_{S(\tilde\mu_T')} \le P_{S(\tilde\mu_T)}$. Rounding each surviving positive weight within relative error $\delta_w$ preserves positivity and yields $(1 - \delta_{\mathrm{drop}})(1 - \delta_w) P_{S(\tilde\mu_T)} \le P_{S(\bar\mu_T)} \le (1 + \delta_w) P_{S(\tilde\mu_T)}$; since $1 + \delta_w \le (1 - \delta_w)^{-1}$, taking logarithms yields \eqref{eq:quanttau}.

Finally, let $E_T = \lfloor \log_2 T \rfloor \ge 0$, so $1 \le T / 2^{E_T} < 2$. Every surviving weight lies in $\bigl[\frac{\delta_{\mathrm{drop}}}{3\kappa_K n_T} T, T\bigr]$, so $\tilde a_j / 2^{E_T} \in \bigl[\frac{\delta_{\mathrm{drop}}}{3\kappa_K m(d_v+1)}, 2\bigr)$. Writing $\tilde a_j / 2^{E_T} = 2^{e_j}(1 + m_j)$ with $m_j \in [0, 1)$ and rounding $1 + m_j$ to $b_m = \lceil \log_2(1/\delta_w)\rceil$ fractional mantissa bits yields relative rounding error at most $2^{-(b_m+1)} < \delta_w$ and rounded exponent $e_j \in \{-E_{\mathrm{rel}},\ldots,1\}$, where $E_{\mathrm{rel}} = \lceil \log_2(3\kappa_K m(d_v+1)/\delta_{\mathrm{drop}})\rceil$ (incrementing $e_j$ by $1$ if rounding reaches $2$). Encoding the shared integer $E_T$ via a prefix-free code takes $2\lfloor \log_2(E_T + 1) \rfloor + 1 = O(\log\log(T+3))$ bits, and each of the at most $m(d_v + 1)$ surviving weights requires $\lceil \log_2(E_{\mathrm{rel}} + 2)\rceil = O(\log\log(3\kappa_K m(d_v+1)/\delta_{\mathrm{drop}}))$ exponent bits plus $b_m = O(\log(2/\delta_w))$ mantissa bits, giving \eqref{eq:bitwidth}.
\end{proof}

\parhead{Representation and finite-precision caveats.}
For the circular harmonic kernel \eqref{eq:harmonic} with real basis \eqref{eq:realfourier}, $m = 2H + 1$, giving at most $(2H+1)(d_v+1)$ atoms independent of $T$. While maintaining $\Psi(\mu_T) \in \R^{m(d_v+1)}$ directly recovers classical linear/feature attention accumulators~\cite{linear,performer}, Theorem~\ref{thm:p1b} remains inside the positive atomic class for exact-$K$ evaluation and merging. Four caveats delineate the exact boundary of these maintenance guarantees: (i)~zero accumulation holds in exact arithmetic (Corollary~\ref{cor:fixed}), whereas approximate numerical reductions in a dyadic merge tree incur at most logarithmic drift $O(\delta_{\mathrm{num}}\log T)$ while keeping kernel error $T$-independent (Corollary~\ref{cor:numdrift}); (ii)~the positive moment-elimination construction does not establish the existence of a continuous atom-selection encoder; (iii)~although exact weights grow linearly with $T$, relative-precision weight quantization reduces weight storage to $O(\log\log T)$ bits in $T$ (Corollary~\ref{cor:bitwidth}), leaving finite-bit key--value coordinate compression separate; and (iv)~changing $\Q$, $K$, $M$, or the feature set mid-stream breaks zero-accumulation maintenance.

\section{Related Work, Novelty Positioning, and Scope}
\label{sec:scope}

\begin{table*}[!t]
\centering
\caption{Positioning Relative to Key Mathematical and Systems Prior Art}
\label{tab:related}
\setlength{\tabcolsep}{4.0pt}
\renewcommand{\arraystretch}{1.12}
\begin{tabular}{@{}L{42mm} L{53mm} L{\dimexpr\textwidth-95mm-4\tabcolsep\relax}@{}}
\toprule
\textbf{Prior work} & \textbf{What it establishes} & \textbf{Relation and distinction in this paper} \\
\midrule
Yang and Li~\cite{yang2026} & Fixed-row tail factorization (Thm.~3.7), a posteriori future-query bound with $|p-q|\le\tanh(|\Lambda_T-z_\star|/4)$ (Thm.~C.1), linear-sketch rank floor (Thm.~C.2). & Shares the $\tanh(\delta/4)$ log-mass sensitivity envelope~\cite[Thm.~C.1]{yang2026}; we extend beyond single-row and linear-sketch bounds via dynamic state equivalence ($\dQ$), continuous nonlinear hidden-mass widths (Thms.~\ref{thm:w4}, \ref{thm:w4orth}; Cor.~\ref{cor:w4phys}), collective widths (Thms.~\ref{thm:c2}, \ref{thm:c3}), and positive-cone maintenance. \\
Liberty--Andoni--Kleiner~\cite{liberty}; Cohen~\cite{cohen} & Query-oblivious attention coresets controlling $(N,D)$ via sampling and recursive halving. & Closest coreset prior art; Corollary~\ref{cor:coreset_inst} distinguishes exact mass restoration (sampling) from certified approximate dyadic rescaling (halving) and clarifies gauge consistency for physical appends. \\
FlashInfer~\cite{cascade,flashinfer}; Hydragen~\cite{hydragen} & Merges output and log-normalizer statistics for shared-prefix execution. & State-pair merging is established; we formalize its append-observational metric geometry and continuous nonlinear widths. \\
Thompson~\cite{thompson}; Cohen--Fausti~\cite{cohenfausti} & Positive-cone Thompson metric; sharp Hilbert--TV bound $\norm{\mu-\nu}_1\le 2\tanh(\mathcal{H}/4)$. & Theorems~\ref{thm:m3}--\ref{thm:m4} apply classical Thompson geometry via $P_S$; Lemma~\ref{lem:p1a} adapts the Hilbert--TV bound on common positive support. \\
DeVore--Howard--Micchelli~\cite{devore}; Brouwer~\cite{brouwer}; Borsuk~\cite{borsuk} & Continuous nonlinear widths; invariance of domain; Borsuk--Ulam theorem. & Theorem~\ref{thm:w3} is a classical ellipsoid-width reference; our primary width results are Theorems~\ref{thm:w4}, \ref{thm:w4orth}, \ref{thm:c3} and Corollary~\ref{cor:w4phys} under hidden historical mass. \\
Carath\'eodory--Tchakaloff~\cite{bayer,piazzon2017}; Bentley--Saxe~\cite{bentley} & Positive moment-preserving cubature; static-to-dynamic merge trees. & Established primitives embedded into $(\tau,\dQ)$; Theorem~\ref{thm:p1b} preserves total atomic mass exactly under $1\in\operatorname{span}\{g_\ell\}$, extended to $O(\delta_{\mathrm{num}}\log T)$ numerical drift and $O(\log\log T)$ weight bits (Cors.~\ref{cor:numdrift}--\ref{cor:bitwidth}). \\
Haris--Onak~\cite{haris}; BalanceKV~\cite{balancekv}; Chen et al.~\cite{tightstream} & Finite-bit streaming space lower bounds and discrepancy summaries. & Complementary models (Table~\ref{tab:width_vs_bits}): we bound continuous real coordinates under future appends, alongside an $O(\log\log T)$ weight bit-width upper bound (Cor.~\ref{cor:bitwidth}, excl.\ KV coords.). \\
Kim--Son~\cite{tworead}; Haverbeck et al.~\cite{risk}; Konda~\cite{konda} & Two-read linear cache lower bounds; distributional risk; deep parameter identifiability. & We place no read-count restriction on decoders, study worst-case appends, and identify single-head \emph{history} states. \\
\bottomrule
\end{tabular}
\end{table*}

\subsection{Relationship to Query-Oblivious Attention Coresets}
\label{sec:coresets}

Query-oblivious attention coreset constructions~\cite{liberty,cohen} control numerator and denominator errors jointly inside their analyses.

To see where append observability adds structure, consider an unweighted subset retaining one-tenth of a collection of identical historical KV pairs with constant value $v = M$. Its current normalized output is exact ($F_S = F_{\hat S} \equiv M$), yet its historical mass drops by a factor of $10$, yielding relaxed append distance (Fig.~\ref{fig:validation}(a))
\begin{equation}\label{eq:tenx}
\begin{aligned}
  \dapp(S, \hat S; q)
  &= 2M\tanh\!\left(\frac{\log 10}{4}\right)\\
  &= 2M\frac{\sqrt{10}-1}{\sqrt{10}+1} \approx 1.0390\,M.
\end{aligned}
\end{equation}
Thus an unweighted coreset---or any theorem whose black-box guarantee bounds only normalized output $\norm{F_S - F_{\hat S}}$---does not automatically control append-observational error.

Conversely, our metric inequalities certify both sufficiency and necessity between append-observational fidelity and joint preservation of normalized responses and relative historical mass:

\begin{proposition}[Weighted coreset guarantees imply append robustness]\label{prop:coreset}
Suppose a weighted summary $\hat S = (\hat N, \hat D) \in \K$ approximates $S = (N, D) \in \K$ uniformly on $\Q$ with relative denominator error $(1-\gamma)D(q) \le \hat D(q) \le (1+\gamma)D(q)$ ($0 \le \gamma < 1$) and either (a) normalized output error $\norm{F_S(q) - F_{\hat S}(q)} \le \e_F$, or (b) numerator error $\norm{N(q) - \hat N(q)} \le \e_N D(q)$. Then
\begin{equation}\label{eq:coresetbridge}
\begin{aligned}
  \dQ(S, \hat S) &\le \e_F + 2M\tanh\!\left(\frac{-\log(1-\gamma)}{4}\right),\\
  \tau(S, \hat S) &\le -\log(1-\gamma) + \frac{\e_F}{M},
\end{aligned}
\end{equation}
where in case (b) one may set $\e_F = (\e_N + M\gamma)/(1-\gamma)$.
\end{proposition}
\begin{proof}
From $(1-\gamma)D \le \hat D \le (1+\gamma)D$ we have $|\log(\hat D/D)| \le -\log(1-\gamma)$. In case (b),
\begin{align*}
  \norm{F_S - F_{\hat S}}
  &= \norm{\frac{N - \hat N}{\hat D} + \frac{\hat D - D}{\hat D} F_S}\\
  &\le \frac{\e_N D + \gamma D M}{(1-\gamma)D}
  = \frac{\e_N + M\gamma}{1-\gamma}.
\end{align*}
Applying Proposition~\ref{prop:obs} and \eqref{eq:reverse} of Theorem~\ref{thm:m4} gives \eqref{eq:coresetbridge}.
\end{proof}

\begin{proposition}[Quantitative converse to coreset certification]\label{prop:converse}
Suppose $S = (N, D)$ and $\hat S = (\hat N, \hat D)$ in $\K$ satisfy $\dQ(S, \hat S) \le \e$ with $0 \le \e < M$, and define $b_\e = 4\operatorname{artanh}(\e/M)$. Then
\begin{equation}\label{eq:converse}
\begin{aligned}
  \sup_{q \in \Q} \norm{F_S(q) - F_{\hat S}(q)} &\le \e,\\
  \sup_{q \in \Q} \left|\log\frac{\hat D(q)}{D(q)}\right| &\le b_\e,\\
  \sup_{q \in \Q} \frac{\norm{\hat N(q) - N(q)}}{D(q)} &\le \e + M\bigl(e^{b_\e} - 1\bigr).
\end{aligned}
\end{equation}
\end{proposition}
\begin{proof}
For every $q \in \Q$, let $\delta(q) = |\log(\hat D(q)/D(q))|$. Proposition~\ref{prop:obs} gives $\norm{F_S(q)-F_{\hat S}(q)} \le \e$ and $M\tanh(\delta(q)/4) \le \e$. Taking the supremum over $q \in \Q$ gives the first inequality, and since $\e/M < 1$ and $\tanh$ is strictly increasing on $[0,\infty)$, $\delta(q) \le 4\operatorname{artanh}(\e/M) = b_\e$, proving the second. Hence $e^{-b_\e} \le \hat D(q)/D(q) \le e^{b_\e}$; because $1 - e^{-b} \le e^b - 1$ for all $b \ge 0$, we have $|\hat D(q) - D(q)|/D(q) \le e^{b_\e} - 1$. Writing
\begin{align*}
  \hat N(q) - N(q)
  &= D(q)\bigl(F_{\hat S}(q) - F_S(q)\bigr)\\
  &\quad + \bigl(\hat D(q) - D(q)\bigr) F_{\hat S}(q)
\end{align*}
and applying the triangle inequality with $\norm{F_{\hat S}(q)} \le M$ gives
\begin{align*}
  \frac{\norm{\hat N(q) - N(q)}}{D(q)}
  &\le \norm{F_{\hat S}(q) - F_S(q)}\\
  &\quad + \frac{|\hat D(q) - D(q)|}{D(q)}\norm{F_{\hat S}(q)}\\
  &\le \e + M(e^{b_\e} - 1),
\end{align*}
yielding the third inequality in \eqref{eq:converse}.
\end{proof}

As a direct consequence of Proposition~\ref{prop:obs}, Proposition~\ref{prop:converse} pairs with Proposition~\ref{prop:coreset} to show that for $\e < M$, append-observational fidelity is quantitatively equivalent to uniform normalized-output and relative-mass (hence relative-numerator) preservation up to explicit bounds, without requiring separate physical storage of numerator and denominator.

\parhead{Instantiating existing coreset theorems: mass preservation and key-gauge semantics.}
We now instantiate Proposition~\ref{prop:coreset} on the primary constructions of Cohen~\cite{cohen} (\texttt{arXiv:2609.06327v3}) and Liberty, Andoni, and Kleiner~\cite{liberty} (\texttt{arXiv:2605.05602v1}), using the Euclidean norm $\norm{\cdot}_2$ on $\R^{d_k}$ and $\R^{d_v}$ and writing $[n] = \{1,\ldots,n\}$ ($n \ge 1$).

Let an original uncentered history $\{(k_i^{\mathrm{orig}}, v_i)\}_{i=1}^n$ satisfy $\norm{v_i}_2 \le M$ on $\Q = \{q \in \R^{d_k} : \norm{q}_2 \le \varrho\}$ ($\varrho \ge 0$). Set $\bar k = \frac{1}{n}\sum_{i=1}^n k_i^{\mathrm{orig}}$ and $r_K = \max_{i\in[n]} \norm{k_i^{\mathrm{orig}} - \bar k}_2$. Define the key rescaling parameter $\kappa > 0$ explicitly by taking $\kappa = r_K$ when $r_K > 0$ and $\kappa = 1$ (or any positive scalar, giving $\tilde k_i = 0$) when all centered keys vanish ($r_K = 0$). Set $\rho = \varrho\kappa$ (enlarging $\rho$ if a cited discrepancy bound requires $\rho > 0$ or $\rho \ge 1$, which only relaxes the query ball) and normalize via
\[
  \tilde k_i = \frac{k_i^{\mathrm{orig}} - \bar k}{\kappa},\qquad
  \tilde q = \kappa q\ (\norm{\tilde q}_2 \le \rho),\qquad
  \tilde v_i = \frac{v_i}{M},
\]
so $\sum_{i=1}^n \tilde k_i = 0$, $\norm{\tilde k_i}_2 \le 1$, and $\norm{\tilde v_i}_2 \le 1$. In centered coordinates, Jensen's inequality yields the denominator floor
\begin{equation}\label{eq:jensen}
  \tilde D(\tilde q) = \sum_{i=1}^n e^{\tilde q^\top \tilde k_i} \ge n \exp\!\left(\tilde q^\top \frac{1}{n}\sum_{i=1}^n \tilde k_i\right) = n.
\end{equation}
By Proposition~\ref{prop:gauge}, translating both states' keys by $-\bar k$ multiplies both $(N, D)$ and $(\hat N, \hat D)$ by $c(q) = e^{-q^\top\bar k} > 0$ and leaves $\dQ$ and $\tau$ unchanged; the metric bounds in Corollary~\ref{cor:coreset_inst} therefore hold identically in centered or uncentered coordinates, whereas restoring uncentered keys $k_i^{\mathrm{orig}}$ (or shifting future tokens by $-\bar k$) is required specifically before \emph{composing} the summary with future uncentered physical KV tokens.

\begin{corollary}[Append robustness of reweighted query-oblivious coresets]\label{cor:coreset_inst}
Let $S = (N, D) \in \K$ be the uncentered state of $\{(k_i^{\mathrm{orig}}, v_i)\}_{i=1}^n$ on $\Q = \{\norm{q}_2 \le \varrho\}$ with Euclidean value bound $\norm{v_i}_2 \le M$ and normalized parameters $(\kappa, \rho)$ above.
\begin{enumerate}
\item \emph{Uniformly reweighted sampling --- exact total atomic mass restoration (instantiating Cohen~\cite[Theorem~3.7]{cohen}):} Suppose an unweighted subset $C \subseteq [n]$ of size $m \in \{1,\ldots,n\}$ satisfies
\begin{equation}\label{eq:avgbounds}
\begin{aligned}
  \norm{\frac{\tilde N(\tilde q)}{n} - \frac{\tilde N_C(\tilde q)}{m}}_2 &\le \delta_N,\\
  \left|\frac{\tilde D(\tilde q)}{n} - \frac{\tilde D_C(\tilde q)}{m}\right| &\le \delta_D
\end{aligned}
\end{equation}
for all $\norm{\tilde q}_2 \le \rho$, with $0 \le \delta_D < 1$. Assigning each retained atom weight $n/m$ restores the total original atomic mass \emph{exactly} ($\sum_{j \in C} \frac{n}{m} = n$, hence $\hat D(0) = D(0) = n$ at $q = 0 \in \Q$), and $\hat S = (\frac{n}{m}N_C, \frac{n}{m}D_C) \in \K$ satisfies \eqref{eq:coresetbridge} with $\gamma = \delta_D$ and $\e_N = M\delta_N$. For any $\e \in (0, 1]$, Cohen~\cite[Theorem~3.7]{cohen} establishes: (a)~\emph{deterministic existence} (by averaging, not a constructive deterministic algorithm) of $C$ with $m = \min\{n, \lceil 32 e^{2\rho}/\e^2 \rceil\}$; and (b)~a \emph{randomized sampling guarantee} that a uniformly random subset without replacement of size $m = \min\{n, \lceil 128 e^{2\rho}/\e^2 \rceil\}$ succeeds with probability at least $3/4$, both achieving $\delta_N = \delta_D = \delta \le \e/4 \le 1/4$ (and trivially $\delta = 0$, $\hat S = S$ when $m = n$), whence
\begin{equation}\label{eq:cohen37}
  \dQ(S, \hat S) \le \frac{5}{6} M\e,\qquad \tau(S, \hat S) \le \e.
\end{equation}
\item \emph{Dyadically reweighted recursive halving --- dyadic rescaling with certified approximate mass preservation (instantiating Cohen~\cite[Lemmas~3.5--3.6, App.~B]{cohen} and Liberty--Andoni--Kleiner~\cite[Thms.~3--5, Cor.~1, Fact~2]{liberty}):} Suppose $\ell \ge 0$ halving steps $P_0 = [n] \supset P_1 \supset \cdots \supset P_\ell$ satisfy $\tilde N_{P_{j+1}} = \tfrac{1}{2}(\tilde N_{P_j} + \delta_{A,j})$ and $\tilde D_{P_{j+1}} = \tfrac{1}{2}(\tilde D_{P_j} + \delta_{B,j})$ with $\norm{\delta_{A,j}}_2 \le D_A$ and $|\delta_{B,j}| \le D_B$ uniformly for $\norm{\tilde q}_2 \le \rho$, and $(2^\ell - 1) D_B < n$ ($\ell = 0$ handles small histories requiring no halving). Assigning every atom in $P_\ell$ dyadic weight $2^\ell$ yields $\hat S_{\mathrm{halv}} = (2^\ell N_{P_\ell}, 2^\ell D_{P_\ell}) \in \K$ with $\hat D_{\mathrm{halv}}(0) = 2^\ell |P_\ell|$, which need \emph{not} equal $n$ when halving steps are not exactly balanced; instead, telescoping yields certified approximate mass preservation
\begin{equation}\label{eq:halv_mass}
  \bigl|2^\ell \tilde D_{P_\ell}(\tilde q) - \tilde D(\tilde q)\bigr| \le (2^\ell - 1) D_B
\end{equation}
and \eqref{eq:coresetbridge} with $\gamma = (2^\ell - 1)D_B / n$ and $\e_N = M(2^\ell - 1)D_A / n$. Under the stopping rule $2^{\ell+1}(D_A + D_B) \le \e n$ ($\e \in (0,1]$, setting $\ell = 0$ if $2(D_A + D_B) > \e n$),
\begin{equation}\label{eq:cohenhalv}
\begin{aligned}
  \dQ(S, \hat S_{\mathrm{halv}}) &\le \left(1 + \frac{\log 2}{2}\right) M\e < 1.35\,M\e,\\
  \tau(S, \hat S_{\mathrm{halv}}) &\le (1 + \log 2)\e < 1.70\,\e.
\end{aligned}
\end{equation}
Two distinct primary constructions instantiate \eqref{eq:cohenhalv} under their own assumptions:
\begin{itemize}
\item \emph{Cohen~\cite[Lemmas~3.5--3.6, App.~B]{cohen}:} For arbitrary $d_k, d_v \ge 1$, Banaszczyk's theorem gives \emph{deterministic existence} of $(D_A, D_B, D_N)$-balanced signings with $D_A = O(e^\rho(\sqrt{d_v} + \sqrt{d_k\log(1+\rho)}))$, $D_B = O(e^\rho(1 + \sqrt{d_k\log(1+\rho)}))$, and $D_N = 10\sqrt{5}$ ($n/2^\ell \le |P_\ell| \le n/2^\ell + D_N$), yielding $|P_\ell| \le \min\{n,\, 4(D_A+D_B)/\e + D_N\} = O(\min\{n,\, e^\rho(\sqrt{d_v} + \sqrt{d_k\log(1+\rho)})/\e\})$. Separately, Cohen's Gram--Schmidt walk~\cite[Theorem~1.2(b), App.~C]{cohen} gives a \emph{randomized polynomial-time} construction achieving a uniform per-level bound $D^*$ across all levels with probability $\ge 1-\delta$ (at an extra $O(\sqrt{\log(\log n/\delta)})$ factor), on which event \eqref{eq:cohenhalv} holds constructively.
\item \emph{Liberty--Andoni--Kleiner~\cite[Thms.~3--5, Cor.~1, Fact~2]{liberty}:} Formulated for \emph{equal dimensions} $d_k = d_v = d$ (general $d_k \ne d_v$ reduce to this setting by zero-padding the smaller vectors to $d = \max\{d_k, d_v\}$, preserving inner products, Euclidean norms, and centering). Their simultaneous Banaszczyk balancing result (Theorem~5 and Fact~2 combining Theorem~3, Corollary~1, and Fact~1) establishes \emph{deterministic existence} at each step of a subset with $|P_{j+1}| \le |P_j|/2$ and their own discrepancy bound $D_A = D_B = D_{\mathrm{LAK}} = c_{\mathrm{LAK}}\sqrt{d}\, e^\zeta$ ($\zeta = \rho + \tfrac{1}{2}\log\rho + o(\log\log\rho)$); stopping at maximal $\ell \ge 0$ with $2^{\ell+2}D_{\mathrm{LAK}} \le \e n$ ($\ell = 0$ if $4D_{\mathrm{LAK}} > \e n$) yields $|P_\ell| \le n/2^\ell \le \min\{n,\, 8D_{\mathrm{LAK}}/\e\} = O(\min\{n,\, \sqrt{d}\,e^\zeta/\e\})$ with \eqref{eq:cohenhalv}.
\end{itemize}
\end{enumerate}
\end{corollary}
\begin{proof}[Proof outline]
Canceling the common gauge factor $e^{q^\top\bar k}$ (Proposition~\ref{prop:gauge}), dividing \eqref{eq:avgbounds} or the telescoped halving discrepancies by the Jensen floor $\tilde D(\tilde q) \ge n$ in \eqref{eq:jensen}, and substituting into Proposition~\ref{prop:coreset}(b) yield \eqref{eq:cohen37} and \eqref{eq:cohenhalv}; complete constant derivations and source-specific size bounds appear in Appendix~\ref{app:coreset}.
\end{proof}

Two structural observations clarify Corollary~\ref{cor:coreset_inst}. First, while the headline statements of Liberty--Andoni--Kleiner~\cite[Theorem~4]{liberty} and Cohen~\cite[Theorem~1.1]{cohen} guarantee only unweighted normalized-output error (which does not rule out \eqref{eq:tenx}), their internal simultaneous $(N,D)$ discrepancy bounds transfer directly to $(\dQ,\tau)$ once the appropriate weights are restored. Second, the three mass-preservation mechanisms across our paper differ fundamentally: uniformly reweighted sampling ($n/m$) and positive moment-preserving recompression with $1 \in \operatorname{span}\{g_1,\ldots,g_m\}$ (Theorem~\ref{thm:p1b}) preserve total atomic mass \emph{exactly}, whereas dyadically reweighted halving ($2^\ell$) achieves \emph{certified approximate mass preservation} via \eqref{eq:halv_mass}.

\subsection{Comparison with Competing Theory and Systems Literature}
\label{sec:related}

Table~\ref{tab:related} positions our results against adjacent mathematical and systems work.

\parhead{Comparison with Yang and Li~\cite{yang2026}.}
Yang and Li (\texttt{arXiv:2608.21541v1}) analyze when softmax attention is compressible by separating threshold-count sparsity, mass coverage, fixed-row output fidelity, and downstream task utility. Three of their theorems merit direct comparison:
\begin{itemize}
\item \emph{Overlap on the $\tanh$ log-mass sensitivity envelope (Theorem~C.1):} When an evicted tail $T$ is replaced by a single synthetic slot $\star$ alongside an unchanged retained set $S$ at a fixed future query, Yang and Li~\cite[Theorem~C.1]{yang2026} prove the a posteriori bound $\norm{O - \widehat O}_2 \le V_{\max}[2|p - q| + \min\{p,q\}\norm{a - r}_1]$ with $|p - q| \le \tanh(|\Lambda_T - z_\star|/4)$, where $\Lambda_T = \log\sum_{j\in T} e^{z_j}$ is the tail log-partition and the unchanged retained mass $Z_S = \sum_{i\in S} e^{z_i}$ plays the role of our appended suffix mass $t$. This relies on the same sigmoid log-mass sensitivity mechanism captured in Lemma~\ref{lem:mass} and Proposition~\ref{prop:obs}, establishing the $\tanh(\delta/4)$ future-query sensitivity envelope in the cache-eviction setting.
\item \emph{Fixed-row factorization and linear-sketch minimality (Theorems~3.7 and~C.2):} Yang and Li's Theorem~3.7 establishes the exact additive completion $u^\star = (1-p_S)\bar V_T$ for one realized attention row, and their Theorem~C.2 proves a $(k+1)d$ rank floor for any \emph{linear} sketch recovering one fixed weight row and all its retained sub-budgets---noting explicitly that Theorem~C.2 does not cover nonlinear summaries or future-query caches.
\item \emph{Where our results go beyond~\cite{yang2026}:} Our work addresses a different mathematical layer: (i)~operational dynamic state equivalence and the two-sided metric characterization of $\dQ$ across future common continuations (Theorems~\ref{thm:t1}, \ref{thm:m1}; Propositions~\ref{prop:metric}, \ref{prop:obs}, \ref{prop:gauge}, \ref{prop:coreset}, \ref{prop:converse}); (ii)~\emph{continuous nonlinear} representation-width lower bounds against arbitrary decoders, especially the constant-output variable-key hidden-mass obstructions in Theorems~\ref{thm:w4} and~\ref{thm:w4orth} and Corollary~\ref{cor:w4phys} and generic exact dimension in Theorem~\ref{thm:t2}; (iii)~multi-session collective continuous widths both for fixed-key circular families on $\Q_L$ (Theorem~\ref{thm:c2}) and for variable-key hidden-mass families on general $\Q$ (Theorem~\ref{thm:c3}); and (iv)~positive-cone $\max$-merge stability and indefinite-horizon positive maintenance with numerical-drift and weight-quantization guarantees (Theorems~\ref{thm:m3}, \ref{thm:m4}, \ref{thm:d1}, \ref{thm:p1b}; Corollaries~\ref{cor:fixed}, \ref{cor:numdrift}, \ref{cor:bitwidth}).
\end{itemize}

\parhead{Cache-replacement reliability versus representation widths.}
Whereas cache-replacement and state-substitution analyses~\cite{reskv,yang2026} ask when a compressed cache reliably preserves downstream attention outputs under normalization-mass shifts, Theorems~\ref{thm:w4}, \ref{thm:w4orth}, and~\ref{thm:c3} and Corollary~\ref{cor:w4phys} address the complementary \emph{continuous nonlinear representation-width} problem: quantifying the minimum continuous coordinate dimension required to preserve hidden historical mass against arbitrary decoders.

\parhead{Cache compression, streaming bounds, and identifiability.}
Recursive and cascade attention merge output and log-normalizer statistics~\cite{cascade,flashinfer}, and Hydragen shares prefix computations across batches~\cite{hydragen}. CommVQ applies commutative vector quantization compatible with rotary embeddings~\cite{commvq}, ResKV reconstructs omitted contributions under a common normalizer~\cite{reskv}, and an informal repository draft (\emph{FOCUS-Fabric}) likewise notes value and log-mass errors in composable memory~\cite{focusfabric}. Multi-agent cache sharing is studied empirically in TokenDance~\cite{tokendance} and PolyKV~\cite{polykv}, echoing Gray--Wyner common/private source coding~\cite{graywyner}; Theorems~\ref{thm:c2} and~\ref{thm:c3} provide stylized continuous-dimension baselines (both fixed-key spectral and variable-key hidden-mass) rather than an empirical comparison with serving systems.

On the theoretical complexity side, Haris and Onak~\cite{haris}, BalanceKV~\cite{balancekv}, and Chen et al.~\cite{tightstream} establish finite-bit streaming lower bounds and discrepancy summaries, complementary to our continuous real-coordinate widths (Table~\ref{tab:width_vs_bits}), while Corollary~\ref{cor:bitwidth} adds an explicit $O(\log\log T)$ weight bit-width bound for fixed-feature positive summaries. Haverbeck et al.~\cite{risk} study distributional minimax risk via response-operator spectra, whereas Theorem~\ref{thm:w3} is a worst-case finite-query reference. Kim and Son~\cite{tworead} prove lower bounds for two-read linear caches, whereas we place no read constraint on decoders. Konda~\cite{konda} studies deep parameter identifiability, whereas Theorems~\ref{thm:t1} and~\ref{thm:t2} address single-head historical state identifiability under external appends.

\subsection{Model Boundaries, Open Problems, and Conclusion}
\label{sec:conclusion}

\parhead{Model boundaries and open problems.}
Our analysis is confined to a single head, bounded values, a static compact query domain $\Q$, fixed visibility, continuous real-coordinate width accounting, and open-loop common continuations (with Corollaries~\ref{cor:numdrift}--\ref{cor:bitwidth} bridging exact-arithmetic maintenance to logarithmic numerical solver drift and $O(\log\log T)$ weight bit-width, excluding key--value coordinates). Sliding-window eviction, token-deleting masks, and learned sinks require modified state algebras; in closed-loop generation, small attention errors can alter future tokens and queries, so single-head $\dQ$ bounds do not imply token agreement or end-to-end task accuracy. Four problems remain open: (i)~quantitative continuous widths for high-dimensional variable-key families with \emph{unknown} (unaligned) shared subspaces and non-orthogonal key geometry; (ii)~finite-bit \emph{key--value coordinate} lower bounds under append observability (beyond the weight bit-width bound of Corollary~\ref{cor:bitwidth}); (iii)~feature counts and collective costs for multi-plane rotary attention; and (iv)~closed-loop trajectory stability under endogenous query feedback.

\parhead{Conclusion.}
Answering what representation complexity is required to preserve attention states that agree on all current normalized observations yet diverge under future continuations leads directly to the append-observational metric $\dQ$ on $\K$. Within this geometry, constant-output variable-key histories force full local continuous dimension $r_{\min} = L$ under both relaxed and repeated physical unit-weight continuations---even in 1D (Theorem~\ref{thm:w4} and Corollary~\ref{cor:w4phys}) and without conditioning decay at an $L$-independent macroscopic threshold within a linear suffix horizon $H_* = O(L)$ in $d_k \ge L$ (Theorem~\ref{thm:w4orth})---and directly drive exact collective dimension separation $R_{\joint}(\e) = s + Bp$ across concurrent sessions (Theorem~\ref{thm:c3}, complementing the fixed-key harmonic baseline of Theorem~\ref{thm:c2}). Positive-cone lifts, fixed-moment cubature with logarithmic numerical-drift and $O(\log\log T)$ weight-quantization bounds (Corollaries~\ref{cor:numdrift}--\ref{cor:bitwidth}), and appropriately reweighted query-oblivious coresets delineate when certified compositional maintenance is achievable.

\appendices

\section{Boundary-Horizon Verification for Theorem~\ref{thm:d1}}
\label{app:d1}

This appendix verifies the persistent active storage bound $R_T \le R_{\mathrm{buffer}} + C_0 + \sum_{\ell=1}^{K_T} C_\ell(\eta_\ell)$ in Theorem~\ref{thm:d1}(2) at boundary horizons after completed carries (excluding temporary merge workspace), separating $c = 1$ from $c \ge 2$:
\begin{itemize}
\item \emph{Case $c = 1$:} Because the incomplete insertion buffer has size $< 1$, it is always empty ($r = T \bmod 1 = 0$, $R_{\mathrm{buffer}} = 0$) and $q = T$. At $T = 1$, $q = 1$ holds one exact level-$0$ block (cost $\le C_0$). At $T = c + 1 = 2 = 2c$, $q = 2 = (10)_2$ triggers a carry producing a single level-$1$ summary (cost $\le C_1(\eta_1)$), rather than a level-$0$ block plus a partial buffer. At $T = 3$, $q = 3 = (11)_2$ holds one level-$0$ block and one level-$1$ summary (cost $\le C_0 + C_1(\eta_1)$).
\item \emph{Case $c \ge 2$:} For $T < c$ ($q = 0, r = T$), only the incomplete buffer is active (cost $\le R_{\mathrm{buffer}}$). At $T = c$ ($q = 1, r = 0$), only the full level-$0$ block is active (cost $\le C_0$). At $T = c + 1 < 2c$ ($q = 1, r = 1$), the full level-$0$ block and a $1$-token insertion buffer coexist (cost $\le R_{\mathrm{buffer}} + C_0$). At $T = 2c$ ($q = 2 = (10)_2, r = 0$), carry produces a single level-$1$ summary (cost $\le C_1(\eta_1)$). At $T = 3c - 1$ ($q = 2, r = c - 1$), the level-$1$ summary coexists with the $(c-1)$-token buffer. At $T = 4c - 1$ ($q = 3 = (11)_2, r = c - 1$), the buffer, level-$0$ block, and level-$1$ summary are simultaneously active (cost $\le R_{\mathrm{buffer}} + C_0 + C_1(\eta_1)$).
\end{itemize}

\section{Proof Details for Corollary~\ref{cor:coreset_inst}}
\label{app:coreset}

Because $N(q) = M e^{q^\top \bar k} \tilde N(\tilde q)$, $D(q) = e^{q^\top \bar k} \tilde D(\tilde q)$, and the same positive factor $e^{q^\top \bar k}$ multiplies the uncentered reweighted subset, $e^{q^\top \bar k}$ cancels identically in all relative ratios (consistent with Proposition~\ref{prop:gauge}).

\parhead{Part~(1) (Uniformly reweighted sampling).}
Total atomic mass is $\sum_{j \in C} \frac{n}{m} = n$, and dividing \eqref{eq:avgbounds} by $\tilde D(\tilde q)/n \ge 1$ (from \eqref{eq:jensen}) gives
\begin{align*}
  \frac{|\hat D(q) - D(q)|}{D(q)} &= \frac{|\tilde D_C(\tilde q)/m - \tilde D(\tilde q)/n|}{\tilde D(\tilde q)/n} \le \delta_D,\\
  \frac{\norm{N(q) - \hat N(q)}_2}{D(q)} &= \frac{M\norm{\tilde N(\tilde q)/n - \tilde N_C(\tilde q)/m}_2}{\tilde D(\tilde q)/n} \le M\delta_N.
\end{align*}
Proposition~\ref{prop:coreset}(b) yields \eqref{eq:coresetbridge} with $\gamma = \delta_D$ and $\e_N = M\delta_N$. In Cohen~\cite[Theorem~3.7]{cohen}, if $m = n$ then $C = [n]$ and $\delta = 0$; otherwise the two-block RKHS argument yields $\delta_N = \delta_D = \delta \le \e/4 \le 1/4$ (by deterministic existence via expectation at $m = \lceil 32e^{2\rho}/\e^2\rceil$, or with probability $\ge 3/4$ by Markov's inequality at $m = \lceil 128e^{2\rho}/\e^2\rceil$). Then $\e_F = \frac{2M\delta}{1-\delta} \le \frac{8}{3}M\delta$, $-\log(1-\delta) \le \frac{4}{3}\delta$ on $[0,1/4]$, and $2\tanh(-\tfrac{1}{4}\log(1-\delta)) \le \frac{2}{3}\delta$, yielding $\dQ \le \frac{10}{3}M\delta \le \frac{5}{6}M\e$ and $\tau \le 4\delta \le \e$, proving \eqref{eq:cohen37}.

\parhead{Part~(2) (Dyadically reweighted recursive halving).}
If $\ell = 0$ then $P_0 = [n]$, $\hat S_{\mathrm{halv}} = S$, and all bounds hold with zero error. For $\ell \ge 1$, multiplying $\tilde N_{P_{j+1}} = \tfrac{1}{2}(\tilde N_{P_j} + \delta_{A,j})$ and $\tilde D_{P_{j+1}} = \tfrac{1}{2}(\tilde D_{P_j} + \delta_{B,j})$ by $2^{j+1}$ and telescoping over $j = 0,\ldots,\ell-1$ gives
\begin{align*}
  \norm{2^\ell \tilde N_{P_\ell}(\tilde q) - \tilde N(\tilde q)}_2 &\le (2^\ell - 1) D_A,\\
  \left|2^\ell \tilde D_{P_\ell}(\tilde q) - \tilde D(\tilde q)\right| &\le (2^\ell - 1) D_B,
\end{align*}
establishing \eqref{eq:halv_mass}. Dividing by $\tilde D(\tilde q) \ge n$ and canceling $e^{q^\top \bar k}$ gives $|\hat D_{\mathrm{halv}} - D| \le \gamma D$ and $\norm{N - \hat N_{\mathrm{halv}}}_2 \le \e_N D$ with $\gamma = (2^\ell - 1)D_B / n < 1$ and $\e_N = M(2^\ell - 1)D_A / n$. Under $2^{\ell+1}(D_A + D_B) \le \e n$, $\gamma \le \e/2 \le 1/2$ and $\e_N + M\gamma \le M \cdot 2^\ell(D_A + D_B)/n \le M\e/2$, so $\e_F = (\e_N + M\gamma)/(1-\gamma) \le M\e$. Using $-\log(1-\gamma) \le (2\log 2)\gamma \le (\log 2)\e$ on $[0, 1/2]$ and $\tanh(u) \le u$ in \eqref{eq:coresetbridge} yields $\dQ \le M\e + \tfrac{M}{2}(\log 2)\e$ and $\tau \le (\log 2)\e + \e$, proving \eqref{eq:cohenhalv}. Finally, under Cohen's Lemmas~3.5--3.6, maximality of $\ell$ gives $n/2^\ell < 4(D_A+D_B)/\e$ and $|P_\ell| \le n/2^\ell + D_N$; under Liberty--Andoni--Kleiner's simultaneous balancing (Theorem~5 / Fact~2 with $D_A = D_B = D_{\mathrm{LAK}}$ and $|P_{j+1}| \le |P_j|/2$), maximality of $\ell$ under $2^{\ell+2}D_{\mathrm{LAK}} \le \e n$ gives $|P_\ell| \le n/2^\ell < 8 D_{\mathrm{LAK}}/\e$ (and $|P_0| = n < 4D_{\mathrm{LAK}}/\e$ when $\ell = 0$).

\section{Quantitative Conditioning Examples}
\label{app:conditioning}

To quantify the conditioning decay of $\mu = \norm{J^{-1}}_{\infty\to\infty}^{-1}$ in Theorem~\ref{thm:w4} and the uniform ceiling-suffix horizon $H_* = \lceil L e^{Q_{\max} K_{\max}} \rceil$ in Corollary~\ref{cor:w4phys} (Remark~\ref{rem:w4scope}), Table~\ref{tab:conditioning} evaluates canonical uniformly spaced positive reference keys $k_j^0 = \frac{j}{L+1} K_{\max}$ ($j = 1,\ldots,L$) for $L \in \{2, 3, 4, 6\}$ under two parameterizations: (1)~a fixed score budget ($K_{\max} = 1$, $Q_{\max} = 1$, probe step $h = 1/L$), and (2)~a fixed probe step ($K_{\max} = 1$, $h = 1$, $Q_{\max} = L$).

In every 1D configuration listed, the quadratic-remainder limit $a_{\mathrm{quad}} = 2\mu / Q_{\max}^2$ is strictly smaller than the geometric key-separation limit $a_{\mathrm{geom}} = \frac{K_{\max}}{2(L+1)}$, so the maximal admissible cube radius $a_{\max} = 2\mu / Q_{\max}^2$ is attained and maximizes the relaxed hidden-mass threshold $\e_{\mathrm{hm}} / M = \tanh\bigl(\mu^2 / (2 Q_{\max}^2)\bigr)$; Corollary~\ref{cor:w4phys}'s physical-continuation threshold is $\frac{8}{9}\e_{\mathrm{hm}}/M$. Under reciprocal rescaling $k \mapsto \alpha k$, $q \mapsto \alpha^{-1} q$ ($\alpha > 0$), the score budget $\rho = Q_{\max} K_{\max}$, $\e_{\mathrm{hm}}/M$, and $H_*$ remain invariant while $\mu \mapsto \alpha^{-1}\mu$ and $a_{\max} \mapsto \alpha a_{\max}$.

Table~\ref{tab:conditioning} exhibits the worst-case 1D Vandermonde conditioning trade-off explicitly: holding $\rho = Q_{\max} K_{\max} = 1$ fixed in 1D keeps the uniform ceiling-suffix horizon small ($H_* \in [6, 17]$), but packs the Vandermonde nodes $z_j = e^{h k_j^0}$ near $1$ so $\e_{\mathrm{hm}}/M$ decays to $2.66\times 10^{-20}$ at $L = 6$. Allowing $Q_{\max} = L$ (fixed step $h = 1$) separates the nodes across $(1, e)$ and improves $\e_{\mathrm{hm}}/M$ by over seven orders of magnitude at $L = 6$ ($5.49\times 10^{-13}$), at the expense of raising $H_*$ to $2421$.

As the bottom block of Table~\ref{tab:conditioning} demonstrates, this super-exponential decay is specifically a worst-case $d_k = 1$ conditioning trade-off rather than a general limitation of hidden-mass widths. In the high-dimensional orthogonal-axis construction of Theorem~\ref{thm:w4orth} ($d_k \ge L$, $Q_{\max} = 1$, $a = 2$), diagonal conditioning stays constant ($\mu = 0.500$), the relaxed and physical thresholds remain macroscopically large ($\e_{\mathrm{orth}}/M = \tanh(1/2) \approx 4.62\times 10^{-1}$ and $\frac{8}{9}\e_{\mathrm{orth}}/M \approx 4.11\times 10^{-1}$) independently of $L$, and the uniform ceiling-suffix horizon grows only linearly ($H_* = \lceil (L-1)(e^2+1)\rceil \in \{9, 17, 26, 42\}$). All entries are computed at 50-digit precision and cross-checked to $\norm{J J^{-1} - I}_\infty < 10^{-40}$ by \href{https://github.com/openkedge/pdds/blob/main/attention-nonlinear-widths/scripts/generate_validation_figures.py}{\texttt{scripts/generate\_validation\_figures.py}}, publicly available at \url{https://github.com/openkedge/pdds/tree/main/attention-nonlinear-widths}.

\begin{table}[!t]
\centering
\caption{Quantitative Conditioning and Open-Loop Horizon Examples for 1D Vandermonde Keys and High-Dimensional Orthogonal Axes}
\label{tab:conditioning}
\begin{minipage}{\ifCLASSOPTIONtwocolumn\columnwidth\else 90mm\fi}
\centering
\setlength{\tabcolsep}{3.0pt}
\renewcommand{\arraystretch}{1.12}
\begin{tabular*}{\linewidth}{@{\extracolsep{\fill}}c c c c r@{}}
\toprule
$L$ & $\mu$ & $a_{\max}$ or $a$ & $\e_{\mathrm{hm}}/M$ or $\e_{\mathrm{orth}}/M$ & $H_*$ \\
\midrule
\multicolumn{5}{@{}l}{\emph{1D fixed score budget ($d_k=1$, $K_{\max}=1$, $Q_{\max}=1$, $h = 1/L$):}} \\
2 & $2.40\times 10^{-2}$ & $4.80\times 10^{-2}$ & $2.89\times 10^{-4}$ & $6$ \\
3 & $3.28\times 10^{-4}$ & $6.56\times 10^{-4}$ & $5.37\times 10^{-8}$ & $9$ \\
4 & $3.87\times 10^{-6}$ & $7.73\times 10^{-6}$ & $7.48\times 10^{-12}$ & $11$ \\
6 & $2.31\times 10^{-10}$ & $4.61\times 10^{-10}$ & $2.66\times 10^{-20}$ & $17$ \\
\midrule
\multicolumn{5}{@{}l}{\emph{1D fixed probe step ($d_k=1$, $K_{\max}=1$, $h = 1$, $Q_{\max} = L$):}} \\
2 & $8.21\times 10^{-2}$ & $4.11\times 10^{-2}$ & $8.43\times 10^{-4}$ & $15$ \\
3 & $8.25\times 10^{-3}$ & $1.83\times 10^{-3}$ & $3.78\times 10^{-6}$ & $61$ \\
4 & $7.13\times 10^{-4}$ & $8.91\times 10^{-5}$ & $1.59\times 10^{-8}$ & $219$ \\
6 & $6.29\times 10^{-6}$ & $3.49\times 10^{-7}$ & $5.49\times 10^{-13}$ & $2421$ \\
\midrule
\multicolumn{5}{@{}l}{\emph{Orthogonal axes (Thm.~\ref{thm:w4orth}, $d_k \ge L$, $Q_{\max} = 1$, $a = 2$):}} \\
2 & $5.00\times 10^{-1}$ & $2.00\times 10^{0}$ & $4.62\times 10^{-1}$ & $9$ \\
3 & $5.00\times 10^{-1}$ & $2.00\times 10^{0}$ & $4.62\times 10^{-1}$ & $17$ \\
4 & $5.00\times 10^{-1}$ & $2.00\times 10^{0}$ & $4.62\times 10^{-1}$ & $26$ \\
6 & $5.00\times 10^{-1}$ & $2.00\times 10^{0}$ & $4.62\times 10^{-1}$ & $42$ \\
\bottomrule
\end{tabular*}
\par\vspace{1.2mm}
\raggedright\scriptsize
\emph{Notes:} 1D regimes (Theorem~\ref{thm:w4}, Corollary~\ref{cor:w4phys}) use $k_j^0 = \tfrac{j}{L+1}K_{\max}$ ($K_{\max}=1$), $a_{\max}=2\mu/Q_{\max}^2$, $\e_{\mathrm{hm}}/M = \tanh(\mu^2/(2Q_{\max}^2))$, and $H_* = \lceil L e^{Q_{\max}K_{\max}}\rceil$. Orthogonal axes (Theorem~\ref{thm:w4orth}) use $d_k \ge L$, $Q_{\max}=1$, $a=2$, $\mu=\tfrac{1}{2}Q_{\max}$, $\e_{\mathrm{orth}}/M=\tanh(Q_{\max}a/4)$, and $H_* = \lceil (L-1)(e^{Q_{\max}a}+1)\rceil$. In all regimes, the physical-continuation threshold is $\tfrac{8}{9}$ of the relaxed threshold.
\end{minipage}
\end{table}

\end{document}

%% file: figures/validation_panels.tex
\begin{tikzpicture}[x=1cm,y=1cm,font=\scriptsize\sffamily]
  \begin{scope}
    \foreach \yt/\yl in {0.0/0, 0.3/0.3, 0.6/0.6, 0.9/0.9, 1.2/1.2} {
      \draw[black!10,very thin] (0,{(\yt/1.20)*3.250}) -- (4.350,{(\yt/1.20)*3.250});
      \draw[primary!70,thin] (-0.07,{(\yt/1.20)*3.250}) -- (0,{(\yt/1.20)*3.250});
      \node[left,font=\tiny\sffamily] at (-0.08,{(\yt/1.20)*3.250}) {\yl};
    }
    \foreach \xt in {0, 5, 10, 15, 20} {
      \draw[black!10,very thin] ({(\xt/20.0)*4.350},0) -- ({(\xt/20.0)*4.350},3.250);
      \draw[primary!70,thin] ({(\xt/20.0)*4.350},-0.07) -- ({(\xt/20.0)*4.350},0);
      \node[below,font=\tiny\sffamily] at ({(\xt/20.0)*4.350},-0.08) {\xt};
    }
    \draw[primary,semithick] (0,0) rectangle (4.350,3.250);

    \draw[alert!80,densely dashed,semithick] (0,2.501) -- (4.350,2.501);
    \node[anchor=north east,font=\tiny\sffamily,color=alert,fill=white,inner sep=1pt] at (4.290,2.461) {$\tfrac{8}{9} d_{\mathrm{app}} = 0.924M$};

    \draw[accent,very thick] (0.000,0.000) -- (0.022,0.439) -- (0.043,0.797) -- (0.076,1.221) -- (0.109,1.548) -- (0.163,1.944) -- (0.217,2.216) -- (0.272,2.407) -- (0.326,2.543) -- (0.381,2.640) -- (0.435,2.708) -- (0.489,2.755) -- (0.544,2.786) -- (0.598,2.804) -- (0.652,2.813) -- (0.688,2.814) -- (0.734,2.812) -- (0.788,2.804) -- (0.843,2.793) -- (0.897,2.778) -- (0.952,2.760) -- (1.006,2.741) -- (1.060,2.719) -- (1.115,2.697) -- (1.169,2.673) -- (1.223,2.649) -- (1.278,2.624) -- (1.332,2.599) -- (1.387,2.573) -- (1.441,2.548) -- (1.495,2.522) -- (1.550,2.496) -- (1.604,2.471) -- (1.658,2.445) -- (1.713,2.420) -- (1.767,2.395) -- (1.822,2.370) -- (1.876,2.346) -- (1.930,2.321) -- (1.985,2.297) -- (2.039,2.274) -- (2.093,2.250) -- (2.148,2.227) -- (2.202,2.205) -- (2.257,2.182) -- (2.311,2.160) -- (2.365,2.139) -- (2.420,2.117) -- (2.474,2.096) -- (2.528,2.076) -- (2.583,2.055) -- (2.637,2.036) -- (2.692,2.016) -- (2.746,1.997) -- (2.800,1.978) -- (2.855,1.959) -- (2.909,1.940) -- (2.963,1.922) -- (3.018,1.905) -- (3.072,1.887) -- (3.127,1.870) -- (3.181,1.853) -- (3.235,1.836) -- (3.290,1.820) -- (3.344,1.804) -- (3.398,1.788) -- (3.453,1.772) -- (3.507,1.757) -- (3.562,1.742) -- (3.616,1.727) -- (3.670,1.712) -- (3.725,1.698) -- (3.779,1.684) -- (3.833,1.670) -- (3.888,1.656) -- (3.942,1.643) -- (3.997,1.629) -- (4.051,1.616) -- (4.105,1.603) -- (4.160,1.591) -- (4.214,1.578) -- (4.268,1.566) -- (4.323,1.554) -- (4.350,1.548);

    \draw[primary!60,densely dotted] (0.688,0) -- (0.688,2.814);
    \fill[primary] (0,0) circle (1.5pt);
    \fill[accent] (0.688,2.814) circle (1.9pt);
    \fill[alert] (0.870,2.786) circle (1.8pt);

    \node[anchor=south west,font=\tiny\sffamily,color=accent,inner sep=1pt] at (0.748,2.894) {$t^*\!=\!\sqrt{10}$ ($1.039M$)};
    \node[anchor=west,font=\tiny\sffamily,color=alert,fill=white,inner sep=1pt] at (1.030,1.536) {$\lceil t^*\rceil\!=\!4$ ($1.029M$)};
    \draw[alert,thin,-{Stealth[length=1.2mm]}] (1.290,1.666) -- (0.900,2.726);
    \node[anchor=south west,font=\tiny\sffamily,color=primary] at (0.14,0.06) {$t\!=\!0$: $0$ current err.};

    \node[above,font=\scriptsize\bfseries\sffamily,color=primary] at (2.175,3.330) {(a) Hidden Historical Mass ($D\!=\!1,\,\hat D\!=\!10$)};
    \node[below,font=\tiny\sffamily] at (2.175,-0.34) {Appended suffix mass $t$ ($g=-M$)};
    \node[rotate=90,above,font=\tiny\sffamily] at (-0.48,1.625) {Output gap / $M$};
  \end{scope}

  \begin{scope}
    \foreach \lv/\yl in {-6/10^{-6}, -4/10^{-4}, -2/10^{-2}, 0/10^{0}} {
      \draw[black!10,very thin] (6.000,{((\lv+6.0)/6.0)*3.250}) -- (10.350,{((\lv+6.0)/6.0)*3.250});
      \draw[primary!70,thin] (5.930,{((\lv+6.0)/6.0)*3.250}) -- (6.000,{((\lv+6.0)/6.0)*3.250});
      \node[left,font=\tiny\sffamily] at (5.930,{((\lv+6.0)/6.0)*3.250}) {$\yl$};
    }
    \foreach \xk in {2, 4, 6, 8, 10, 12} {
      \draw[black!10,very thin] ({6.000+((\xk-0.2)/11.8)*4.350},0) -- ({6.000+((\xk-0.2)/11.8)*4.350},3.250);
      \draw[primary!70,thin] ({6.000+((\xk-0.2)/11.8)*4.350},-0.07) -- ({6.000+((\xk-0.2)/11.8)*4.350},0);
      \node[below,font=\tiny\sffamily] at ({6.000+((\xk-0.2)/11.8)*4.350},-0.08) {\xk};
    }

    \fill[accent!14] (6.000,1.676) -- (6.074,1.997) -- (6.147,2.181) -- (6.221,2.306) -- (6.295,2.399) -- (6.369,2.470) -- (6.442,2.527) -- (6.516,2.573) -- (6.590,2.611) -- (6.664,2.642) -- (6.737,2.669) -- (6.811,2.691) -- (6.885,2.710) -- (6.958,2.727) -- (7.032,2.741) -- (7.106,2.754) -- (7.180,2.765) -- (7.253,2.774) -- (7.327,2.783) -- (7.401,2.791) -- (7.475,2.798) -- (7.548,2.804) -- (7.622,2.810) -- (7.696,2.815) -- (7.769,2.820) -- (7.843,2.824) -- (7.917,2.828) -- (7.991,2.832) -- (8.064,2.835) -- (8.138,2.839) -- (8.212,2.842) -- (8.286,2.844) -- (8.359,2.847) -- (8.433,2.849) -- (8.507,2.852) -- (8.581,2.854) -- (8.654,2.856) -- (8.728,2.858) -- (8.802,2.860) -- (8.814,2.860) -- (8.802,2.859) -- (8.728,2.851) -- (8.654,2.842) -- (8.581,2.833) -- (8.507,2.824) -- (8.433,2.814) -- (8.359,2.804) -- (8.286,2.793) -- (8.212,2.781) -- (8.138,2.769) -- (8.064,2.756) -- (7.991,2.742) -- (7.917,2.727) -- (7.843,2.712) -- (7.769,2.695) -- (7.696,2.676) -- (7.622,2.657) -- (7.548,2.636) -- (7.475,2.613) -- (7.401,2.589) -- (7.327,2.562) -- (7.253,2.533) -- (7.180,2.501) -- (7.106,2.465) -- (7.032,2.427) -- (6.958,2.384) -- (6.885,2.336) -- (6.811,2.282) -- (6.737,2.221) -- (6.664,2.152) -- (6.590,2.072) -- (6.516,1.980) -- (6.442,1.872) -- (6.369,1.742) -- (6.295,1.584) -- (6.221,1.385) -- (6.147,1.122) -- (6.074,0.745) -- (6.000,0.094) -- cycle;

    \draw[accent!65,densely dashed,semithick] (6.000,1.677) -- (6.074,2.004) -- (6.147,2.194) -- (6.221,2.330) -- (6.295,2.435) -- (6.369,2.520) -- (6.442,2.593) -- (6.516,2.656) -- (6.590,2.711) -- (6.664,2.761) -- (6.737,2.806) -- (6.811,2.847) -- (6.885,2.884) -- (6.958,2.919) -- (7.032,2.952) -- (7.106,2.982) -- (7.180,3.010) -- (7.253,3.037) -- (7.327,3.063) -- (7.401,3.087) -- (7.475,3.110) -- (7.548,3.132) -- (7.622,3.153) -- (7.696,3.173) -- (7.769,3.192) -- (7.843,3.210) -- (7.917,3.228);
    \draw[alert!65,densely dashed,semithick] (6.000,0.091) -- (6.074,0.743) -- (6.147,1.125) -- (6.221,1.396) -- (6.295,1.606) -- (6.369,1.777) -- (6.442,1.922) -- (6.516,2.048) -- (6.590,2.159) -- (6.664,2.258) -- (6.737,2.347) -- (6.811,2.429) -- (6.885,2.505) -- (6.958,2.574) -- (7.032,2.639) -- (7.106,2.700) -- (7.180,2.757) -- (7.253,2.811) -- (7.327,2.862) -- (7.401,2.910) -- (7.475,2.956) -- (7.548,3.000) -- (7.622,3.042) -- (7.696,3.082) -- (7.769,3.120) -- (7.843,3.157) -- (7.917,3.192) -- (7.991,3.227);
    \draw[accent,very thick] (6.000,1.676) -- (6.074,1.997) -- (6.147,2.181) -- (6.221,2.306) -- (6.295,2.399) -- (6.369,2.470) -- (6.442,2.527) -- (6.516,2.573) -- (6.590,2.611) -- (6.664,2.642) -- (6.737,2.669) -- (6.811,2.691) -- (6.885,2.710) -- (6.958,2.727) -- (7.032,2.741) -- (7.106,2.754) -- (7.180,2.765) -- (7.253,2.774) -- (7.327,2.783) -- (7.401,2.791) -- (7.475,2.798) -- (7.548,2.804) -- (7.622,2.810) -- (7.696,2.815) -- (7.769,2.820) -- (7.843,2.824) -- (7.917,2.828) -- (7.991,2.832) -- (8.064,2.835) -- (8.138,2.839) -- (8.212,2.842) -- (8.286,2.844) -- (8.359,2.847) -- (8.433,2.849) -- (8.507,2.852) -- (8.581,2.854) -- (8.654,2.856) -- (8.728,2.858) -- (8.802,2.860) -- (8.875,2.861) -- (8.949,2.863) -- (9.023,2.864) -- (9.097,2.866) -- (9.170,2.867) -- (9.244,2.869) -- (9.318,2.870) -- (9.392,2.871) -- (9.465,2.872) -- (9.539,2.873) -- (9.613,2.874) -- (9.686,2.875) -- (9.760,2.876) -- (9.834,2.877) -- (9.908,2.878) -- (9.981,2.879) -- (10.055,2.880) -- (10.129,2.881) -- (10.203,2.882) -- (10.276,2.882) -- (10.350,2.883) -- (10.350,2.883);
    \draw[alert,very thick] (6.000,0.094) -- (6.074,0.745) -- (6.147,1.122) -- (6.221,1.385) -- (6.295,1.584) -- (6.369,1.742) -- (6.442,1.872) -- (6.516,1.980) -- (6.590,2.072) -- (6.664,2.152) -- (6.737,2.221) -- (6.811,2.282) -- (6.885,2.336) -- (6.958,2.384) -- (7.032,2.427) -- (7.106,2.465) -- (7.180,2.501) -- (7.253,2.533) -- (7.327,2.562) -- (7.401,2.589) -- (7.475,2.613) -- (7.548,2.636) -- (7.622,2.657) -- (7.696,2.676) -- (7.769,2.695) -- (7.843,2.712) -- (7.917,2.727) -- (7.991,2.742) -- (8.064,2.756) -- (8.138,2.769) -- (8.212,2.781) -- (8.286,2.793) -- (8.359,2.804) -- (8.433,2.814) -- (8.507,2.824) -- (8.581,2.833) -- (8.654,2.842) -- (8.728,2.851) -- (8.802,2.859) -- (8.875,2.866) -- (8.949,2.873) -- (9.023,2.880) -- (9.097,2.887) -- (9.170,2.893) -- (9.244,2.900) -- (9.318,2.905) -- (9.392,2.911) -- (9.465,2.916) -- (9.539,2.922) -- (9.613,2.927) -- (9.686,2.932) -- (9.760,2.936) -- (9.834,2.941) -- (9.908,2.945) -- (9.981,2.949) -- (10.055,2.953) -- (10.129,2.957) -- (10.203,2.961) -- (10.276,2.965) -- (10.350,2.969) -- (10.350,2.969);

    \draw[primary!65,densely dotted] (8.814,0) -- (8.814,2.860);
    \fill[primary] (8.814,2.860) circle (1.7pt);
    \node[anchor=north west,font=\tiny\sffamily,color=primary,inner sep=1pt] at (8.874,2.545) {$\kappa_\times\!\approx\!7.83$};

    \draw[primary,semithick] (6.000,0) rectangle (10.350,3.250);

    \node[anchor=south,font=\tiny\sffamily,color=accent,inner sep=1pt] at (8.230,2.869) {$\varepsilon_-(\kappa)$};
    \node[anchor=north west,font=\tiny\sffamily,color=alert,inner sep=1pt] at (7.309,2.536) {$\varepsilon_+(\kappa)$};
    \node[anchor=center,font=\tiny\sffamily,color=accent!85!black,align=center,inner sep=0.5pt] at (6.387,2.128) {valid\\[-1.5pt]gap};
    \node[anchor=south east,font=\tiny\sffamily,color=black!65] at (10.300,0.05) {dashed: small-$\kappa$ asymp.};

    \node[above,font=\scriptsize\bfseries\sffamily,color=primary] at (8.175,3.330) {(b) Spectral Separation ($L\!=\!16$, $a\!=\!b\!=\!1$)};
    \node[below,font=\tiny\sffamily] at (8.175,-0.34) {Concentration parameter $\kappa$};
    \node[rotate=90,above,font=\tiny\sffamily] at (5.340,1.625) {Threshold ($\log_{10}$)};
  \end{scope}

  \begin{scope}
    \foreach \yg in {0, 2, 4, 6, 8} {
      \draw[black!10,very thin] (12.000,{(\yg/8.8)*3.250}) -- (16.350,{(\yg/8.8)*3.250});
      \draw[primary!70,thin] (11.930,{(\yg/8.8)*3.250}) -- (12.000,{(\yg/8.8)*3.250});
      \node[left,font=\tiny\sffamily] at (11.920,{(\yg/8.8)*3.250}) {\yg};
    }
    \foreach \xb in {1, 8, 16, 24, 32} {
      \draw[black!10,very thin] ({12.000+((\xb-1.0)/31.0)*4.350},0) -- ({12.000+((\xb-1.0)/31.0)*4.350},3.250);
      \draw[primary!70,thin] ({12.000+((\xb-1.0)/31.0)*4.350},-0.07) -- ({12.000+((\xb-1.0)/31.0)*4.350},0);
      \node[below,font=\tiny\sffamily] at ({12.000+((\xb-1.0)/31.0)*4.350},-0.08) {\xb};
    }

    \draw[primary!55,densely dashed,thin] (12.000,0.739) -- (16.350,0.739);
    \draw[accent!65,densely dashed,thin] (12.000,1.477) -- (16.350,1.477);
    \draw[alert!65,densely dashed,thin] (12.000,2.955) -- (16.350,2.955);

    \draw[primary,thick] (12.000,0.369) -- (12.140,0.492) -- (12.281,0.554) -- (12.421,0.591) -- (12.561,0.616) -- (12.702,0.633) -- (12.842,0.646) -- (12.982,0.657) -- (13.123,0.665) -- (13.263,0.671) -- (13.403,0.677) -- (13.544,0.682) -- (13.684,0.686) -- (13.824,0.689) -- (13.965,0.692) -- (14.105,0.695) -- (14.245,0.698) -- (14.385,0.700) -- (14.526,0.702) -- (14.666,0.703) -- (14.806,0.705) -- (14.947,0.707) -- (15.087,0.708) -- (15.227,0.709) -- (15.368,0.710) -- (15.508,0.711) -- (15.648,0.712) -- (15.789,0.713) -- (15.929,0.714) -- (16.069,0.715) -- (16.210,0.716) -- (16.350,0.716);
    \draw[accent,very thick] (12.000,0.369) -- (12.140,0.591) -- (12.281,0.739) -- (12.421,0.844) -- (12.561,0.923) -- (12.702,0.985) -- (12.842,1.034) -- (12.982,1.074) -- (13.123,1.108) -- (13.263,1.136) -- (13.403,1.161) -- (13.544,1.182) -- (13.684,1.200) -- (13.824,1.217) -- (13.965,1.231) -- (14.105,1.244) -- (14.245,1.256) -- (14.385,1.266) -- (14.526,1.276) -- (14.666,1.285) -- (14.806,1.293) -- (14.947,1.300) -- (15.087,1.307) -- (15.227,1.313) -- (15.368,1.319) -- (15.508,1.324) -- (15.648,1.330) -- (15.789,1.334) -- (15.929,1.339) -- (16.069,1.343) -- (16.210,1.347) -- (16.350,1.351);
    \draw[alert,very thick] (12.000,0.369) -- (12.140,0.657) -- (12.281,0.886) -- (12.421,1.074) -- (12.561,1.231) -- (12.702,1.364) -- (12.842,1.477) -- (12.982,1.576) -- (13.123,1.662) -- (13.263,1.738) -- (13.403,1.806) -- (13.544,1.866) -- (13.684,1.920) -- (13.824,1.970) -- (13.965,2.014) -- (14.105,2.055) -- (14.245,2.093) -- (14.385,2.127) -- (14.526,2.159) -- (14.666,2.189) -- (14.806,2.216) -- (14.947,2.241) -- (15.087,2.265) -- (15.227,2.287) -- (15.368,2.308) -- (15.508,2.328) -- (15.648,2.346) -- (15.789,2.364) -- (15.929,2.380) -- (16.069,2.396) -- (16.210,2.410) -- (16.350,2.424);

    \draw[primary,semithick] (12.000,0) rectangle (16.350,3.250);

    \node[anchor=south east,font=\tiny\sffamily,color=alert,inner sep=1pt] at (16.290,2.985) {$s/p\!=\!7\;(\to 8)$};
    \node[anchor=south east,font=\tiny\sffamily,color=accent,inner sep=1pt] at (16.290,1.507) {$s/p\!=\!3\;(\to 4)$};
    \node[anchor=south east,font=\tiny\sffamily,color=primary,inner sep=1pt] at (16.290,0.769) {$s/p\!=\!1\;(\to 2)$};

    \node[above,font=\scriptsize\bfseries\sffamily,color=primary] at (14.175,3.330) {(c) Collective Scaling $G_C(B)$};
    \node[below,font=\tiny\sffamily] at (14.175,-0.34) {Number of sessions $B$};
    \node[rotate=90,above,font=\tiny\sffamily] at (11.620,1.625) {Gain $G_C = R_{\mathrm{ind}}/R_{\mathrm{joint}}$};
  \end{scope}
\end{tikzpicture}